\def\year{2022}\relax
\documentclass[letterpaper]{article} 
\usepackage{aaai22}  
\usepackage{times}  
\usepackage{helvet}  
\usepackage{courier}  
\usepackage[hyphens]{url}  
\usepackage{graphicx} 
\usepackage{natbib}  
\usepackage{caption} 
\DeclareCaptionStyle{ruled}{labelfont=normalfont,labelsep=colon,strut=off} 
\usepackage{algorithm}
\usepackage[noend]{algpseudocode}
\usepackage{amssymb}
\usepackage{amsmath}
\usepackage{amsthm}
\usepackage{blindtext}
\usepackage{todonotes}
\usepackage{enumitem}
\usepackage{caption}
\usepackage{subcaption}
\usepackage{placeins}
\usepackage{thmtools,thm-restate}
\usepackage{microtype}

\newtheorem{problem}{Problem}
\newtheorem{assumption}{Assumption}

\newtheorem{lemma}{Lemma}
\newtheorem*{proposition*}{Proposition}

\DeclareMathOperator*{\argmax}{arg\,max}

\renewcommand{\epsilon}{\varepsilon}

\usepackage{commath}
\usepackage{newfloat}
\usepackage{listings}
\floatstyle{ruled}
\newfloat{listing}{tb}{lst}{}
\floatname{listing}{Listing}
\title{Optimal Admission Control for Multiclass Queues with Time-Varying Arrival Rates via State Abstraction}
\author{
    Marc Rigter,\textsuperscript{\rm 1, 2}
    Danial Dervovic,\textsuperscript{\rm 2}
    Parisa Hassanzadeh,\textsuperscript{\rm 2}\\
    Jason Long,\textsuperscript{\rm 2}
    Parisa Zehtabi, \textsuperscript{\rm 2}
    Daniele Magazzeni \textsuperscript{\rm 2}
}
\affiliations{
\textsuperscript{\rm 1}Oxford Robotics Institute, University of Oxford\\
\textsuperscript{\rm 2}J. P. Morgan AI Research\\
mrigter@robots.ox.ac.uk,
\{danial.dervovic, parisa.hassanzadeh, jason.x.long, parisa.zehtabi, daniele.magazzeni\}@jpmorgan.com

}

\usepackage{bibentry}

\begin{document}

\maketitle

\begin{abstract}
We consider a novel queuing problem where the decision-maker must choose to accept or reject randomly arriving tasks into a no buffer queue which are processed by $N$ identical servers. Each task has a~\emph{price}, which is a positive real number, and a~\emph{class}. Each class of task has a different price distribution and service rate, and arrives according to an inhomogenous Poisson process. The objective is to decide which tasks to accept so that the total price of tasks processed is maximised over a finite horizon. We formulate the problem as a discrete time Markov Decision Process (MDP) with a hybrid state space. We show that the optimal value function has a specific structure, which enables us to solve the hybrid MDP exactly. Moreover, we prove that as the time step is reduced, the discrete time solution approaches the optimal solution to the original continuous time problem. To improve the scalability of our approach to a greater number of task classes, we present an approximation based on state abstraction. We validate our approach on synthetic data, as well as a real financial fraud data set, which is the motivating application for this work.
\end{abstract}

\section{Introduction}
In many service systems, the rate at which tasks arrive may greatly exceed the capacity of servers to process the tasks.
We are motivated by the problem of financial fraud detection and monitoring~\cite{dal2014learned}, where the rate at which suspicious transactions are detected may considerably exceed the rate at which human operators (servers) can intervene, for example, by holding payments and calling customers to verify their identities.

We associate each task with a~\emph{price}, and a~\emph{class}. 
The price of a task represents the value associated with processing that task. 
In the financial fraud domain, the value of intervening to verify a suspicious transaction may be equal to that transaction's monetary value.
Tasks of each class may have different service time requirements and price distributions. 
For example, validating a high-value bank transfer (task class A) may require a more thorough process, and therefore a slower service rate, than a low-value credit card payment (task class B).
In the financial fraud domain, the rate of task arrivals varies substantially with time: the volume of transactions is far greater during business hours than at night.
Therefore, we allow the arrival rate of each task class to vary over time.

For such systems, we are interested in~\emph{admission control}, the problem of deciding which tasks should be processed, and which tasks should be ignored.
We seek to optimise the expected total price of tasks processed over a finite horizon by a fixed number of identical servers.

The optimal control of multiclass queues has been considered by previous works~\cite{ata2006dynamic, ata2013scheduling, bertsimas1994optimization, cao2016optimal,  harrison2004dynamic, klimov1974time}.
However, all of these works consider an infinite horizon with constant task arrival rates, which is a considerably easier problem as the optimal solution is stationary (i.e. does not depend on time).
~\citet{yoon2004optimal} consider optimal admission control in a time-varying problem where there is only one task class, and the task prices can only take on values from a discrete set. 
The authors present an approximation algorithm, but do not provide any theoretical analysis.

In contrast, we consider multiple task classes with time-varying arrival rates and allow task prices to be continuous. 
We rigorously prove that our discrete-time algorithm approaches the optimal solution as the time step is reduced.
To improve the scalability of our approach to a greater number of task classes, we present an approximation based on state abstraction to reduce the size of the state space.

Our main contributions are: 1) formulating a novel multiclass queuing problem with continuous task prices and time-varying arrival rates, 2) rigorously proving that our discrete-time algorithm approaches the optimal continuous time solution as the time step is reduced, and 3) an approach based on state abstraction to improve scalability.
We validate our approach on synthetic domains, as well as a real financial fraud data set. 
Our results demonstrate that our approach is computationally efficient and significantly improves the average total price of tasks processed compared to baselines.
\section{Related Work}
The stochastic sequential assignment problem (SSAP) with random arrival times was proposed by~\citet{albright1974optimal}, building on earlier work from~\citet{derman1972sequential}.
In this problem, tasks arrive according to an inhomogenous Poisson process and must be accepted or rejected upon arrival.
Furthermore, each task is associated with a continuous reward value drawn from an arbitrary distribution.
This problem is also referred to as the dynamic and stochastic knapsack problem~\cite{kleywegt1998dynamic}.
The SSAP was recently revisited by~\citet{dervovic2021non} to address the problem when the arrival and reward distributions must be learnt from historical data.
In contrast to our work, the SSAP problem assumes that the number of tasks to be accepted is known a priori, whereas we assume that tasks have stochastic processing times. 


Semi-MDPs were introduced by~\cite{jewell1963markov} and~\cite{howard1963semi} and have been applied extensively to the optimal control of queues~\cite{rue1985application, stidham1985optimal, sennott1989average}. However, these works assume that the arrival rate of tasks is~\emph{stationary}, i.e. does not vary with time, and optimise the expected reward over an infinite horizon. Under these assumptions, the optimal policy is also stationary~\cite{sennott1989average}.  
The queue control problem with time-varying arrival rates that we address is an instance of a non-stationary Semi-MDP~\cite{ghosh2013non}.
Finding the exact optimal solution for non-stationary Semi-MDPs is generally not possible as the solution depends on time, and the optimality equations involve integrals over time which are intractable for most problems~\cite{mcmahon2008time}.
~\citet{mamer1986successive} proposes an approximation which limits the number of possible transitions, while while~\citet{duckworth2021time} present an approximation algorithm which utilises sample-based planning.
%

Multiclass queues have multiple task classes which may have different arrival rates and service rates. The optimisation of multiclass queues has been addressed in numerous works~\cite{ata2006dynamic, ata2013scheduling, bertsimas1994optimization, cao2016optimal,  harrison2004dynamic, klimov1974time}. However, none of these existing works consider time-varying arrival rates, which is a core focus of this work.
%

The most related work to ours is that of~\citet{yoon2004optimal}.~\citet{yoon2004optimal} consider tasks arriving according to a periodic inhomogenous (i.e. time-varying) Poisson process.
All tasks have the same service rate, but may take on price values from a discrete set.
The authors present a piecewise stationary approximation of the optimal admission policy.
%
%
The authors do not provide performance guarantees for this approach.
In contrast to~\citet{yoon2004optimal}, our work a) admits continuous task prices values, b) considers multiple task classes with different service rates and time-varying arrival rates, and c) we rigorously prove that our discrete-time algorithm approaches the optimal continuous-time solution as the time step is reduced.

\section{Problem Formulation}
\begin{problem}[Continuous-time formulation]
\normalfont
\label{prob:continuous_time}
Consider a multi-server queue with $N_{serv}$ identical servers. A \emph{task} is defined by a pair $\tau = (k, p)$, where $k \in K = \{1, 2, \ldots, |K| \}$ is the \emph{task class}, and $p \in [0, \infty)$ is the \emph{task price}. Tasks of each class arrive according to an inhomogenous Poisson process, $\Lambda_k(t)$. The arrival rate function for each class is assumed to be finite and Lipschitz continuous with constant $C_k$. The price distribution for each task is independent of the arrival time. The probability density function for the price of task class $k$ is $f_k(p)$, with corresponding cumulative density function, $F_k(p)$. When a new task arrives, the decision maker observes the task class and price, but does not observe the required processing time. If there is at least one free server who is not currently processing a task, the decision-maker can decide to accept or reject the new task. If the new task is accepted, it immediately starts being processed by one of the servers who was previously free. If there are no servers who are free, the decision-maker cannot accept new tasks (i.e. there is no buffer or ``waiting area"). For each task class, the time to process a task is exponentially distributed with rate $\mu_k$. The objective is to find a decision-making policy to maximise the expected total price of tasks which are completed during a finite horizon of length $t_H$. 
\end{problem}
\section{Discrete-Time Solution}
In Problem~\ref{prob:continuous_time}, tasks arrive in continuous time and the rate of arrivals varies continuously with time. To solve Problem~\ref{prob:continuous_time} exactly, it can be formalised as a Semi-MDP with time-varying dynamics~\cite{ghosh2013non}. However, solving Semi-MDPs with time-varying dynamics exactly is intractable~\cite{mcmahon2008time}. Therefore, we take the approach of discretising time and approximate Problem~\ref{prob:continuous_time} as a discrete-time hybrid factored MDP (HMDP) with finite horizon. In Proposition~\ref{eq:approx_error} we prove that as the resolution of the discretisation of time approaches zero, the solution to our HMDP formulation approaches the solution to Problem~\ref{prob:continuous_time}. We begin by defining HMDPs.

\subsection{Discrete-Time Hybrid Factored MDPs}
A discrete-time hybrid factored MDP (HMDP)~\cite{kveton2006solving} with a finite horizon is a tuple, $\mathcal{M} = (\mathbf{X}, A, R, P,  D)$. $D = \{t_0, t_1, \ldots, t_H \}$ is a finite sequence of decision epochs, or time steps.
For simplicity, we assume that the time step size, $\Delta t=t_{i+1} - t_i$, is a constant.
$\mathbf{X}$ is a state space represented by a set of state variables, $\{X_1, X_2, \ldots, X_n \}$. 
A state  is defined by a vector $\mathbf{x}$ of assignments to each state variable, which splits into discrete and continuous components denoted by $\mathbf{x} = (\mathbf{x}_D, \mathbf{x}_C)$.
For any state vector, $\mathbf{x}$, we write $\mathbf{x}[j]$ to refer to the value of state variable $j$.
$A$ is a finite set of actions. 
$R : \mathbf{X} \times A \times D \rightarrow \mathbb{R}_{\geq 0}$ is a non-negative reward function. 
$P : \mathbf{X} \times A  \times D \times \mathbf{X}  \rightarrow \mathbb{R}_{\geq0}$ is a time-varying transition function which describes the transition dynamics conditioned upon the previous state, action, and time step.
A deterministic Markovian policy is a mapping from the state and time step to an action: $\pi : \mathbf{X} \times D \rightarrow A$.
The objective is to find a policy which optimises the expected total reward over the finite horizon,  $\mathbb{E}[\sum_{t=t_0}^{t_H}R(\mathbf{x}_t, \pi(\mathbf{x}_t))]$. The optimal value function~\cite{bellman1966dynamic}, $V_{\pi^*}$, satisfies
\vspace{-2mm}
\begin{multline}
    \label{eq:bellman}
    V_{\pi^*}(\mathbf{x}, t_i) = \max_{a \in A}\Big[R(\mathbf{x}, a, t_i) + \\ \sum_{\mathbf{x}'_D} \int_{\mathbf{x}'_C} P(\mathbf{x'} \mid \mathbf{x}, a, t_i) \cdot V_{\pi^*}(\mathbf{x}', t_{i+1}) \dif \mathbf{x}_C \Big],
\end{multline}
\noindent where we denote the optimal policy corresponding to Eq.~\ref{eq:bellman} by $\pi^*$. The optimal $Q$-values are
\vspace{-0.5mm}
\begin{multline}
    \label{eq:q-val}
    Q_{\pi^*}(\mathbf{x}, a, t_i) =  R(\mathbf{x}, a, t_i) + \\  \sum_{\mathbf{x}'_D} \int_{\mathbf{x}'_C} P(\mathbf{x'} \mid \mathbf{x}, a, t_i) \cdot V_{\pi^*}(\mathbf{x}', t_{i+1}) \dif \mathbf{x}_C .
\end{multline}

\subsection{HDMP Formulation}
To approximate the continuous-time formulation in Problem~\ref{prob:continuous_time} by an HMDP, we make the following assumption. 
\vspace{-0.5mm}
\begin{assumption}
    \label{ass:arrivals}
    At each decision epoch, at most one task can be accepted.
\end{assumption}
\vspace{-0.5mm}
Assumption~\ref{ass:arrivals} is necessary to ensure that the action space is finite. If a task, $\tau$, arrives between $t_i$ and $t_{i+1}$, at decision epoch $t_{i+1}$ the decision maker can choose to accept or reject $\tau$.  In our formulation, we assume that if more than one task arrives between decision epochs, one of these tasks is selected uniformly at random to be the task that the decision maker can choose to accept or reject. The rest of the arrivals are automatically rejected. Intuitively, as the time step size is reduced, it becomes less likely that tasks will be automatically rejected due to Assumption~\ref{ass:arrivals}. In Proposition~\ref{eq:approx_error} we show that as the time step size approaches zero, the solution to the HDMP formulation approaches the solution to Problem~\ref{prob:continuous_time}.

We now introduce our HDMP formulation of the problem which we refer to as the Stochastic Task Admission HDMP (STA-HMDP).
The STA-HDMP has the following set of state variables: $\{n_1, \ldots, n_{|K|},k^+, p\}$.
$n_k$ is the number of servers who are currently processing tasks of class $k$. 
Because there are $N_{serv}$ servers, $\sum_{k\in K} n_k \leq N_{serv}$.
If task $\tau = (k_{\mathtt{arr}}, p_{\mathtt{arr}})$ arrives between $t_i$ and $t_{i+1}$, then at time $t_{i+1}$: $k^+ = k_{\mathtt{arr}}$ and $p = p_{\mathtt{arr}}$. 
%
%
If no task has arrived, then we say $k^+ = \bot$ and $p = 0$, i.e. $F_\bot(p) = H (p)$, where $H$ is the Heaviside step function.  
Thus, $k^+$ equals the task class if a task has arrived since the last decision epoch, and equals $\bot$ if no task has arrived.

Note that we must include the combination of task classes currently being processed, $(n_1, \ldots, n_{|K|})$, in the state because each task class has a different service rate.
Therefore, the rate at which tasks are being processed depends not only on the number of tasks being processed, but also on the classes of tasks being processed.
To simplify the notation, at times we will use the abbreviation ${\mathbf{n} = (n_1, \ldots, n_{|K|}) \in \mathbf{N}}$, where $\mathbf{N}$ is the set of possible task class combinations.

The action space is $A = \{\mathtt{acc}, \mathtt{rej}\}$, corresponding to accepting or rejecting a task. 
The $\mathtt{acc}$ action can only be executed if $k^+ \neq \bot$ and  $\sum_{k\in K} n_k < N_{serv}$, i.e. a task has arrived and there is at least one free server available. The reward for accepting a task is \vspace{-1mm}
\begin{equation}
    \label{eq:reward_func}
    R(\mathbf{x}, \mathtt{acc}, t) = 
        \mathbf{x}[p] \cdot \Pr(\mathtt{fin}(\mathbf{x}[k^+]) \mid t) ,
        \vspace{-1mm}
\end{equation}
where $\Pr(\mathtt{fin}(k) \mid t) = 1 - \exp(-(t_H - t)\mu_{k})$ is the probability that task class $k$ will be completed before the horizon, $t_H$. Eq.~\ref{eq:reward_func} computes the expected price  as the task price is only received if the task is completed before the horizon (Problem~\ref{prob:continuous_time}). 
The reward for rejecting a task is zero.

We now define the transition function for the $\mathtt{rej}$ action.
Due to independence between state variables, we can write the transition function in the following product form
\vspace{-1mm}
\begin{multline}
    \label{eq:indep_trans_func}
    \small
    P(\mathbf{x}' \mid \mathbf{x}, \mathtt{rej}, t_i) = 
    \Pr(\mathbf{x}'[k^+] \mid t_i) \times \\ f_{\mathbf{x}'[k^+]}(\mathbf{x}'[p])  \times \prod_{k=1}^K  \Pr(\mathbf{x}'[n_k] \mid  \mathbf{x}[n_k]),
    \vspace{-1mm}
\end{multline}
\noindent where the probability that no task arrives is    
\begin{equation}
    \textstyle \Pr\big( \mathbf{x'}[k^+] = \bot \mid t_i \big) =  \exp(- \Delta t \sum_{k = 1}^{K}\Lambda_k(t_i) ), 
 \end{equation}
 
\noindent and the probability that a task of class $k'$ arrives is \vspace{-1mm}
\begin{multline}
    \Pr(\mathbf{x}'[k^+]=k' \mid t_i) = \\
    \frac{\Lambda_{k'}(t_i)}{\sum_{k = 1}^{K}\Lambda_{k}(t_i)}\big[1 -\Pr\big( \mathbf{x'}[k^+] = \bot \mid t_i \big) \big].
    \vspace{-1mm}
\end{multline}

The transition probabilities for $n_k$, the number of servers currently processing a task of class $k$, can be computed as \vspace{-1mm}
\begin{multline}
    \label{eq:server_fin}
    \Pr(\mathbf{x}'[n_k] \mid  \mathbf{x}[n_k]) = \\
    \hspace{3mm}
    \begin{cases}
        \binom{\mathbf{x}[n_k]}{\mathbf{x}[n_k] - \mathbf{x}'[n_k]} \times (1 - e^{-\mu_k \Delta t})^{(\mathbf{x}[n_k] - \mathbf{x}'[n_k])} \times \vspace{0mm} \\ \hspace{70pt}e^{-\mu_k \mathbf{x}'[n_k] \Delta t }, \hspace{5pt} \textnormal{if } \mathbf{x}'[n_k] \leq \mathbf{x}[n_k]. \vspace{0.5mm}\\
        0, \textnormal{ otherwise.} \vspace{-2mm}
    \end{cases}
    \raisetag{15mm}
\end{multline}
\noindent where $\binom{\cdot}{\cdot}$ denotes the binomial coefficient. Eq.~\ref{eq:server_fin} follows a binomial distribution because each server finishing or not finishing during $\Delta t$ is an independent Bernoulli trial.

We now consider the $\mathtt{acc}$ action. If a task of class $k$ is accepted, the reward $ R(\mathbf{x}, \mathtt{acc}, t_i)$ is received. The state instantaneously transitions to a successor state, $\mathbf{x}^{\mathtt{acc}}_{(\mathbf{n}, k)}$, where there is additional server processing task class $k$, and no task available to be accepted. Therefore, the $Q$-value for accepting a task can be computed by 
\begin{multline}
    \label{eq:qv_accept_act}
    Q_{\pi^*}(\mathbf{x}, \mathtt{acc}, t_i)  =  \mathbf{x}[p] \cdot \Pr( \mathtt{fin}(\mathbf{x}[k^+]) \mid t_i)  \vspace{0.5mm}\\
    + V_{\pi^*}(\mathbf{x}^{\mathtt{acc}}_{(\mathbf{x}[\mathbf{n}], \mathbf{x}[k^+])}, t_{i}), \hspace{5pt} \textnormal{where}
\end{multline}
\vspace{-3mm}
\begin{equation}
    \label{eq:x_accept}
    \mathbf{x}^{\mathtt{acc}}_{(\mathbf{x}[\mathbf{n}], \mathbf{x}[k^+])}[n_k] = 
    \begin{cases}
      \mathbf{x}[n_k] + 1, \textnormal{ if }  \mathbf{x}[k^+]  = k \\
       \mathbf{x}[n_k], \textnormal{ otherwise. }
    \end{cases}
\end{equation}
\noindent and $\mathbf{x}^{\mathtt{acc}}_{(\mathbf{x}[\mathbf{n}], \mathbf{x}[k^+])}[k^+] = \bot$. 
The subscript of $\mathbf{x}^{\mathtt{acc}}_{(\mathbf{n}, k)}$ indicates that it is the successor state after accepting a task of class $k$ when the combination of tasks being processed was $\mathbf{n}$.



\subsection{STA-HDMP Solution Algorithm}
General solutions to HMDPs resort to function approximation due to the difficulty of optimising over the continuous state space~\cite{kveton2006solving}. However, as we shall show in this section, it is possible to solve the STA-HMDP exactly (i.e. without function approximation) using a finite number of Bellman backups. This is due to the specific structure of the optimal value function in the STA-HDMP with respect to the continuous state variable, $p$. 

As a first step, we make the observation that the $Q$-value for rejecting the task only depends on $\mathbf{n}$ and $t$, i.e. $Q_{\pi^*}(\mathbf{x}, \mathtt{rej}, t) = Q_{\pi^*}(\mathbf{x}', \mathtt{rej}, t)$ if $\mathbf{x}[\mathbf{n}] = \mathbf{x}'[\mathbf{n}]$. 
Intuitively, this is because if a task is rejected, no immediate reward is received, and the transition dynamics to the next time step are the same irrespective of what task class and price was rejected. 
To make this explicit, we will write $Q_{\pi^*}(\mathbf{x}[\mathbf{n}], \mathtt{rej}, t)$ in place of $Q_{\pi^*}(\mathbf{x}, \mathtt{rej}, t)$.
Following from Eq.~\ref{eq:q-val}, the $Q$-value for the $\mathtt{rej}$ action is \vspace{-0.5mm}
\begin{multline}
    \label{eq:q_reject}
    Q_{\pi^*}(\mathbf{x}[\mathbf{n}], \mathtt{rej}, t_i) =  \\ \sum_{\mathbf{x}'[\mathbf{n}]}  \sum_{\mathbf{x}'[{k^+}]} \int_{\mathbf{x}'[p]} P(\mathbf{x'} \mid \mathbf{x}, \mathtt{rej}, t_i) \cdot V_{\pi^*}(\mathbf{x}', t_{i+1})\cdot \dif \, (\mathbf{x}'[p]) \raisetag{15mm}
\end{multline}

We now show that the optimal value function is piecewise-linear with respect to the task price, $p$. Full proofs of all propositions are in the supplementary material (supp. mat.).
\begin{restatable}{proposition}{pwlvf}
    \label{prop:pwl_vf}
    Let $V_{\pi^*}(\mathbf{x}, t)$ be the optimal value function for the STA-HMDP. $V_{\pi^*}(\mathbf{x}, t)$ has the following form 
\end{restatable}
\vspace{-5mm}
\begin{multline}
    V_{\pi^*}(\mathbf{x}, t) = \\
    \begin{cases}
         \Pr\big(\mathtt{fin}(\mathbf{x}[{k^+}])\mid t \big) \cdot \big(\mathbf{x}[p] - p^*_{cr}(\mathbf{x}[\mathbf{n}], \mathbf{x}[{k^+}], t) \big)  
            \\  \hspace{122pt}+ Q_{\pi^*}(\mathbf{x}[\mathbf{n}],  \mathtt{rej}, t), \\
        \hspace{28pt} \textnormal{if } \mathbf{x}[{k^+}] \neq \bot \textnormal{ and } \mathbf{x}[p] \geq p^*_{cr}(\mathbf{x}[\mathbf{n}], \mathbf{x}[{k^+}], t). \vspace{1mm}\\
        Q_{\pi^*}(\mathbf{x}[\mathbf{n}], \mathtt{rej}, t), \textnormal{ otherwise.}
    \end{cases} \raisetag{5mm}
\end{multline}
\noindent where $p^*_{cr}$ is the \emph{critical price function}, defined as
\begin{multline}
    \label{eq:def_pcrit}
    p^*_{cr}(\mathbf{x}[\mathbf{n}], \mathbf{x}[{k^+}], t) =  \frac{1}{\Pr(\mathtt{fin}(\mathbf{x}[{k^+}]) \mid t) } \times \\
    \Big(Q_{\pi^*}(\mathbf{x}[\mathbf{n}], \mathtt{rej}, t) -  Q_{\pi^*}(\mathbf{x}^{\mathtt{acc}}_{(\mathbf{x}[\mathbf{n}], \mathbf{x}[k^+])}[\mathbf{n}], \mathtt{rej}, t)
    \Big). \raisetag{14mm}
\end{multline}
\noindent if $\mathbf{x}[{k^+}] \neq \bot$ and $t < t_H$. We define $p^*_{cr}(\mathbf{x}[\mathbf{n}], \mathbf{x}[{k^+}], t) = 0$ if $\mathbf{x}[{k^+}] = \bot$ or $t = t_H$.

Proposition~\ref{prop:pwl_vf} shows that the optimal value function is piecewise-linear with respect to $\mathbf{x}[p]$.
For a given $\mathbf{x}[\mathbf{n}], \mathbf{x}[{k^+}]$, and $t$, if we know $Q_{\pi^*}(\mathbf{x}[\mathbf{n}], \mathtt{rej}, t)$ and  the critical price function, we can compute the value function for any value of $\mathbf{x}[p]$. 
Additionally, the optimal policy is to accept tasks only if the task price exceeds the critical price.

\begin{restatable}{proposition}{thrpol}[Optimal threshold policy]
    \label{prop:thr_pol}
    The optimal policy, $\pi^*$, for the STA-HMDP may be expressed as follows
\begin{align*}
    \pi^*(\mathbf{x}, t) \hspace{-0.5mm} =  \hspace{-0.5mm}
    \begin{cases}
        \mathtt{acc}, \hspace{2pt} \textnormal{if } \mathbf{x}[{k^+}] \neq \bot \textnormal { and } \mathbf{x}[p] \geq p^*_{cr}(\mathbf{x}[\mathbf{n}], \mathbf{x}[{k^+}], t) \\
        \mathtt{rej}, \hspace{2pt} \textnormal{otherwise}
    \end{cases}
     \vspace{-1mm}
\end{align*}
\end{restatable}
Propositions~\ref{prop:pwl_vf} and \ref{prop:thr_pol} enable us to compute the optimal solution for all states using a finite number of Bellman backups. This is because to define the optimal value and policy for any state and time, with any continuous price, we only need to know $p^*_{cr}(\mathbf{n}, {k}, t) $ and $Q_{\pi^*}(\mathbf{n}, \mathtt{rej}, t)$ for all $\mathbf{n} \in \mathbf{N}, k \in K$, and $t \in D$. 
Therefore, there are only a finite number of $Q$-values and critical prices that we need to compute. 

One remaining issue is that Eq.~\ref{eq:q_reject} contains integrals which would be expensive to compute at every Bellman backup. In Proposition~\ref{prop:simple_calcs} we show that the computations in Eq.~\ref{eq:q_reject} can be simplified in the following manner.
\begin{restatable}{proposition}{msf}[Mean shortage function]
    \label{prop:simple_calcs}
    \vspace{-3mm}
\end{restatable}
\begin{align}
    \begin{split}
    \label{eq:simple_calcs}
     &  Q_{\pi^*}(\mathbf{x}[\mathbf{n}], \mathtt{rej}, t_i) = 
    \sum_{\mathbf{x}'[\mathbf{n}]} \prod_{k=1}^K  \Pr(\mathbf{x}'[n_k] \mid  \mathbf{x}[n_k]) \cdot \vspace{-1mm} \\  
    & \sum_{\mathbf{x}'[{k^+}]} \Pr(\mathbf{x}'[{k^+}]= {k^+}' \mid t_i) \cdot 
    \Big[ Q_{\pi^*}(\mathbf{x}'[\mathbf{n}], \mathtt{rej}, t_{i+1} ) +  \vspace{-1.5mm} \\ 
    & \Pr\big(\mathtt{fin}(\mathbf{x}'[{k^+}])  \mid t_{i+1} \big) \cdot  \phi_{\mathbf{x}'[{k^+}]} \big(p^*_{cr}(\mathbf{x}'[\mathbf{n}], \mathbf{x}'[{k^+}], t_{i+1}) \big) \Big]
    \raisetag{26mm}
    \end{split}
\end{align}

\noindent where $\phi_{{k^+}}(p)$ is the \emph{mean shortage function} of the price distribution for task class ${k^+}$ defined as $
    { \phi_{{k^+}}(p) = \int_p^\infty (1 - {F_{{k^+}}}(y)) \dif y.}
$

Proposition~\ref{prop:simple_calcs} means that we do not have to compute the integrals in Eq.~\ref{eq:q_reject} separately for each Bellman backup, but instead we can query the mean shortage function. 
The mean shortage function can be computed in closed form for many common distributions (e.g. exponential, Pareto), and can be computed efficiently for the nonparametric representation of the task price distribution used in~\citet{dervovic2021non}.

We are now ready to introduce our finite horizon value iteration algorithm for the STA-HMDP (Alg.~\ref{alg:1}). In the algorithm, we first initialise the value function to zero at the horizon, $t_H$. We then iterate over each time step backwards in time. For a given time step, we 1) compute the $Q$-value corresponding to the $\mathtt{rej}$ action using Eq.~\ref{eq:q_reject} and~\ref{eq:simple_calcs}, and 2) compute the critical price values using Eq.~\ref{eq:def_pcrit}. After we have computed the critical price function for all time steps, we can derive the optimal policy using Proposition~\ref{prop:thr_pol}.
\setlength{\intextsep}{6pt}
\renewcommand\algorithmicthen{:}
\renewcommand\algorithmicdo{:}
\begin{algorithm}[b!th]
\caption{Value iteration for STA-HDMP \label{alg:1}}
\begin{algorithmic}
\State {initialise} $Q_{\pi^*}(\mathbf{n}, \mathtt{rej}, t_H) = 0
\textnormal{ for all } \mathbf{n} \in \mathbf{N}$
\For{ $t_i = t_{H-1}, t_{H-2}, \ldots, t_0$}
    \For{$\mathbf{n} \in \mathbf{N}$}
        \State{compute $Q_{\pi^*}(\mathbf{n}, \mathtt{rej}, t_i)$ using Eq.~\ref{eq:q_reject} and~\ref{eq:simple_calcs}}
    \EndFor
    \For{$\mathbf{n} \in \mathbf{N}$}
        \For{$k \in K$}
            \State{compute $p^*_{cr}(\mathbf{n}, k, t_i)$ using Eq.~\ref{eq:def_pcrit}}
        \EndFor
    \EndFor
\EndFor
\end{algorithmic}
\end{algorithm}

\subsection{STA-HDMP Approximation Error}
We now establish that as the time step size, $\Delta t$, approaches zero the optimal solution to the STA-HMDP approaches the optimal continuous time solution to Problem~\ref{prob:continuous_time}.

\begin{restatable}{proposition}{approxerr}
 \label{eq:approx_error}
    Let $\overline{V}^*_\pi(\mathbf{x}, t)$, and $\overline{p}^*_{cr}(\mathbf{n}, {k^+}, t)$ be the optimal value function and critical price function for the continuous time problem defined by Problem~\ref{prob:continuous_time}. Let $V_{\pi^*}(\mathbf{x}, t_i)$, and $p^*_{cr}(\mathbf{n}, {k^+}, t_i)$ be the optimal value function and critical price function for the corresponding STA-HMDP at decision epochs $t_i \in D$. Then: \vspace{-1mm}
\begin{equation*}
    \lim_{\Delta t \rightarrow 0^+} \bigm\lvert \overline{V}_{\pi^*}(\mathbf{x}, t_i) - V_{\pi^*}(\mathbf{x}, t_i) \bigm\lvert = 0, \ \forall\ \mathbf{x},\ t_i \in D
\end{equation*}
 \begin{equation*}
    \lim_{\Delta t \rightarrow 0^+} \bigm\lvert \overline{p}^*_{cr}(\mathbf{n}, {k^+}, t_i) - p^*_{cr}(\mathbf{n}, {k^+}, t_i) \bigm\lvert = 0,\ \forall\ \mathbf{n},\ {k^+},\ t_i \in D
    \vspace{-1mm}
\end{equation*}
\end{restatable}
\section{Approximation via State Abstraction}
In the previous section, we have shown how to solve the STA-HDMP. However, Alg.~\ref{alg:1} requires iterating over the set $\mathbf{N}$ of possible combinations of tasks currently being processed. The size of this set grows rapidly with the number of task classes, prohibiting scalability to more than a few classes. Intuitively, we expect that some combinations of tasks result in a similar ``workload''. For example, processing two tasks with a medium service rate might represent a similar workload to processing one fast, and one slow task. If this were the case, we would expect that the optimal policy would apply a similar threshold to accepting new tasks in both cases. Therefore, we can expect to obtain a good solution by treating both of these cases as the same. This idea of grouping similar states together is referred to as~\emph{state abstraction}~\cite{li2006towards}. In this section, we show how to use state abstraction to improve the scalability of our approach.

We introduce the following notation following from~\citet{li2006towards}. Let $\mathcal{M} = (\mathbf{X}, A, R, P, D)$ be referred to as the~\emph{ground} MDP, with optimal policy $\pi^*$. The~\emph{abstract} version of $\mathcal{M}$ is $\widehat{\mathcal{M}}= (\widehat{\mathbf{X}}, A, \widehat{R}, \widehat{P},  D)$, and has optimal policy $\widehat{\pi}^*$. The abstraction function, $\psi : \mathbf{X} \rightarrow \widehat{\mathbf{X}}$, maps each ground state to its corresponding abstract state, and $\psi^{-1}(\widehat{\mathbf{x}})$ is the inverse of $\psi(\mathbf{x})$. The~\emph{weighting function} is $w: \mathbf{X} \rightarrow [0, 1]$, where for each $\widehat{\mathbf{x}} \in \widehat{\mathbf{X}}$, $\sum_{\mathbf{x} \in \psi^{-1}(\widehat{\mathbf{x}})} w(\mathbf{x})= 1$. The abstract reward and transition functions are a weighted sum over the corresponding functions for the ground states
\begin{equation}
    \widehat{R}(\widehat{\mathbf{x}}, a, t) = \sum_{\mathbf{x} \in \psi^{-1}(\widehat{\mathbf{x}})} w(\mathbf{x}) R(\mathbf{x}, a, t)
    \vspace{-1mm}
\end{equation}
\begin{equation}
    \label{eq:abstract_trans_probs}
    \widehat{P}(\widehat{\mathbf{x}}' \mid \widehat{\mathbf{x}}, a, t) =  \hspace{-3mm} \sum_{\mathbf{x} \in \psi^{-1}(\widehat{\mathbf{x}})} \sum_{\mathbf{x}' \in \psi^{-1}(\widehat{\mathbf{x}}')}  \hspace{-3mm} w(\mathbf{x}) P(\mathbf{x}' \mid \mathbf{x}, a, t)
    \raisetag{5mm}
\end{equation}

To generate the state aggregation function, one possibility is to consider aggregating states together when their optimal $Q$-values in the ground MDP are within $\epsilon$, i.e. \vspace{-1mm}
\begin{multline}
    \label{eq:aggregation_cond}
    \psi(\mathbf{x}_1) = \psi(\mathbf{x}_2) \rightarrow \\ \forall_{a \in A} \mid Q_{\pi^*} (\mathbf{x}_1, a, t) - Q_{\pi^*} (\mathbf{x}_2, a, t) \mid\  \leq \epsilon
\end{multline}

Let $\pi^{GA}$ denote the policy in the ground MDP derived from the optimal policy in the abstract MDP,
\begin{equation}
    \label{eq:abstract_to_ground}
    \pi^{GA}(\mathbf{x}) = \widehat{\pi}^*\big(\psi(\mathbf{x}) \big).
\end{equation}

\citet{abel2016near} prove that for any valid weighting function, if the state aggregation function satisfies Eq.~\ref{eq:aggregation_cond}, then the suboptimality of  $\pi^{GA}$ in the ground MDP is bounded by a function linear in $\epsilon$:
\begin{equation}
    \label{eq:abel_theorem}
    V_{\pi^*}(\mathbf{x}) - V_{\pi^{GA}}(\mathbf{x}) \leq  \mathcal{O}(\epsilon), \textnormal{ for all }\mathbf{x} \in \mathbf{X}
\end{equation}

\subsection{State Abstraction in the STA-HDMP}
We are now ready to present our state abstraction approach for approximating the optimal solution to the STA-HMDP.

\begin{restatable}{proposition}{abstractionerr}
    \label{prop:5}
    Consider two states, $\mathbf{x}_1$ and $\mathbf{x}_2$, in the STA-HDMP where $\mathbf{x}_1[k^+] = \mathbf{x}_2[k^+] $ and $\mathbf{x}_1[p] = \mathbf{x}_2[p].$ If ${\mid Q_{\pi^*}(\mathbf{x}_1[\mathbf{n}], \mathtt{rej}, t) - Q_{\pi^*}(\mathbf{x}_2[\mathbf{n}], \mathtt{rej}, t) \mid\ \leq \epsilon / 2}$, and $\ {\forall_{k^+}{\mid p^*_{cr}(\mathbf{x}_1[\mathbf{n}], k^+, t) -p^*_{cr}(\mathbf{x}_2[\mathbf{n}], k^+, t) \mid} \leq \epsilon/2 }$, then
    \end{restatable}
    \vspace{-4mm}
    \begin{equation}
\forall_{a \in A} \mid Q_{\pi^*} (\mathbf{x}_1, a, t) - Q_{\pi^*} (\mathbf{x}_2, a, t) \mid\ \leq \epsilon
    \end{equation}

Proposition~\ref{prop:5} suggests that we can define the state aggregation function as follows
\begin{multline}
    \label{eq:da_aggregation}
    \psi(\mathbf{x}_1) =  \psi(\mathbf{x}_2) \rightarrow \mathbf{x}_1[k^+] = \mathbf{x}_2[k^+],\  \mathbf{x}_1[p] = \mathbf{x}_2[p], \\
    Q_{\pi^*}(\mathbf{x}_1[\mathbf{n}], \mathtt{rej}, t) \approx Q_{\pi^*}(\mathbf{x}_2[\mathbf{n}], \mathtt{rej}, t) \textnormal{\hspace{3pt} and } \\ 
    p^*_{cr}(\mathbf{x}_1[\mathbf{n}], k^+, t) \approx p^*_{cr}(\mathbf{x}_2[\mathbf{n}], k^+, t) \textnormal{\hspace{3pt} for all } k^+,
\end{multline}
\noindent and the performance of the policy derived from the abstract MDP will be near-optimal due to the result stated in Eq.~\ref{eq:aggregation_cond}-\ref{eq:abel_theorem}.
The resulting abstract states are represented by the state variables $\{n_A,k^+, p\}$, where $n_A \in \mathcal{N}_A$ is an abstraction of $\mathbf{n}$, the combination of task classes being processed in the ground state. $\mathcal{N}_A$ is the set of abstractions of task class combinations. We write $\psi_\mathbf{n} : \mathbf{N} \rightarrow \mathcal{N}_A$ to denote the aggregation function for combinations of task classes such that
\begin{multline}
    \psi(\mathbf{x}_1) = \psi(\mathbf{x}_2) \Longleftrightarrow \mathbf{x}_1[k^+] = \mathbf{x}_2[k^+],\  \mathbf{x}_1[p] = \mathbf{x}_2[p], \\ \textnormal{ and \hspace{2pt}}
    \psi_\mathbf{n}(\mathbf{x}_1[\mathbf{n}]) = \psi_\mathbf{n}(\mathbf{x}_2[\mathbf{n}])
\end{multline}

We cannot directly apply Proposition~\ref{prop:5} to aggregate task combinations, as we cannot compute the optimal STA-HDMP values required. This is because solving for these values via Alg.~\ref{alg:1} does not scale for many task classes. Guided by the intuition obtained from Eq.~\ref{eq:da_aggregation} we propose two more scalable methods for computing the state aggregation function. 

\subsubsection{State Aggregation via Semi-MDP Stationary Solution}
As an efficient alternative to aggregating based on the optimal STA-HDMP solution as suggested by Eq.~\ref{eq:da_aggregation}, we propose to compute a stationary approximation of the STA-HMDP solution to use to determine the aggregation function. 

We first compute the average arrival rate over the horizon for each task class, $\mathbb{E}_t[\Lambda_k(t)]$. We then find the solution which optimises the average reward over an infinite horizon using the average arrival rates. This problem can be solved more quickly than the STA-HDMP as the optimal solution is stationary. We solve this stationary problem by adapting policy iteration for stationary Semi-MDPs~\cite{puterman2014markov}. Details of this are in the supp. mat. For each $\mathbf{n}\in \mathbf{N}$, we then compose a vector $v_\mathbf{n}$ of the associated~\emph{relative}\footnote{We use the relative value (see~\citet{puterman2014markov}, Chapter 8) as the value is infinite the average-reward infinite horizon setting.} $Q$-value and critical price values in the stationary solution:  $v_{\mathbf{n}}^{ss} = [ \widetilde{Q}_{\pi^*}(\mathbf{n}, \mathtt{rej}),\ \widetilde{p}^*_{cr}(\mathbf{n}, k_1), \ldots,\ \widetilde{p}^*_{cr}(\mathbf{n}, k_{|K|}) ]$, where we use ``$\sim$" to denote the solution for the stationary problem.  We then perform clustering on the $v_\mathbf{n}^{ss}$ vectors to aggregate the task class combinations into the desired number of abstractions, $|\mathcal{N}_A|$. These clusters define aggregation function, $\psi_\mathbf{n}$. For clustering, we use the k-means algorithm.

\subsubsection{State Aggregation via Order Statistics}
The scalability of the state aggregation approach we have just outlined is limited, as it requires solving for the stationary solution which may not be feasible for a large number of task classes. Here, we present an alternative state aggregation approach, based on summary statistics for each combination of tasks. For each $\mathbf{n} \in \mathbf{N}$, we compose a vector of the form 
$$v_{\mathbf{n}}^{os} = \\ \big[ \mathbb{E}[t_{N_{f} \geq 0} \mid \mathbf{n}], \mathbb{E}[t_{N_{f} \geq 1} \mid \mathbf{n}], \ldots, \mathbb{E}[t_{N_{f} = N_{serv}} \mid \mathbf{n}] \big]$$

\noindent where $N_{f}$ denotes the number of servers free, i.e. $N_f = N_{serv} - \sum_{k\in K} n_k$. Additionally, $\mathbb{E}[t_{N_{f} \geq q} \mid \mathbf{n}]$ denotes the expected time until at least $q$ servers are free given that the combination of tasks currently being processed is $\mathbf{n}$, and no further tasks are accepted. The intuition for $v_{\mathbf{n}}^{os}$ is that it approximately summarises the distribution over task completion times for the tasks currently being processed. Once the vectors have been computed, we perform clustering using k-means, and the resulting clusters correspond to $\psi_\mathbf{n}$.

Computing $v_{\mathbf{n}}^{os}$ requires computing the mean of order statistics of independent and non-identical exponential distributions. This is computationally challenging, so in practice we use a Monte Carlo approximation (details in supp. mat.).

\subsubsection{Value Iteration for STA-HMDP with State Abstraction}
We refer to the abstract version of the STA-HDMP as the Abstract STA-HDMP. 
We provide a brief summary of the algorithm for the Abstract STA-HDMP here, and more details can be found in the supp. mat.
We denote by  $\widehat{Q}_{\pi^*}$, and $\widehat{p}^*_{cr}$ the optimal $Q$-value function and critical price respectively in the Abstract STA-HDMP. Like the original STA-HMDP, the optimal policy for the Abstract STA-HMDP has a threshold form, as stated in Proposition~\ref{prop:6}.

\begin{restatable}{proposition}{abstractvalue}
\label{prop:6}
The optimal policy for the Abstract STA-HMDP, $\widehat{\pi}^*$, may be expressed as follows \vspace{-1mm}
\begin{equation*}
    \widehat{\pi}^*(\widehat{\mathbf{x}}, t) = 
    \begin{cases}
        \mathtt{acc}, \hspace{3pt} \textnormal{if } \widehat{\mathbf{x}}[{k^+}] \neq \bot \textnormal { and } \widehat{\mathbf{x}}[p] \geq \widehat{p}^*_{cr}(\widehat{\mathbf{x}}[n_A], \widehat{\mathbf{x}}[{k^+}], t) \\
        \mathtt{rej}, \hspace{3pt} \textnormal{otherwise} \vspace{-1mm}
    \end{cases}
\end{equation*}
\end{restatable}
\noindent \textit{where the critical price function is \vspace{-1mm}}
\begin{multline}
    \widehat{p}^*_{cr}(n_A, {k^+}, t) =  \frac{1}{\Pr(\mathtt{fin}({k^+}) \mid t) } \times 
    \Big(\widehat{Q}_{\pi^*}(n_A, \mathtt{rej}, t)  -   \\ \sum_{\mathbf{n} \in \psi_{\mathbf{n}}^{-1}(n_A)}  \hspace{-3mm} w(\mathbf{n}) \cdot \widehat{Q}_{\pi^*}(\psi(\mathbf{x}^{\mathtt{acc}}_{(\mathbf{n}, {k^+})}), \mathtt{rej}, t)
    \Big). 
\end{multline}
\noindent \textit{if $\mathbf{x}[{k^+}] \neq \bot$ and $t < t_H$, and 0 otherwise. }

The value iteration algorithm for solving the Abstract STA-HDMP proceeds in a similar manner to Alg.~\ref{alg:1}. Pseudocode is provided in Alg. 2 in the supp. mat. Once the policy has been computed for the Abstract STA-HDMP, the policy to apply to the original STA-HMDP is derived using Eq.~\ref{eq:abstract_to_ground}.

\begin{figure*}[bth]
    \centering
        \includegraphics[width=0.8\textwidth]{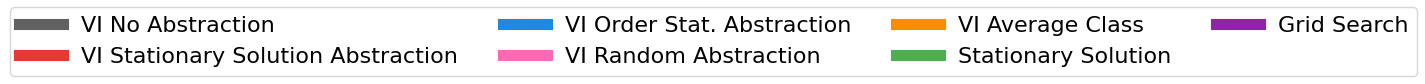}
    \vspace{-4mm}
\end{figure*}
\setlength{\dbltextfloatsep}{12pt}
\begin{figure*}
    \centering
    \begin{minipage}[t]{0.30\textwidth}
       \includegraphics[width=\textwidth]{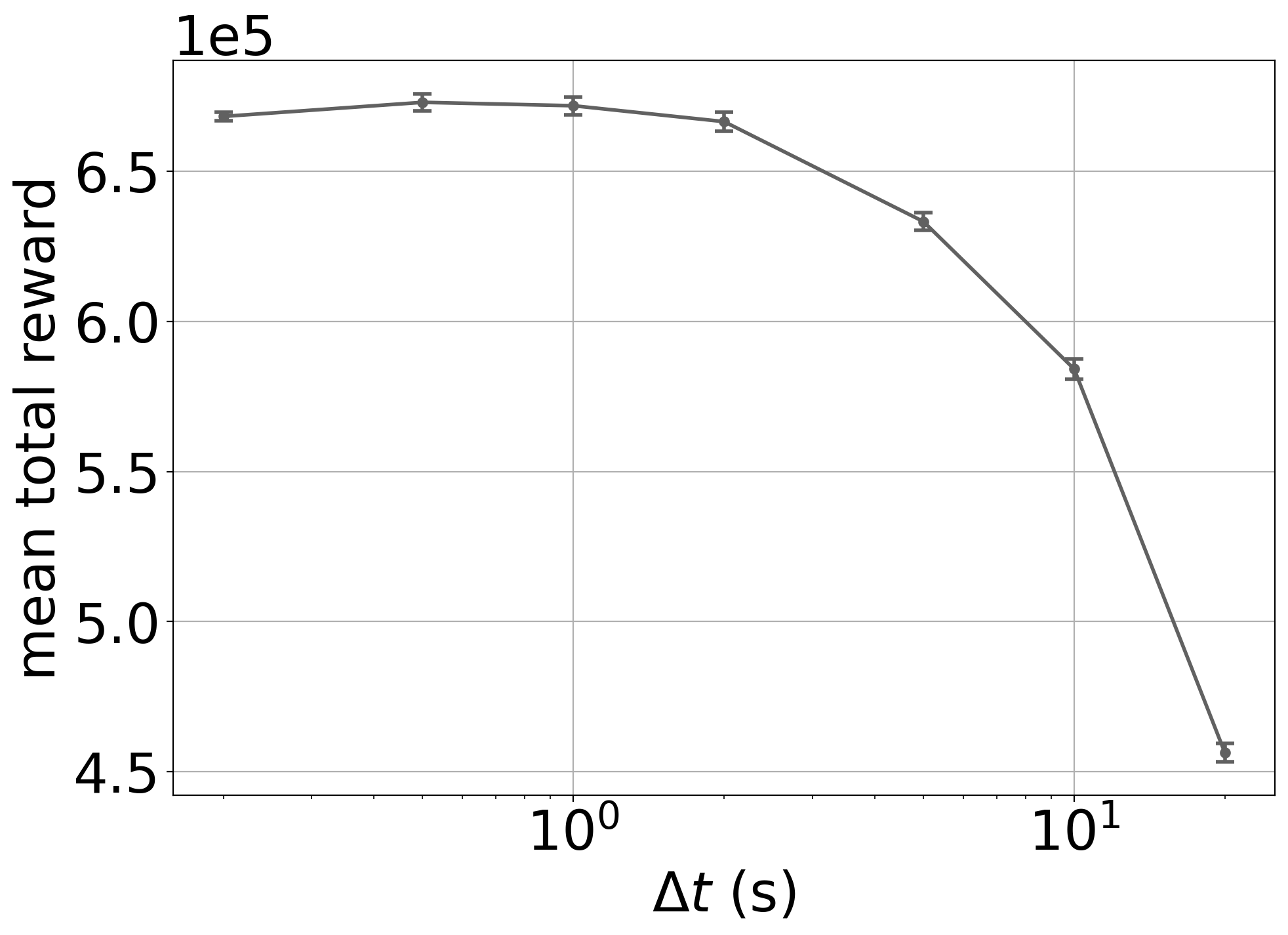}
    \caption{Mean reward vs $\Delta t$ for~\emph{VI No Abstractions} in the Synthetic Small domain (sinusoid arrival rates).\label{fig:dt_performance}}
    \end{minipage}
    \hspace{3pt}
    \begin{minipage}[t]{0.32\textwidth}
        \centering
        \includegraphics[width=\textwidth]{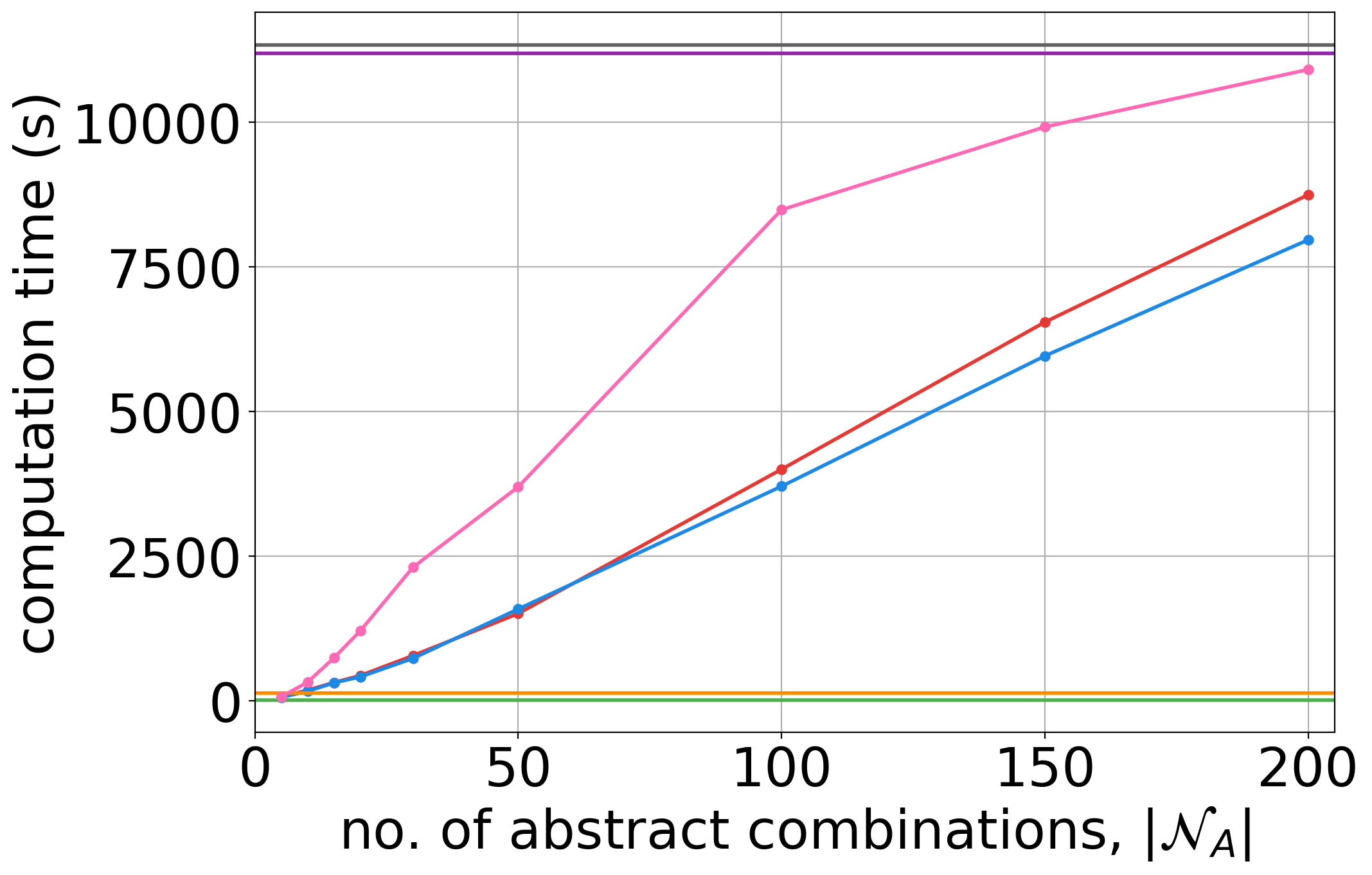}
        \caption{ Computation times for each method in the Synthetic Small domain (sinusoid arrival rates).}
        \label{fig:comp_time}
    \end{minipage}
    \hspace{5pt}
    \begin{minipage}[t]{0.31\textwidth}
        \centering
        \includegraphics[width=\textwidth]{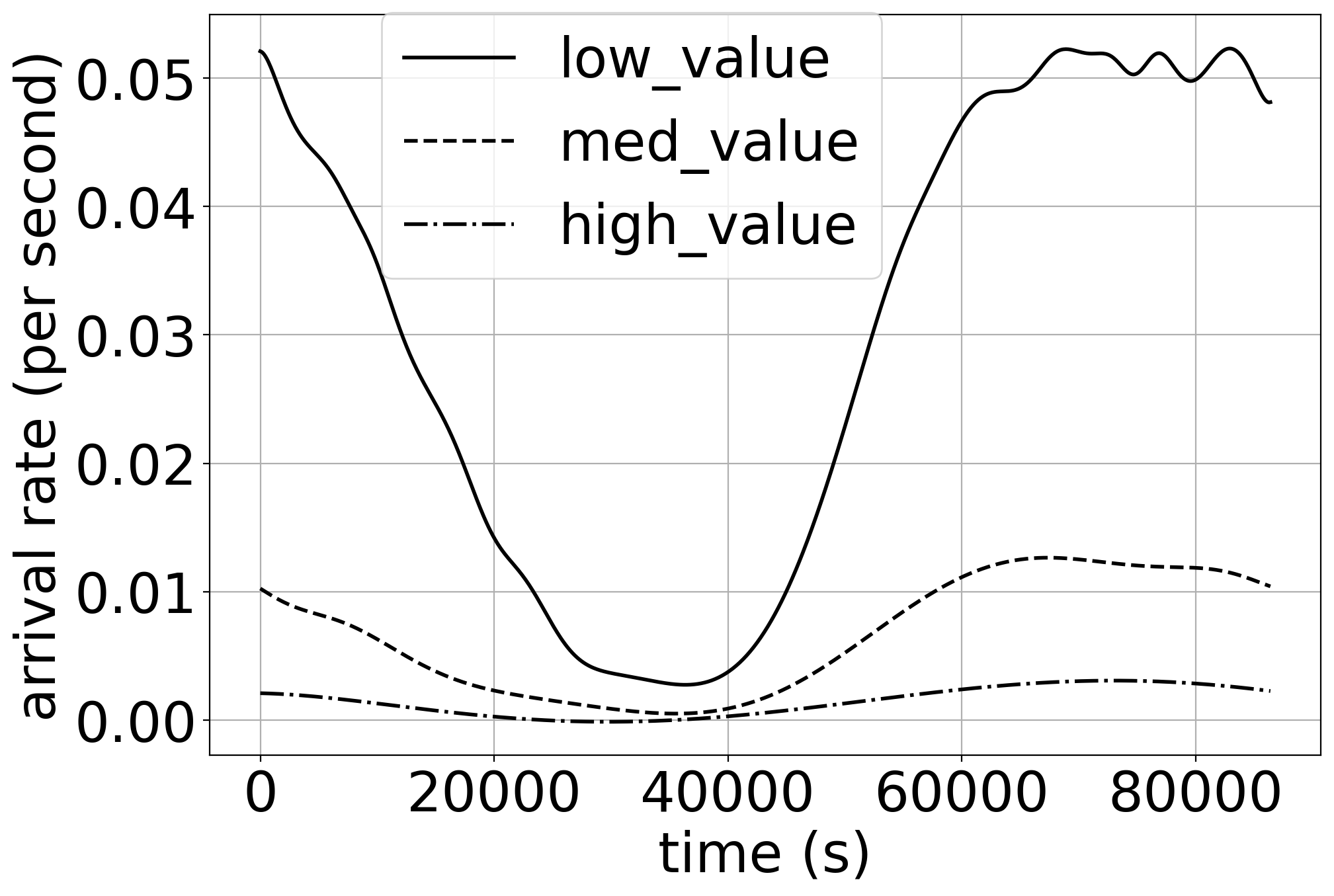}
        \caption{Arrival rate functions for the public fraud dataset.}
        \label{fig:ieee_arrival_rates}
    \end{minipage}
    \vspace{-3mm}
\end{figure*}

\begin{figure*}[t!]
    \centering
    \begin{subfigure}[t]{0.31\textwidth}
        \centering
        \includegraphics[width=\textwidth]{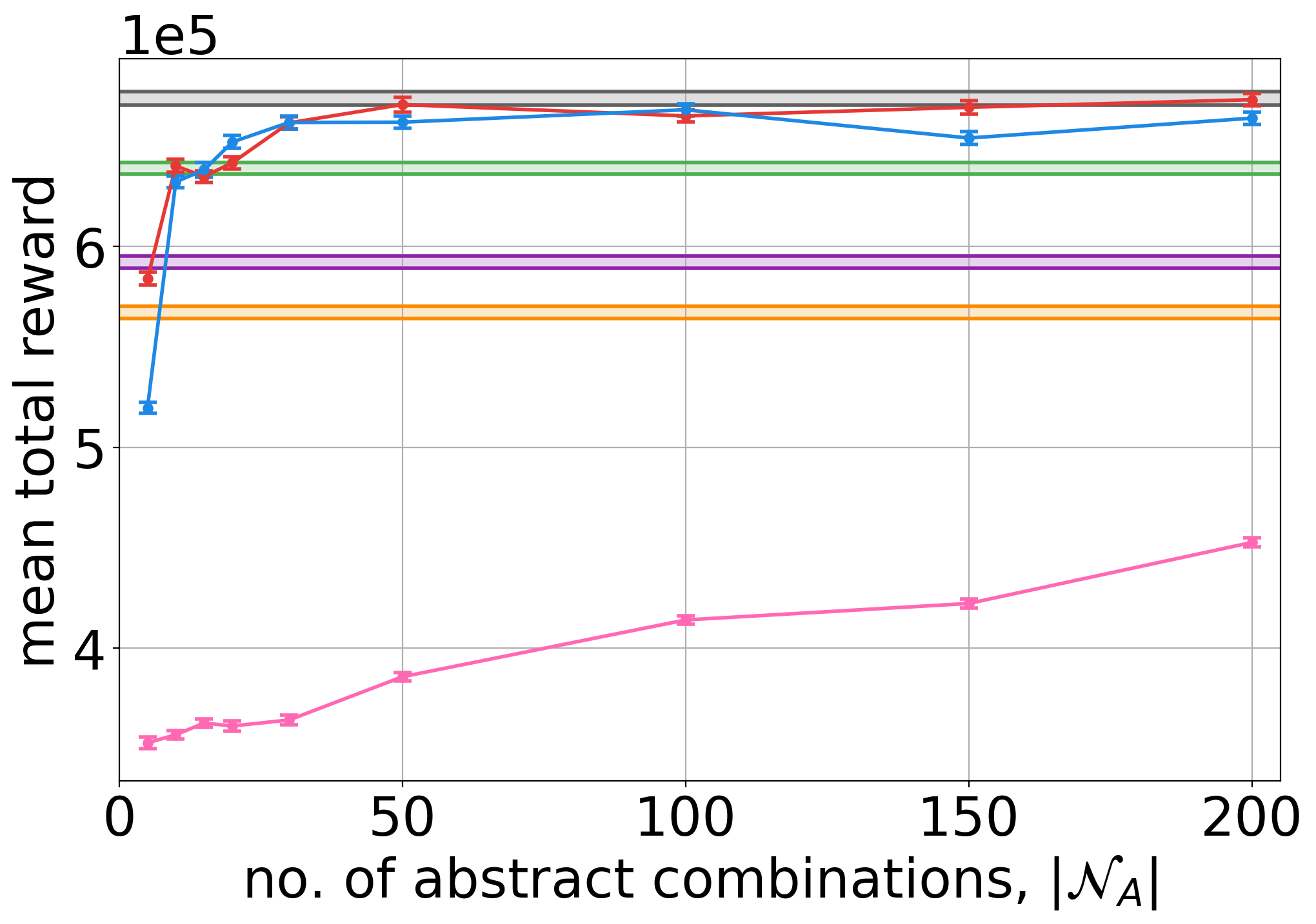}
        \caption{Synthetic Small (sinusoid arrival rates). \label{fig:synthetic_small_sinusoid}}
    \end{subfigure}%
    ~
    \begin{subfigure}[t]{0.32\textwidth}
        \centering
        \includegraphics[width=\textwidth]{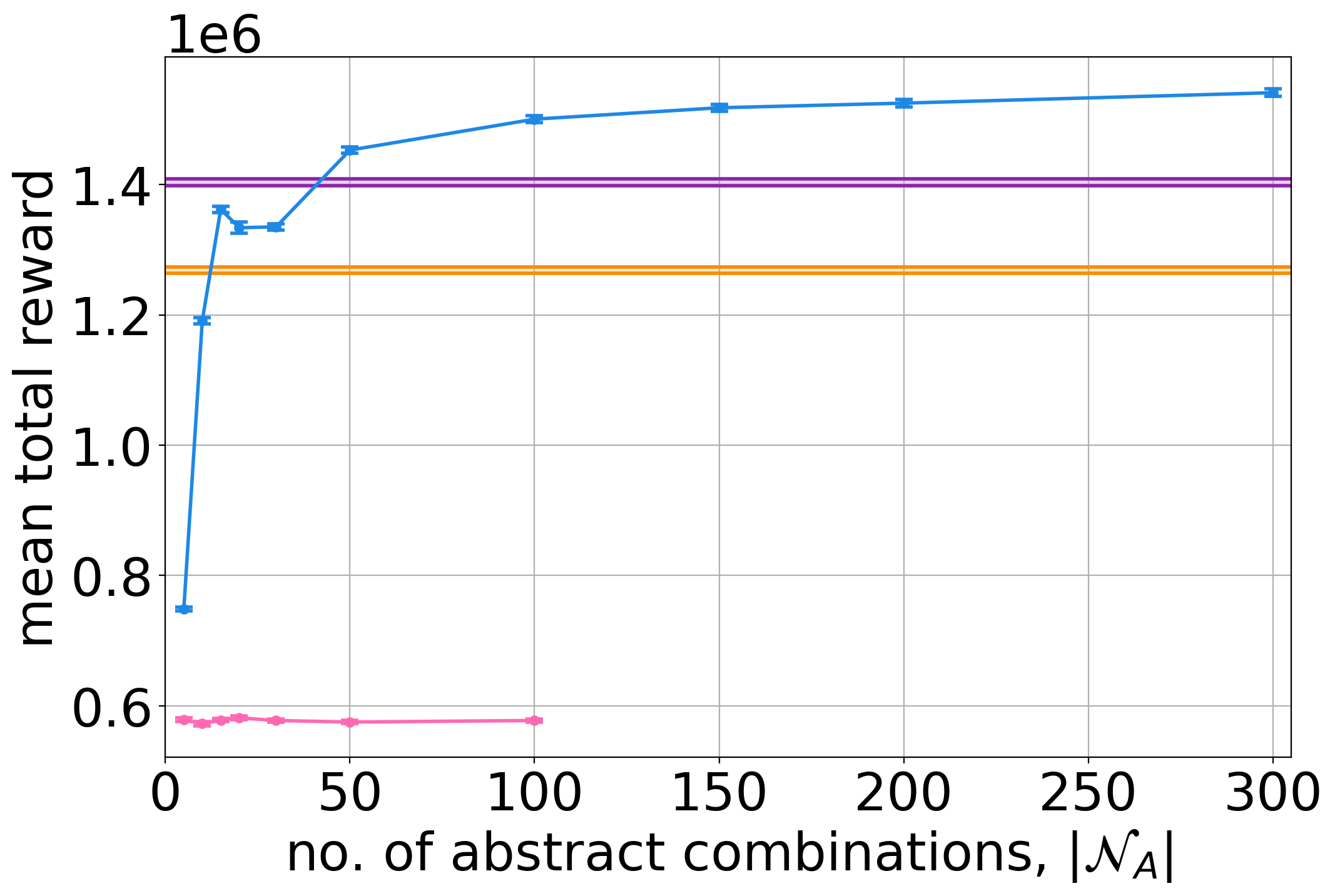}
        \caption{Synthetic Large (sinusoid arrival rates). \label{fig:synthetic_large_sinusoid}}
    \end{subfigure}
    ~
    \begin{subfigure}[t]{0.33\textwidth}
        \centering
        \includegraphics[width=\textwidth]{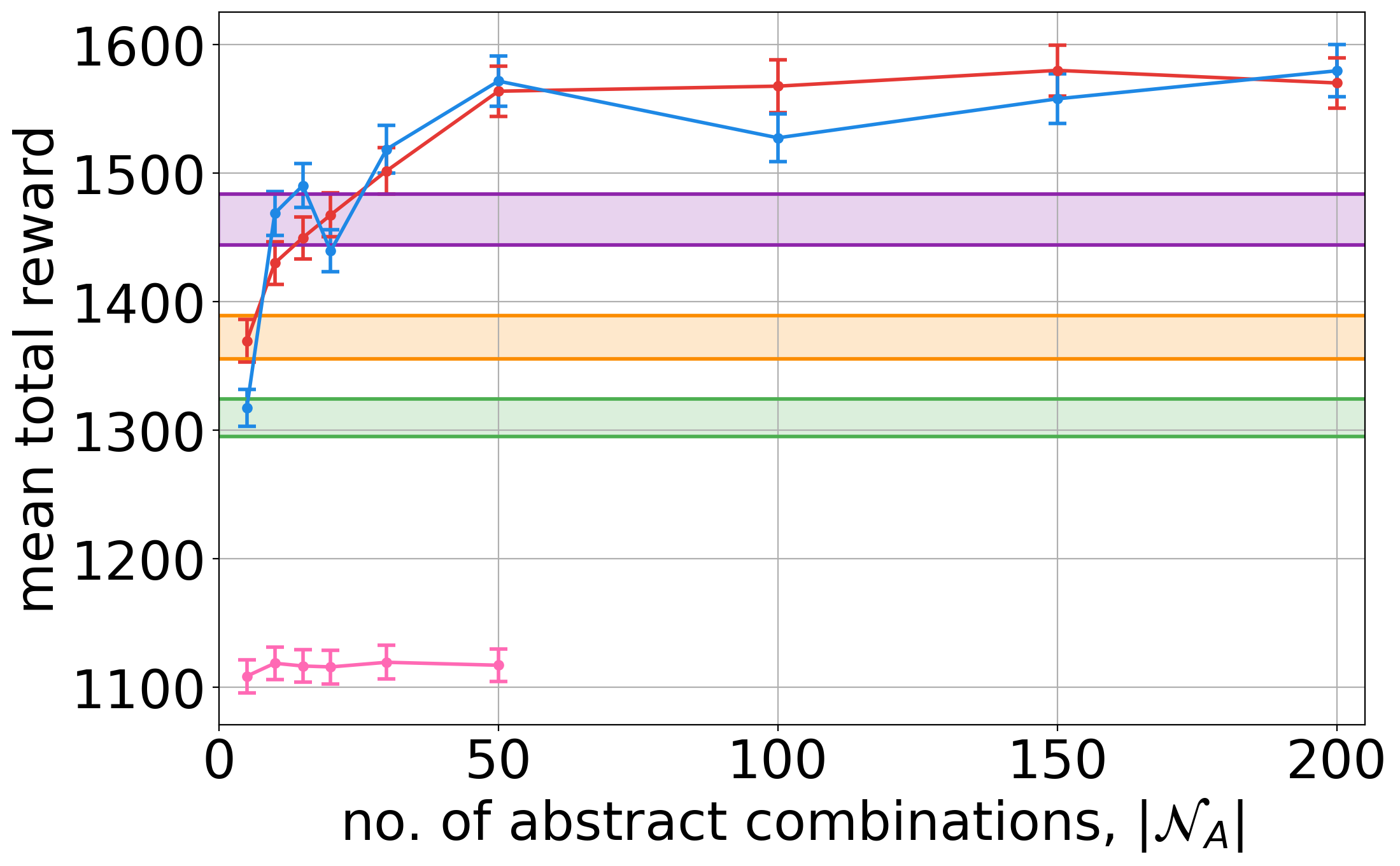}
        \caption{Public Fraud dataset. \label{fig:ieee_performance}}
    \end{subfigure}
    \caption{Mean total reward performance on 300 evaluation runs for each domain. Error bars and shaded regions indicate standard errors. We only include methods and numbers of abstract combinations which can be computed within 30,000 seconds. }
\end{figure*}

\section{Experiments}
The code used to run the experiments is implemented in Python and is included in the supp. mat. The experiments were run using an Intel Xeon 
E3-1585L v5 3GHz processor with 64GB of RAM. We compare several methods based on our Value Iteration (VI) approach as well as additional baselines. The following methods are compared:
\begin{itemize}[leftmargin=*, noitemsep, itemindent=5pt]
    \item \textit{VI No Abstr.}: our VI approach from Alg.~\ref{alg:1}.
    \item \textit{VI Stationary Sol. Abstr.}: our VI approach with state abstraction based on the Semi-MDP stationary solution.
   \item \textit{VI Order Stat. Abstr.}: our VI approach with state abstraction based on the mean of order statistics.
   \item \textit{VI Random Abstr.}: our VI approach with state abstraction with random state aggregation.
   \item \textit{VI Avg. Class}: we compute an ``average" task class, $k_{\mathtt{avg}}$, as follows. The arrival rate function is $\Lambda_{k_{\mathtt{avg}}}(t) = \sum_{k \in K} \Lambda_k (t)$. The price distribution and service rate are a weighted average over each task. The weighting for each class, $\omega_k$, is proportional to the mean arrival rate for each class: $\omega_k \propto \mathbb{E}_t[\Lambda_k(t)].$ We perform VI using the single average class, and assume that all tasks are the average class.
   \item \textit{Stationary Sol.}: uses the  critical prices from the stationary solution, which assumes constant arrival rates.
   \item \textit{Grid Search}: the critical price for each task class, $p_k$, is proportional to the expected service time, i.e. $p_k = C/\mu_k$, where $C$ is a constant. To find the best value for $C$, we evaluate the performance of 50 log-spaced values of $C$ for 300 episodes each, and choose the best value.
\end{itemize}

\subsubsection{Synthetic Domains}
\vspace{-2mm} For each synthetic domain, the horizon is 28800s, representing an 8hr working day. The task prices are distributed according to a Lomax distribution, with shape parameter of 3. We test two arrival rate functions:~\textit{sinusoid} and~\textit{step} functions which are plotted in the supp. mat.

\noindent \textit{Small}: There are 10 servers and 3 task classes which we call $\{ \mathtt{slow},\ \mathtt{medium},\ \mathtt{fast} \}$. The scale parameters of the Lomax price distribution for each task class are: $\{ \mathtt{slow} \textnormal{=} 1600,\ \mathtt{medium} \textnormal{=} 900,\ \mathtt{fast} \textnormal{=} 400 \}$. The service rates for each task class are: $\{ \mathtt{slow} \textnormal{=} \frac{1}{2000},\ \mathtt{medium} \textnormal{=} \frac{1}{1000},\ \mathtt{fast} \textnormal{=} \frac{1}{500} \}$. 

\noindent \textit{Large}: There are 40 servers and 4 task classes which we call $\{ \mathtt{very\ slow},\ \mathtt{slow},\ \mathtt{medium},\ \mathtt{fast} \}$. The Lomax price distribution scale parameters are: $\{\mathtt{very\ slow} \textnormal{=} 2500,\ \mathtt{slow} \textnormal{=} 1600,\ \mathtt{medium} \textnormal{=} 900,\ \mathtt{fast} \textnormal{=} 400 \}$. The service rates for each of the task classes are: ${\{ \mathtt{very\ slow} \textnormal{=} \frac{1}{4500}, \mathtt{slow} \textnormal{=} \frac{1}{3000},\ \mathtt{medium} \textnormal{=} \frac{1}{1500},\ \mathtt{fast} \textnormal{=} \frac{1}{750}\}}$.

For both versions of the synthetic domains, tasks with slower service times tend to have higher prices. 

\subsubsection{Public Fraud Data}
For this domain we use a public dataset, $M$, of financial transactions which are labelled as fraudulent or non-fraudulent~\cite{ieee2019}. We wish to optimise the expected value of fraudulent transactions reviewed by staff. The horizon for each episode is 86400s (1 day). We use half of the dataset, $M_{train}$, to train a classifier to predict the probability that a transaction is fraudulent based on additional features of the data. We assume that transactions with a probability $\geq 0.1$  are automatically reviewed. Our objective is to optimise the value of the remaining transactions which are reviewed by staff. To adjust for the fraud likelihood, we consider the~\emph{adjusted price} of each task (transaction) to be the value of the transaction multiplied by the probability of fraud predicted by the classifier. We assume that tasks are divided into the following classes according to the value of the transactions: $\{\mathtt{low\_val}$: $[\$0, \$150),\ \mathtt{med\_val}$: $[\$150, \$ 500),\ \mathtt{high\_val}$: $[\$500, \infty ) \}$. We assume that there are 20 staff (servers), and that transactions of different values require different processes to review, and therefore have the following service rates: $\{\mathtt{low\_val}\textnormal{=} \frac{1}{900},\ \mathtt{med\_val} 
\textnormal{=} \frac{1}{3600},\ \mathtt{high\_val}\textnormal{=} \frac{1}{10,800} \}$. We learn the adjusted price distributions and arrival rate functions from $M_{train}$ with the method from~\citet{dervovic2021non}. We then simulate the total adjusted price of transactions reviewed using the remainder of the data, $M_{test}$.

\subsection{Results}
We evaluate each approach for 300 episodes on each domain. In the plots, the error bars and the shaded regions indicate the mean $\pm$ the std. error. Additional results are in the supp. mat.

\subsubsection{Influence of time step} Figure~\ref{fig:dt_performance} shows the influence of $\Delta t$ for the~\emph{VI No Abstr.} method in the Synthetic Small domain with sinusoid arrival rate functions. At large values of $\Delta t$, the performance is poor, but as $\Delta t$ is decreased below approximately 1 second, the performance plateaus. This empirically confirms Proposition~\ref{eq:approx_error}. Based on this finding, we use a time step of $\Delta t = 0.5$s for all other results on the synthetic domains. For the fraud domain we use $\Delta t = 1$s as this is the resolution of the time information in the dataset.

\subsubsection{Computation Time} The computation times for each approach on the Synthetic Small domain are shown in Figure~\ref{fig:comp_time}.~\emph{VI No Abstr.} and~\emph{Grid Search} are the slowest methods. For the methods based on state abstraction, the computation time decreases as the number of abstractions is decreased, verifying that our abstraction method reduces the computation required. We omit results for methods which take more than 30,000s. This means we do not include results for~\emph{VI No Abstr.} on either of the larger domains. Computation times for the other domains are included in the supp. mat.

\subsubsection{Synthetic Domain Results} Results on the Synthetic Small domain with sinusoid arrival rates are in Figure~\ref{fig:synthetic_small_sinusoid}.~\emph{VI No Abstr.} has the strongest performance, and the best baseline is \textit{Stationary Sol}. Both \textit{VI Order Stat. Abstr.} and \textit{VI Stationary Sol. Abstr.}  perform better than all baselines, and comparably to \emph{VI No Abstr.} if a sufficient number of abstractions (${\geq 30}$) are used.~\textit{VI Random Abstr.} performs poorly across all domains, indicating that the state aggregation approaches we have proposed are crucial for strong performance. 

The results on the Synthetic Large domain in Figure~\ref{fig:synthetic_large_sinusoid} show that \textit{VI Order Stat. Abstr.} is the best performing method, provided that a sufficient number of state abstractions are used. For the large domain, the number of possible combinations of task classes is over  130,000. This means that we cannot provide results for~\emph{VI No Abstr.}, or any of the methods which require the stationary solution because these methods (considerably) exceed our 30,000s computation limit.

In the supp. mat., we include results for the synthetic domains using step functions for the arrival rates. We observe similar results for the step functions, indicating that our method performs well for a variety of arrival rate functions.

\subsubsection{Public Fraud Data Results}
For the fraud domain, the arrival rates are plotted in Figure~\ref{fig:ieee_arrival_rates}, and the performance of each method is in Figure~\ref{fig:ieee_performance}. Our methods based on state abstraction outperform all of the baselines when using $\geq 30$ state abstractions, and require less computation time than the grid search approach as shown in the supp. mat. 

Our results validate that a) our state abstraction approach reduces the computation requirements for value iteration, making it applicable to complex problems, and b) our approach significantly improves performance over simpler baselines, and therefore has the potential to improve the efficiency of  real world systems, such as those used to monitor fraud.

\FloatBarrier

\section{Conclusion}
We have introduced a novel queuing problem that corresponds to a challenge faced in financial fraud detection and monitoring.
Moreover, we demonstrate that it has a computationally tractable solution by discretising time and utilising state abstraction. 
Future work includes adversarial modelling of incoming tasks and proving reward lower bounds as a function of the size of the abstract state space.

\newpage
\paragraph{Disclaimer}
This paper was prepared for informational purposes in part by
the Artificial Intelligence Research group of JPMorgan Chase \& Co. and its affiliates (``JP Morgan''),
and is not a product of the Research Department of JP Morgan.
JP Morgan makes no representation and warranty whatsoever and disclaims all liability,
for the completeness, accuracy or reliability of the information contained herein.
This document is not intended as investment research or investment advice, or a recommendation,
offer or solicitation for the purchase or sale of any security, financial instrument, financial product or service,
or to be used in any way for evaluating the merits of participating in any transaction,
and shall not constitute a solicitation under any jurisdiction or to any person,
if such solicitation under such jurisdiction or to such person would be unlawful.
\bibliography{aaai22}

\newpage
\onecolumn

\section{Proof of Proposition~\ref{prop:pwl_vf}}
\pwlvf*
\vspace{-4mm}
\begin{equation*}
    V_{\pi^*}(\mathbf{x}, t) = 
    \begin{cases}
        \Pr\big( \mathtt{fin}(\mathbf{x}[{k^+}]) \mid t \big)  \cdot \big(\mathbf{x}[p] - p^*_{cr}(\mathbf{x}[\mathbf{n}], \mathbf{x}[{k^+}], t) \big)  
         + Q_{\pi^*}(\mathbf{x}[\mathbf{n}], \mathtt{rej}, t), \\
         \hspace{200pt} \textnormal{if } \mathbf{x}[{k^+}] \neq \bot \textnormal{ and } \mathbf{x}[p] >= p^*_{cr}(\mathbf{x}[\mathbf{n}], \mathbf{x}[{k^+}], t). \vspace{2mm}\\
        Q_{\pi^*}(\mathbf{x}[\mathbf{n}], \mathtt{rej}, t), \textnormal{ otherwise.}
    \end{cases}
\end{equation*}

\noindent where $p^*_{cr}(\mathbf{x}[\mathbf{n}], \mathbf{x}[{k^+}], t)$ is the \emph{critical price function}, defined as
\begin{equation*}
    p^*_{cr}(\mathbf{x}[\mathbf{n}], \mathbf{x}[{k^+}], t) =  \frac{1}{\Pr(\mathtt{fin}(\mathbf{x}[{k^+}]) \mid t) } \times
    \Big(Q_{\pi^*}(\mathbf{x}[\mathbf{n}], \mathtt{rej}, t) -  Q_{\pi^*}(\mathbf{x}^{\mathtt{acc}}_{(\mathbf{x}[\mathbf{n}], \mathbf{x}[k^+])}[\mathbf{n}], \mathtt{rej}, t)
    \Big).
\end{equation*}

\noindent if $\mathbf{x}[{k^+}] \neq \bot$ and $t < t_H$. We define $p^*_{cr}(\mathbf{x}[\mathbf{n}], \mathbf{x}[{k^+}], t) = 0$ if $\mathbf{x}[{k^+}] = \bot$ or $t = t_H$.

\bigskip
\textit{Proof}: 
First we address the case where $t = t_H$. $\Pr(\mathtt{fin}(\mathbf{x}[{k^+}]) \mid t_H) = Q_{\pi^*}(\mathbf{x}[\mathbf{n}], \mathtt{rej}, t) = 0$, yielding $V_{\pi^*}(\mathbf{x}, t_H) = 0$ as required.

Next we address the case where $\mathbf{x}[{k^+}] = \bot$. In this case, the only action available is $\mathtt{rej}$. Therefore, $V_{\pi^*}(\mathbf{x}, t) = Q_{\pi^*}(\mathbf{x}[\mathbf{n}], \mathtt{rej}, t)$.

We now address the case where $t < t_H$ and $\mathbf{x}[{k^+}] \neq \bot$. Because a task is available, both the $\mathtt{acc}$ and $\mathtt{rej}$ actions are available to the decision maker. From Equation~\ref{eq:qv_accept_act}, the expression for the  optimal $Q$-value for the accept action is
\begin{multline}
\label{eq:q_val_accept}
    Q_{\pi^*}(\mathbf{x}, \mathtt{acc}, t) = \mathbf{x}[p] \cdot \Pr(\mathtt{fin}(\mathbf{x}[{k^+}] ) \mid t) + V_{\pi^*}(\mathbf{x}^{\mathtt{acc}}_{(\mathbf{x}[\mathbf{n}], \mathbf{x}[k^+])}, t) \\
    = \mathbf{x}[p] \cdot \Pr(\mathtt{fin}(\mathbf{x}[{k^+}]) \mid t) + Q_{\pi^*}(\mathbf{x}^{\mathtt{acc}}_{(\mathbf{x}[\mathbf{n}], \mathbf{x}[k^+])}[\mathbf{n}], \mathtt{rej}, t)
\end{multline}
\noindent where the second equality is because $\mathbf{x}^{\mathtt{acc}}_{(\mathbf{x}[\mathbf{n}], \mathbf{x}[k^+])}[{k^+}] = \bot$, so only the reject action is enabled at $\mathbf{x}^{\mathtt{acc}}_{(\mathbf{x}[\mathbf{n}], \mathbf{x}[k^+])}$.


By adding and subtracting $p^*_{cr}(\mathbf{x}[\mathbf{n}], \mathbf{x}[{k^+}], t)$, in Eq.~\ref{eq:q_val_accept} we may express $Q_{\pi^*}(\mathbf{x}, \mathtt{acc}, t)$ as follows
\begin{equation}
    \label{eq:expr}
    Q_{\pi^*}(\mathbf{x}, \mathtt{acc}, t) = \Pr\big(\mathtt{fin}(\mathbf{x}[{k^+}]\big) \mid t)  \times 
    \Big(\mathbf{x}[p] - p^*_{cr}(\mathbf{x}[\mathbf{n}], \mathbf{x}[{k^+}], t) + p^*_{cr}(\mathbf{x}[\mathbf{n}], \mathbf{x}[{k^+}], t) \Big) \\ + Q_{\pi^*}(\mathbf{x}^{\mathtt{acc}}_{(\mathbf{x}[\mathbf{n}], \mathbf{x}[k^+])}[\mathbf{n}], \mathtt{rej}, t)
\end{equation}
    
By substituting the definition for $p^*_{cr}$,  Eq.~\ref{eq:expr} simplifies to
\begin{equation}
    \label{eq:q_val_relation}
    Q_{\pi^*}(\mathbf{x}, \mathtt{acc}, t_i) = \\ \Pr\big(\mathtt{fin}(\mathbf{x}[{k^+}]) \mid t \big) \cdot \big(\mathbf{x}[p] - p^*_{cr}(\mathbf{x}[\mathbf{n}], \mathbf{x}[{k^+}], t) \big)  \\
     + Q_{\pi^*}(\mathbf{x}[\mathbf{n}], \mathtt{rej}, t)
\end{equation}

The optimal value function is 
\begin{equation}
    V_{\pi^*}(\mathbf{x}, t) = \max_{a \in \{\mathtt{acc}, \mathtt{rej}\} } Q^*(\mathbf{x}, a, t)
\end{equation}

Because for $t < t_H$, $\Pr\big(\mathtt{fin}(\mathbf{x}[{k^+}]) \mid t \big) > 0$, from Eq.~\ref{eq:q_val_relation} we have that
\begin{equation}
    \label{eq:iff}
    Q_{\pi^*}(\mathbf{x}, \mathtt{acc}, t_i) \geq Q_{\pi^*}(\mathbf{x}[\mathbf{n}], \mathtt{rej}, t) \iff \\
    \mathbf{x}[p] \geq p^*_{cr}(\mathbf{x}[\mathbf{n}], \mathbf{x}[{k^+}], t) 
\end{equation}

Therefore, we can expression the optimal value function as 

\begin{equation}
    V_{\pi^*}(\mathbf{x}, t) =
    \begin{cases}
        Q_{\pi^*}(\mathbf{x}, \mathtt{acc}, t_i), \textnormal{\ \    if } \mathbf{x}[k^+] \neq \bot \textnormal{ and } \mathbf{x}[p] \geq p^*_{cr}(\mathbf{x}[\mathbf{n}], \mathbf{x}[{k^+}], t). \\
        Q_{\pi^*}(\mathbf{x}[\mathbf{n}], \mathtt{rej}, t), \textnormal{ otherwise. }
    \end{cases}
\end{equation}

This combined with Eq.~\ref{eq:q_val_relation} completes the proof. $\square$

\section{Proof of Proposition~\ref{prop:thr_pol}}

\thrpol*

\noindent \textit{Proof}: This follows directly from the fact that 
\begin{equation}
    \pi^*(\mathbf{x}, t) = \argmax_{a \in \{\mathtt{acc}, \mathtt{rej}\} } Q^*(\mathbf{x}, a, t),
\end{equation}
and Eq.~\ref{eq:iff}.

\section{Proof of Proposition~\ref{prop:simple_calcs}}

\msf*
\vspace{-4mm}
\begin{multline*}
    Q_{\pi^*}(\mathbf{x}[\mathbf{n}], \mathtt{rej}, t_i) =  
    \sum_{\mathbf{x}'[\mathbf{n}]} \prod_{k=1}^K  \Pr(\mathbf{x}'[n_k] \mid  \mathbf{x}[n_k]) \cdot  \sum_{\mathbf{x}'[{k^+}]} \Pr(\mathbf{x}'[{k^+}]= {k^+}' \mid t_i) \cdot \\
    \Big[ Q_{\pi^*}(\mathbf{x}'[\mathbf{n}], \mathtt{rej}, t_{i+1} ) + \Pr\big(\mathtt{fin}(\mathbf{x}'[{k^+}])  \mid t_{i+1} \big) \cdot \phi_{\mathbf{x}'[{k^+}]}(p^*_{cr}(\mathbf{x}'[\mathbf{n}], \mathbf{x}'[{k^+}], t_{i+1})) \Big]
\end{multline*}

\noindent where $\phi_{{k^+}}(p)$ is the \emph{mean shortage function} of the price distribution for task class ${k^+}$ defined as 

\begin{equation*}
    \phi_{{k^+}}(p) = \int_p^\infty (1 - {F_{{k^+}}}(y)) dy.
\end{equation*}

\bigskip
\noindent \textit{Proof}: We begin by substituting the transition function from Equation~\ref{eq:indep_trans_func} into Equation~\ref{eq:q_reject}

\begin{multline}
    \label{eq:eq_with_integral}
    Q_{\pi^*}(\mathbf{x}[\mathbf{n}], \mathtt{rej}, t_i) =
    \sum_{\mathbf{x}'[\mathbf{n}]}  \sum_{\mathbf{x}'[{k^+}]} \int_{\mathbf{x}'[p]} P(\mathbf{x'} \mid \mathbf{x}, \mathtt{rej}, t_i) \cdot V_{\pi^*}(\mathbf{x}', t_{i+1}) \cdot \dif(\mathbf{x}'[p]) = 
    \\ 
     \sum_{\mathbf{x}'[\mathbf{n}]}  \sum_{\mathbf{x}'[{k^+}]} \int_{\mathbf{x}'[p]} \Big[  \prod_{k=1}^K  \big(\Pr(\mathbf{x}'[n_k] \mid \mathbf{x}[n_k]) \big) \cdot \Pr(\mathbf{x}'[{k^+}] \mid t_i) \cdot f_{\mathbf{x}'[{k^+}]}(\mathbf{x}'[p])
    \cdot V_{\pi^*}(\mathbf{x}', t_{i+1}) \Big] \cdot \dif (\mathbf{x}'[p]) = \\
    \sum_{\mathbf{x}'[\mathbf{n}]}  \prod_{k=1}^K  \Pr(\mathbf{x}'[n_k] \mid  \mathbf{x}[n_k]) \sum_{\mathbf{x}'[{k^+}]} \Pr(\mathbf{x}'[{k^+}] \mid t_i) 
    \cdot \int_{\mathbf{x}'[p]} \Big[ f_{\mathbf{x}'[{k^+}]}(\mathbf{x}'[p]) \cdot V_{\pi^*}(\mathbf{x}', t_{i+1}) \Big] \cdot \dif (\mathbf{x}'[p])
\end{multline}

We split the integral in Eq.~\ref{eq:eq_with_integral} into two parts using the piecewise-linear value function representation in Proposition~\ref{prop:pwl_vf}
\begin{multline}
    \label{eq:with_int_2}
    \int_{\mathbf{x}'[p]} \Big[ f_{\mathbf{x}'[{k^+}]}(\mathbf{x}'[p]) \cdot V_{\pi^*}(\mathbf{x}', t_{i+1}) \Big] \cdot \dif (\mathbf{x}'[p]) =  \\
    \int_0^{p^*_{cr}(\mathbf{x}'[\mathbf{n}], \mathbf{x}'[{k^+}], t_{i+1})} \Big[ f_{\mathbf{x}'[{k^+}]}(\mathbf{x}'[p]) \cdot Q_{\pi^*}(\mathbf{x}'[\mathbf{n}], \mathtt{rej}, t_{i+1} ) \Big] \cdot \dif (\mathbf{x}'[p])  +  \\
    \int_{p^*_{cr}(\mathbf{x}'[\mathbf{n}], \mathbf{x}'[{k^+}], t_{i+1})}^\infty  f_{\mathbf{x}'[{k^+}]}(\mathbf{x}'[p]) \cdot \Big[  \Pr\big(\mathtt{fin}(\mathbf{x}'[{k^+}]) \mid t_{i+1} \big) \cdot \big(\mathbf{x}'[p] - p^*_{cr}(\mathbf{x}'[\mathbf{n}], \mathbf{x}'[{k^+}], t_{i+1}) \big)  + Q_{\pi^*}(\mathbf{x}'[\mathbf{n}], \mathtt{rej}, t_{i+1})  \Big] \cdot \dif (\mathbf{x}'[p]) \\
    = Q_{\pi^*}(\mathbf{x}'[\mathbf{n}], \mathtt{rej}, t_{i+1}) + \Pr\big(\mathtt{fin}(\mathbf{x}'[{k^+}]) \mid t_{i+1} \big) \cdot 
    \int_{p^*_{cr}(\mathbf{x}'[\mathbf{n}], \mathbf{x}'[{k^+}], t_{i+1})}^\infty f_{\mathbf{x}'[{k^+}]}(\mathbf{x}'[p]) \cdot \big(\mathbf{x}'[p] - p^*_{cr}(\mathbf{x}'[\mathbf{n}], \mathbf{x}'[{k^+}], t_{i+1}) \big) \cdot  \dif (\mathbf{x}'[p])
\end{multline}

The integral in Eq.~\ref{eq:with_int_2} is in the form $\int_p^\infty f_{{k^+}}(y) \cdot (y - p)  \dif y $, which is equal to the mean shortage function, i.e. 
\begin{equation}
    \label{eq:msf}
    \int_p^\infty f_{{k^+}}(y) \cdot (y - p)  \dif y  = \int_p^\infty (1 - F_{{k^+}}(y)) \dif y = \phi_{{k^+}}(p).
\end{equation}

A proof of Eq.~\ref{eq:msf} can be found in~\citet{dervovic2021non} (Lemma 1). 
Eq.~\ref{eq:msf} allows us to rewrite Eq.~\ref{eq:with_int_2} as 
\begin{multline}
    \label{eq:simplify_eq}
    \int_{\mathbf{x}'[p]} \Big[ f_{\mathbf{x}'[{k^+}]}(\mathbf{x}'[p]) \cdot V_{\pi^*}(\mathbf{x}', t_{i+1}) \Big] \cdot d (\mathbf{x}'[p]) = \\ 
    \Big[ Q_{\pi^*}(\mathbf{x}'[\mathbf{n}], \mathtt{rej}, t_{i+1} ) +  \Pr\big(\mathtt{fin}(\mathbf{x}'[{k^+}]) \mid t_{i+1} \big) \cdot \phi_{\mathbf{x}'[{k^+}]}(p^*_{cr}(\mathbf{x}'[\mathbf{n}], \mathbf{x}'[{k^+}], t_{i+1})) \Big]
\end{multline}

Substituting this expression into Eq.~\ref{eq:eq_with_integral} completes the proof. $\square$

\section{Proof of Proposition~\ref{eq:approx_error}}
\approxerr*
\bigskip

Let $\overline{V}_{\pi^*}(\mathbf{x}, t)$, $\overline{Q}_{\pi^*}(\mathbf{x}, a, t)$, and $\overline{p}^*_{cr}(\mathbf{x}[\mathbf{n}], \mathbf{x}[{k^+}], t)$ be the analogues of $V_{\pi^*}(\mathbf{x}, t)$, $Q_{\pi^*}(\mathbf{x}, a, t)$, and $p^*_{cr}(\mathbf{x}[\mathbf{n}], \mathbf{x}[{k^+}], t)$ in the continuous time problem described in Problem~\ref{prob:continuous_time}. 
Before commencing with the main proof, we first establish the following lemma which states that the value function in the continuous-time problem can be represented in the same piecewise-linear fashion as given in Proposition~\ref{prop:pwl_vf}.

\begin{lemma}[Piecewise-linear value function for continuous time]
    \label{lemma:pwl_vf}
    Let $\overline{V}_{\pi^*}(\mathbf{x}, t)$ be the optimal value function for Problem~\ref{prob:continuous_time}. $\overline{V}_{\pi^*}(\mathbf{x}, t)$ has the following form
    \begin{equation}
        \overline{V}_{\pi^*}(\mathbf{x}, t) = 
        \begin{cases}
            \Pr\big(\mathtt{fin}(\mathbf{x}[{k^+}])  \mid t \big)  \cdot \big(\mathbf{x}[p] - \overline{p}^*_{cr}(\mathbf{x}[\mathbf{n}], \mathbf{x}[{k^+}], t) \big)  
             + \overline{Q}_{\pi^*}(\mathbf{x}[\mathbf{n}], \mathtt{rej}, t), \vspace{0.5mm}\\
            \hspace{140pt} \textnormal{if } \mathbf{x}[{k^+}] \neq \bot \textnormal{ and } \mathbf{x}[p] \geq \overline{p}^*_{cr}(\mathbf{x}[\mathbf{n}], \mathbf{x}[{k^+}], t) \vspace{2mm}\\
            \overline{Q}_{\pi^*}(\mathbf{x}[\mathbf{n}], \mathtt{rej}, t), \textnormal{ otherwise.}
        \end{cases}
    \end{equation}
\end{lemma}

\noindent where $\overline{p}^*_{cr}$ is defined as
\begin{equation}
    \label{eq:pcrit_cont_time}
    \overline{p}^*_{cr}(\mathbf{x}[\mathbf{n}], \mathbf{x}[{k^+}], t) =  \frac{1}{\Pr(\mathtt{fin}(\mathbf{x}[{k^+}])  \mid t) } \times \\
    \Big(\overline{Q}_{\pi^*}(\mathbf{x}[\mathbf{n}], \mathtt{rej}, t) -  \overline{Q}_{\pi^*}(\mathbf{x}^{\mathtt{acc}}_{(\mathbf{x}[\mathbf{n}], \mathbf{x}[k^+])}[\mathbf{n}], \mathtt{rej}, t)
    \Big).
\end{equation}
\noindent if $\mathbf{x}[{k^+}] \neq \bot$ and $t < t_H$. We define $\overline{p}^*_{cr}(\mathbf{x}[\mathbf{n}], \mathbf{x}[{k^+}], t) = 0$ if $\mathbf{x}[{k^+}] = \bot$ or $t = t_H$.

A proof of Lemma~\ref{lemma:pwl_vf} can be made by following precisely the same reasoning as the proof of Proposition~\ref{prop:pwl_vf}.
We now commence with the main proof.

\bigskip

\noindent \textit{Main Proof}: 
We start by defining the following error functions:
\begin{equation}
    \label{eq:err1}
    \delta_{max}(t_i) \geq |\overline{Q}_{\pi^*}(\mathbf{n}, \mathtt{rej}, t_i) - Q_{\pi^*}(\mathbf{n}, \mathtt{rej}, t_i) |, \hspace{2pt} \textnormal{ for all } \mathbf{n} \in \mathbf{N}
\end{equation}

\begin{equation}
    \label{eq:err2}
    \epsilon_{max}(t_i) \geq |\overline{p}_{cr}^*(\mathbf{n}, k, t_i) - p_{cr}^*(\mathbf{n}, k, t_i)|, \hspace{2pt} \textnormal{ for all } \mathbf{n} \in \mathbf{N},\ k \in K
\end{equation}

\noindent where $t_i \in D$ is a decision epoch of the STA-HDMP.

The proof will proceed by induction over the decision epochs of the STA-HMDP. Consider any two arbitrary decision epochs of the STA-HMDP: $t_i$, and $t_{i+1}$. In the discrete time STA-HMDP approach, we only allow at most one arrival in $[t_i, t_{i+1}]$ to be accepted due to Assumption~\ref{ass:arrivals}. Any additional task arrivals are automatically rejected. Therefore, the STA-HMDP may miss out on reward due to additional task arrivals which are automatically rejected. On the other hand, the continuous time solution to Problem~\ref{prob:continuous_time} may accept tasks at any time without this constraint.
Therefore, the discrete-time STA-HDMP solution underestimates the expected value, i.e. $\overline{Q}_{\pi^*}(\mathbf{n}, \mathtt{rej}, t) \geq Q_{\pi^*}(\mathbf{n}, \mathtt{rej}, t)$. This implies that



\begin{equation}
    |\overline{Q}_{\pi^*}(\mathbf{n}, \mathtt{rej}, t_i) - Q_{\pi^*}(\mathbf{n}, \mathtt{rej}, t_i)| = \overline{Q}_{\pi^*}(\mathbf{n}, \mathtt{rej}, t_i) - Q_{\pi^*}(\mathbf{n}, \mathtt{rej}, t_i).
\end{equation}

We now proceed by finding a suitable upper bound for 
 $\overline{Q}_{\pi^*}(\mathbf{n}, \mathtt{rej}, t)$, and a suitable lower bound for $Q_{\pi^*}(\mathbf{n}, \mathtt{rej}, t)$.
 
\subsection{Upper Bound for $\overline{Q}_{\pi^*}(\mathbf{n}, \mathtt{rej}, t)$}
The value function can be expressed in the following form
\begin{equation}
    \overline{Q}_{\pi^*}(\mathbf{n}, \mathtt{rej}, t_i) = \sum_{N_{\mathtt{arr}} = 0}^\infty \overline{Q}_{\pi^*}(\mathbf{n}, \mathtt{rej}, t_i \mid \mathtt{arr}([t_i, t_{i+1}]) = N_{\mathtt{arr}}) \cdot \Pr \big( \mathtt{arr}([t_i, t_{i+1}])  = N_{\mathtt{arr}} \big),
\end{equation}

\noindent where $\mathtt{arr}([t_i, t_{i+1}]) = N_{\mathtt{arr}}$ means that $N_{\mathtt{arr}}$ tasks arrived between $t_i$ and $t_{i+1}$. We next consider three different cases depending on the value of $N_{\mathtt{arr}}$.

\subsubsection{Case where $N_{\mathtt{arr}} = 0$}

\begin{equation}
    \Pr(\mathtt{arr}([t_i, t_{i+1}]) = 0) = \exp \big(-\int_{t_i}^{t_{i+1}} \textstyle \sum_{k \in K} \Lambda_k(\tau) \dif \tau \big) \leq \exp \big( - \Delta t \sum_{k \in K} [\Lambda_k(t_i) - C_k \Delta t] \big) 
\end{equation}

\noindent where $C_k$ is the Lipschitz constant for each arrival rate function as described in the problem formulation (Problem~\ref{prob:continuous_time}). Given that no tasks arrive in the interval, the value at $t_i$ can be computed based on the value at $t_{i+1}$ as follows
\begin{equation}
    \label{eq:to_sub}
    \overline{Q}_{\pi^*}\big(\mathbf{n}, \mathtt{rej}, t_i \mid \mathtt{arr}([t_i, t_{i+1}]) = 0 \big) = \sum_{\mathbf{n}' \in \mathbf{N}} \Pr (\mathbf{n}', t_{i+1} \mid \mathbf{n}, t_i) \cdot \overline{Q}_{\pi^*}(\mathbf{n}', \mathtt{rej}, t_{i+1}),
\end{equation}

\noindent where we introduce the notation $\Pr(\mathbf{n}', t' \mid \mathbf{n}, t)$ to indicate the probability of there being $\mathbf{n}'$ servers busy at time $t'$ given that there were $\mathbf{n}$ servers busy at time $t$, and given that no tasks were accepted between $t$ and $t'$.

\subsubsection{Case where $N_{arr} = 1$} 
\begin{multline}
     \Pr( \mathtt{arr}([t_i, t_{i+1}]) = 1) = \int_{t_i}^{t_{i+1}}  \textstyle \sum_{k \in K} \displaystyle \Lambda_k(\tau) \dif \tau \cdot \exp \big(-\int_{t_i}^{t_{i+1}} \textstyle{\sum_{k \in K}} \Lambda_k(\tau) \dif \tau \big) \\ \leq 
      \textstyle \Delta t \sum_{k \in K} \big[ \Lambda_k(t_i) + C_k \Delta t \big] \exp \big( - \Delta t \sum_{k \in K} \big[ \Lambda_k(t_i) - C_k \Delta t \big] \big)
\end{multline}

We can express the value as follows
\begin{multline}
    \overline{Q}_{\pi^*}\big(\mathbf{n}, \mathtt{rej}, t_i \mid \mathtt{arr}([t_i, t_{i+1}]) = 1 \big) =  
    \sum_{k \in K} \Bigg[ \Pr(k_\mathtt{arr} = k \mid \mathtt{arr} \big([t_i, t_{i+1}]) = 1 \big) \cdot \\
    \int_{t_i}^{t_{i+1}} \sum_{\mathbf{n}_{\mathtt{arr}} \in \mathbf{N}}  \Pr(\mathbf{n}_{\mathtt{arr}}, \tau_{\mathtt{arr}} \mid \mathbf{n}, t_i)
    \cdot \Big[ \int_0^\infty  f_k(p) \cdot \overline{V}_{\pi^*}(\mathbf{x}', \tau_{\mathtt{arr}}\mid \mathtt{arr}([\tau_{\mathtt{arr}}, t_{i+1}]) = 0) \cdot \dif p \Big] \cdot \dif P \big(t_{\mathtt{arr}} \leq \tau_{\mathtt{arr}} \mid \mathtt{arr}_k([t_i, t_{i+1}]) = 1 \big) \Bigg]
\end{multline}

\noindent where:  $\Pr \big(k_\mathtt{arr} = k \mid \mathtt{arr}([t_i, t_{i+1}]) = 1 \big)$ is the probability that the task arrival is of class $k$, given that one task arrives; $\mathtt{arr}_k([t_i, t_{i+1}]) = 1$ indicates that there is one arrival of class $k$; $t_{\mathtt{arr}}$ is the time at which the task of class $k$ arrives;  $\mathbf{n}_{\mathtt{arr}}$ is the number of servers who are busy when the task arrives; $\mathbf{x}'[\mathbf{n}] = \mathbf{n}_\mathtt{arr}$; $\mathbf{x}'[k^+] = k$; and $\mathbf{x}'[p] = p$.  From Equation~\ref{eq:simplify_eq}, we may simplify the inner integral as follows

\begin{multline}
    \label{eq:36}
    \overline{Q}_{\pi^*}\big(\mathbf{n}, \mathtt{rej}, t_i \mid \mathtt{arr}([t_i, t_{i+1}]) = 1 \big) = \\
      \sum_{k \in K}  \Pr \big(k_\mathtt{arr} = k \mid \mathtt{arr}([t_i, t_{i+1}]) = 1 \big) \cdot 
    \Bigg[ \int_{t_i}^{t_{i+1}} \sum_{\mathbf{n}_{\mathtt{arr}} \in \mathbf{N} }  \Pr(\mathbf{n}_{\mathtt{arr}}, \tau_{\mathtt{arr}} \mid \mathbf{n}, t_i) \cdot \Big[ \overline{Q}_{\pi^*}\big(\mathbf{n}_{\mathtt{arr}}, \mathtt{rej}, \tau_{\mathtt{arr}} \mid \mathtt{arr}([\tau_{\mathtt{arr}}, t_{i+1}]) = 0 \big) + \\
    \Pr\big(\mathtt{fin}(k)  \mid \tau_{\mathtt{arr}} \big) \cdot \phi \big(\overline{p}_{cr}^* (\mathbf{n}_{\mathtt{arr}}, k, \tau_{\mathtt{arr}}) \big) \Big] \cdot \dif P \big(t_{\mathtt{arr}} \leq \tau_{\mathtt{arr}} \mid \mathtt{arr}_k([t_i, t_{i+1}]) = 1 \big) \Bigg]
\end{multline}

By substituting Equation~\ref{eq:to_sub} we have that

\begin{multline}
    \label{eq:simplify_da_probs}
    \sum_{\mathbf{n}_{\mathtt{arr}} \in \mathbf{N}} \Pr(\mathbf{n}_{\mathtt{arr}}, \tau_{\mathtt{arr}} \mid \mathbf{n}, t_i) \cdot \overline{Q}_{\pi^*}\big(\mathbf{n}_{\mathtt{arr}}, \mathtt{rej}, \tau_{\mathtt{arr}} \mid \mathtt{arr}([\tau_{\mathtt{arr}}, t_{i+1}]) = 0 \big) = \\
    \sum_{\mathbf{n}_{\mathtt{arr}} \in \mathbf{N}}  \Pr(\mathbf{n}_{\mathtt{arr}}, \tau_{\mathtt{arr}} \mid \mathbf{n}, t_i) \sum_{\mathbf{n}' \in \mathbf{N} } \Pr (\mathbf{n}', t_{i+1} \mid \mathbf{n}_{\mathtt{arr}}, \tau_{\mathtt{arr}}) \cdot \overline{Q}_{\pi^*}(\mathbf{n}', \mathtt{rej}, t_{i+1})
    =  \\
    \sum_{\mathbf{n}' \in \mathbf{N}} \Pr (\mathbf{n}', t_{i+1} \mid \mathbf{n}, t_i) \cdot \overline{Q}_{\pi^*}(\mathbf{n}', \mathtt{rej}, t_{i+1})
\end{multline}

\noindent where the second equality in Equation~\ref{eq:simplify_da_probs} can be verified by observing that $ \sum_{\mathbf{n}_{\mathtt{arr}} \in \mathbf{N}}  \Pr(\mathbf{n}_{\mathtt{arr}}, \tau_{\mathtt{arr}} \mid \mathbf{n}, t_i) \cdot  \Pr (\mathbf{n}', t_{i+1} \mid \mathbf{n}_{\mathtt{arr}}, \tau_{\mathtt{arr}}) = \Pr (\mathbf{n}', t_{i+1} \mid \mathbf{n}_{\mathtt{arr}}, \tau_{\mathtt{arr}})  $. Noting that the right hand side of Equation~\ref{eq:simplify_da_probs} does not depend on $\tau_{\mathtt{arr}}$, we can substitute Equation~\ref{eq:simplify_da_probs} to rewrite Equation~\ref{eq:36} as

\begin{multline}
    \overline{Q}_{\pi^*}\big(\mathbf{n}, \mathtt{rej}, t_i \mid \mathtt{arr}([t_i, t_{i+1}]) = 1 \big) = 
    \sum_{k \in K} \Pr \big(k_\mathtt{arr} = k \mid \mathtt{arr}([t_i, t_{i+1}]) = 1 \big) \cdot
    \Bigg[  \sum_{\mathbf{n}' \in \mathbf{N}} \Pr (\mathbf{n}', t_{i+1} | \mathbf{n}, t_i) \cdot \overline{Q}_{\pi^*}(\mathbf{n}', \mathtt{rej}, t_{i+1}) + \\
    \int_{t_i}^{t_{i+1}} \sum_{\mathbf{n}' \in \mathbf{N}}  \Pr(\mathbf{n}', \tau_{\mathtt{arr}} \mid \mathbf{n}, t_i) \cdot  \Pr\big(\mathtt{fin}(k)  \mid \tau_{\mathtt{arr}} \big) \cdot \phi(\overline{p}_{cr}^*(\mathbf{n}', k, \tau_{\mathtt{arr}}))  \cdot \dif P\big(t_{\mathtt{arr}} \leq \tau_{\mathtt{arr}} \mid \mathtt{arr}_k([t_i, t_{i+1}]) = 1 \big) \Bigg]
\end{multline}

For each of the terms in the integral, we add and subtract the corresponding values at $t_{i+1}$

\begin{multline}
    \label{eq:39}
    \overline{Q}_{\pi^*}\big(\mathbf{n}, \mathtt{rej}, t_i \mid \mathtt{arr}([t_i, t_{i+1}]) = 1 \big) = 
     \sum_{k \in K} \Pr \big(k_\mathtt{arr} = k \mid \mathtt{arr}([t_i, t_{i+1}]) = 1 \big) \cdot
   \Bigg[  \sum_{\mathbf{n}' \in \mathbf{N}} \Pr (\mathbf{n}', t_{i+1} | \mathbf{n}, t_i) \cdot \overline{Q}_{\pi^*}(\mathbf{n}', \mathtt{rej}, t_{i+1}) +  \\ 
    \int_{t_i}^{t_{i+1}} \sum_{\mathbf{n}' \in \mathbf{N}}  \Big( \Pr(\mathbf{n}', t_{i+1} \mid \mathbf{n}, t_i) + \Pr(\mathbf{n}', \tau_{\mathtt{arr}} \mid \mathbf{n}, t_i) - \Pr(\mathbf{n}', t_{i+1} \mid \mathbf{n}, t_i)  \Big) \cdot \\ 
    \Big(\Pr \big(\mathtt{fin}(k)  \mid t_{i+1} \big) + \Pr \big(\mathtt{fin}(k)  \mid \tau_{\mathtt{arr}} \big) - \Pr \big(\mathtt{fin}(k)  \mid t_{i+1} \big)  \Big)   \cdot \\
    \Big(\phi(\overline{p}_{cr}^*(\mathbf{n}', k, t_{i+1})) + \phi(\overline{p}_{cr}^*(\mathbf{n}', k, \tau_{\mathtt{arr}}))  - \phi(\overline{p}_{cr}^*(\mathbf{n}', k, t_{i+1})) \Big)  \cdot 
    \dif P \big(t_{\mathtt{arr}} \leq \tau_{\mathtt{arr}} \mid \mathtt{arr}_k([t_i, t_{i+1}]) = 1 \big) \Bigg]
\end{multline}

We wish to upper bound each of the terms in Equation~\ref{eq:39}. We start by upper bounding ${\Pr \big(k_\mathtt{arr} = k \mid \mathtt{arr}([t_i, t_{i+1}]) = 1 \big)}$ as follows:
\begin{equation}
    \label{eq:the_other_won}
    \Pr \big(k_\mathtt{arr} = k \mid \mathtt{arr}([t_i, t_{i+1}]) = 1 \big) = \frac{\int_{t_i}^{t_{i+1}}  \Lambda_k(\tau) \dif \tau}{ \int_{t_i}^{t_{i+1}}  \textstyle \sum_{k' \in K} \displaystyle \Lambda_{k'}(\tau) \dif \tau } \leq  \frac{ [ \Lambda_k(t_i) + C_k \Delta t ]}{ \sum_{k' \in K}[ \Lambda_{k'}(t_i) - C_{k'} \Delta t ]}
\end{equation}

We will now find a suitable upper bound for the differences in values at $\tau_{\mathtt{arr}}$ and $t_{i+1}$ of each of the three terms in the integral in Equation~\ref{eq:39}.

\medskip
\noindent \textit{First Term}: \\
The probability distribution over the number of servers at the next time step is given by Equation~\ref{eq:server_fin}. This leads to

\begin{multline}
    \Bigm\lvert \Pr(\mathbf{n}', \tau_{\mathtt{arr}} \mid \mathbf{n}, t_i) - \Pr(\mathbf{n}', t_{i+1} \mid \mathbf{n}, t_i) \Bigm\lvert = \\
    \Biggm\lvert \prod_{k \in K} \binom{n_k}{n_k - n_k'} \Big(1 - \exp \big(-\mu_k (\tau_{\mathtt{arr}} - t_i) \big) \Big)^{n_k - n_k'}\exp \big(-\mu_k n_k'(\tau_{\mathtt{arr}} - t_i) \big) - \\
    \prod_{k \in K} \binom{n_k}{n_k - n_k'} \Big(1 - \exp \big(-\mu_k \Delta t \big) \Big)^{n_k - n_k'}\exp \big(-\mu_k n_k'\Delta t \big)  \Biggm\lvert
\end{multline}

\noindent where to simplify notation we write $n_k$ to mean the $k^{th}$ component of $\mathbf{n}$, and $n_k'$ to mean the $k^{th}$ component of $\mathbf{n'}$.
The expression is maximised when $\tau_{\mathtt{arr}} = t_i$, leading to 

\begin{multline}
    \label{eq:can_be_verified}
    \Bigm\lvert \Pr(\mathbf{n}', \tau_{\mathtt{arr}} \mid \mathbf{n}, t_i) - \Pr(\mathbf{n}', t_{i+1} \mid n_k, t_i) \Bigm\lvert \leq \\
    \Biggm\lvert \prod_{k \in K} \binom{n_k}{n_k - n_k'} (0 )^{n_k - n_k'} - 
    \prod_{k \in K} \binom{n_k}{n_k - n_k'} \Big(1 - \exp \big(-\mu_k \Delta t \big) \Big)^{n_k - n_k'}\exp \big(-\mu_k n_k'\Delta t \big) 
    \Biggm\lvert \\
    = 
    \begin{cases}
        1 - \textstyle \prod_{k \in K} \exp\big(- \mu_k n_k' \Delta t \big), 
        & \textnormal{ if } n_k' = n_k \ \forall k\in K\\
        \prod_{k \in K} \binom{n_k}{n_k - n_k'} \Big(1 - \exp \big(-\mu_k \Delta t \big) \Big)^{n_k - n_k'}\exp \big(-\mu_k n_k'\Delta t \big), 
        & \textnormal{ if } n_k' < n_k\ \textnormal{for some }k \in K
    \end{cases}
\end{multline}

From Equation~\ref{eq:can_be_verified} it can be verified that 
\begin{equation}
    \Bigm\lvert \Pr(\mathbf{n}', \tau_{\mathtt{arr}} \mid n_k, t_i) - \Pr(\mathbf{n}', t_{i+1} \mid n_k, t_i) \Bigm\lvert \leq N_{serv} !^{|K|} \big(1 - \exp(-\mu_k   N_{serv} |K| \Delta t) \big)
\end{equation}

\medskip
\noindent \textit{Second Term}: \\
\begin{multline}
    \label{eq:43}
    \Bigm\lvert \Pr \big(\mathtt{fin}(k)  \mid \tau_{\mathtt{arr}} \big) - \Pr \big(\mathtt{fin}(k)  \mid t_{i+1} \big) \Bigm\lvert \\
    =\ \Bigm\lvert \Big(1 - \exp \big(-(t_H - \tau_{\mathtt{arr}})\mu_{k} \big) \Big) - \Big(1 - \exp \big(-(t_H - t_{i+1})\mu_{k} \big) \Big) \Bigm\lvert \\ 
    =\ \Bigm\lvert \exp \big(-(t_H - t_{i+1})\mu_{k} \big) - \exp \big(-(t_H - t_{\mathtt{arr}})\mu_{k} \big)  \Bigm\lvert
\end{multline}

Equation~\ref{eq:43} is maximised if $t_{\mathtt{arr}} = t_i = t_{i+1} - \Delta t$. Therefore
\begin{multline}
    \Bigm\lvert \Pr \big(\mathtt{fin}(k)  \mid \tau_{\mathtt{arr}} \big) - \Pr \big(\mathtt{fin}(k)  \mid t_{i+1} \big) \Bigm\lvert\  \leq\ 
    \Bigm\lvert \exp \big(-(t_H - t_{i+1})\mu_{k} \big) - \exp \big(-(t_H - t_{i+1} + \Delta t)\mu_{k} \big)  \Bigm\lvert \\
     =\ \Bigm\lvert (1 - \exp \big(-\mu_{k} \Delta t ) \big)\cdot  \exp \big(-(t_H - t_{i+1})\mu_{k} \big) \Bigm\lvert  
    \leq \big(1 - \exp(-\Delta t \cdot \mu_{k}) \big)
\end{multline}

\medskip
\noindent \textit{Third Term}: \\
Finally, we wish to upper bound the third term in the integral, $\mid \phi(\overline{p}_{cr}^*(\mathbf{n}, k, \tau_{\mathtt{arr}}))  - \phi(\overline{p}_{cr}^*(\mathbf{n}, k, t_{i+1})) \mid$. Before bounding this term, we establish some necessary intermediate results. 

We start with the definition of $\overline{p}_{cr}^*(\mathbf{n}, k, t)$ for $t < t_H$ and we use the product rule to bound the magnitude of derivative with respect to time
\begin{equation}
    \overline{p}_{cr}^*(\mathbf{n}, k, t) = \frac{1}{\Pr \big(\mathtt{fin}(k)  \mid t \big)} \Big( \overline{Q}_{\pi^*}(\mathbf{n}, \mathtt{rej}, t) - \overline{Q}_{\pi^*}(\mathbf{x}^{\mathtt{acc}}_{(\mathbf{n}, k)}[\mathbf{n}], \mathtt{rej}, t) \Big)
\end{equation}
\begin{multline}
   \frac{\dif \overline{p}_{cr}^*(\mathbf{n}, k, t) }{\dif t} = 
   \Big( \overline{Q}_{\pi^*}(\mathbf{n}, \mathtt{rej}, t) - \overline{Q}_{\pi^*}(\mathbf{x}^{\mathtt{acc}}_{(\mathbf{n}, k)}[\mathbf{n}], \mathtt{rej}, t) \Big) \cdot \frac{\dif }{\dif t} \Big( \frac{1}{1 - \exp(-\mu_k(t_H - t))}  \Big) + \\
   \frac{1}{1 - \exp(-\mu_k(t_H - t))}  \cdot \frac{\dif }{\dif t} \Big( \overline{Q}_{\pi^*}(\mathbf{n}, \mathtt{rej}, t) - \overline{Q}_{\pi^*}(\mathbf{x}^{\mathtt{acc}}_{(\mathbf{n}, k)}[\mathbf{n}], \mathtt{rej}, t) \Big) \\
\end{multline}
\begin{multline}
    \label{eq:ineq}
   \Bigm\lvert \frac{\dif  \overline{p}_{cr}^*(\mathbf{n}, k, t) }{\dif t} \Bigm\lvert \leq 2 \sum_{k' \in K} \Big[ \max_{t'} \big( \Lambda_{k'} (t') \big) \cdot  \mathbb{E}[p_{k'}]  \cdot \Pr \big(\mathtt{fin}(k')  \mid t \big) \cdot (t_H - t) \Big] \frac{\mu_k\exp(-\mu_k(t_H - t))}{\big(1-\exp(-\mu_k(t_H-t)) \big)^2}  \\
   +  \frac{2}{1 - \exp(-\mu_k(t_H - t))} \sum_{k' \in K} \Big[ \max_{t'}  \big( \Lambda_{k'} (t') \big) \cdot  \mathbb{E}[p_{k'}]  \cdot \Pr \big(\mathtt{fin}(k')  \mid t \big) \Big]
\end{multline}

Equation~\ref{eq:ineq} holds because the right hand side assumes that tasks arrive at the maximum rate and that all tasks can be accepted and therefore overestimates the $Q$-values and the rate of change of the $Q$-values. We multiply and divide each expression on the right hand side by $\Pr \big(\mathtt{fin}(k)  \mid t \big)$, substitute the expression for $\Pr \big(\mathtt{fin}(k)  \mid t \big)$, and simplify
\begin{multline}
    \label{eq:55}
   \Bigm\lvert \frac{\dif  \overline{p}_{cr}^*(\mathbf{n}, k, t) }{\dif t} \Bigm\lvert 
   \leq 2 \sum_{k' \in K} \Big[ \max_{t'} \big( \Lambda_{k'} (t') \big) \cdot  \mathbb{E}[p_{k'}] \cdot \frac{\Pr \big(\mathtt{fin}(k')  \mid t \big)}{\Pr \big(\mathtt{fin}(k)  \mid t \big)} \Big] \cdot  \frac{\mu_k\exp(-\mu_k(t_H - t))(t_H - t)}{1-\exp(-\mu_k(t_H-t)) } \\
   + 2 \sum_{k' \in K} \Big[ \max_{t'} \big( \Lambda_{k'} (t') \big) \cdot  \mathbb{E}[p_{k'}]  \frac{\Pr \big(\mathtt{fin}(k')  \mid t \big)}{\Pr \big(\mathtt{fin}(k)  \mid t \big)} \Big]
\end{multline}

Because 
\begin{equation}
    \frac{\mu_k\exp(-\mu_k(t_H - t))(t_H - t)}{1-\exp(-\mu_k(t_H-t)) } \leq 1 \textnormal{, \hspace{3pt} for } t\in[0, t_H)
\end{equation}
\noindent we have from Equation~\ref{eq:55} that
\begin{equation}
    \Bigm\lvert \frac{\dif  \overline{p}_{cr}^*(\mathbf{n}, k, t) }{\dif t} \Bigm\lvert \leq 4 \sum_{k' \in K} \Big[ \max_{t'} \big( \Lambda_{k'} (t') \big) \cdot  \mathbb{E}[p_{k'}] \cdot \frac{\Pr \big(\mathtt{fin}(k')  \mid t \big)}{\Pr \big(\mathtt{fin}(k)  \mid t \big)} \Big] \textnormal{, \hspace{3pt} for } t\in[0, t_H)
\end{equation}
\begin{equation}
    \label{eq:diff_bound_1}
    \bigm \lvert \overline{p}_{cr}^*(\mathbf{n}, k, t_{\mathtt{arr}}) - \overline{p}_{cr}^*(\mathbf{n}, k, t_{i+1}) \bigm \lvert \leq 4 \sum_{k' \in K} \Big[ \max_{t'} \big( \Lambda_{k'} (t') \big) \cdot  \mathbb{E}[p_{k'}] \cdot \frac{\Pr \big(\mathtt{fin}(k')  \mid t \big)}{\Pr \big(\mathtt{fin}(k)  \mid t \big)} \Big]\cdot\Delta t \textnormal{, \hspace{3pt} for } t_{i+1}\in[0, t_H)
\end{equation}

For the case where $t_{i+1} = t_H$, we have that $\overline{p}_{cr}^*(\mathbf{n}, k, t_{H}) = 0$ by definition. Additionally, if $t_{i+1} = t_H$, then
\begin{multline}
    \overline{p}_{cr}^*(\mathbf{n}, k, t_{\mathtt{arr}}) = \frac{1}{\Pr \big(\mathtt{fin}(k)  \mid t_{\mathtt{arr}} \big)} \Big( \overline{Q}_{\pi^*}(\mathbf{n}, \mathtt{rej}, t_{\mathtt{arr}}) - \overline{Q}_{\pi^*}(\mathbf{x}^{\mathtt{acc}}_{(\mathbf{n}, k)}[\mathbf{n}], \mathtt{rej}, t_{\mathtt{arr}}) \Big) \\
    \leq \frac{2}{\Pr \big(\mathtt{fin}(k)  \mid t_{\mathtt{arr}} \big)} \sum_{k' \in K} \Big[ \max_{t'} \big( \Lambda_{k'} (t') \big) \cdot  \mathbb{E}[p_{k'}]  \cdot (t_H - t_{\mathtt{arr}}) \cdot \Pr \big(\mathtt{fin}(k')  \mid t_{\mathtt{arr}} \big) \Big] \\ 
    \leq 2 \sum_{k' \in K} \Big[ \max_{t'} \big( \Lambda_{k'} (t') \big) \cdot  \mathbb{E}[p_{k'}] \cdot \frac{\Pr \big(\mathtt{fin}(k')  \mid t_{\mathtt{arr}}) \big)}{\Pr \big(\mathtt{fin}(k)  \mid t_{\mathtt{arr}}) \big)} \Big]\cdot\Delta t 
\end{multline}

Therefore, in the case where $t_{i+1} = t_H$ we have 
\begin{equation}
    \label{eq:diff_bound_2}
     \bigm\lvert \overline{p}_{cr}^*(\mathbf{n}, k, t_{\mathtt{arr}}) - \overline{p}_{cr}^*(\mathbf{n}, k, t_{i+1}) \bigm\lvert  \leq 2 \sum_{k' \in K} \Big[ \max_{t'} \big( \Lambda_{k'} (t') \big) \cdot  \mathbb{E}[p_{k'}] \cdot \frac{\Pr \big(\mathtt{fin}(k')  \mid t_{\mathtt{arr}}) \big)}{\Pr \big(\mathtt{fin}(k)  \mid t_{\mathtt{arr}}) \big)} \Big]\cdot \Delta t \textnormal{, \hspace{3pt} for } t_{i+1} = t_H
\end{equation}

Because ${\Pr \big(\mathtt{fin}(k')  \mid t) \big)}/{\Pr \big(\mathtt{fin}(k)  \mid t) \big)} \leq \max \{1, \mu_{k'}/\mu_k \}$, for any $t \in [0, t_H)$, from Equations~\ref{eq:diff_bound_1} and~\ref{eq:diff_bound_2} we have that 
\begin{equation}
    \label{eq:final_diff_bound}
    \bigm\lvert \overline{p}_{cr}^*(\mathbf{n}, k, t_{\mathtt{arr}}) - \overline{p}_{cr}^*(\mathbf{n}, k, t_{i+1}) \bigm\lvert \leq M \Delta t
\end{equation}

\noindent for some finite constant $M$, and any $t_{i+1} \in [0, t_H]$.

Next we consider the mean shortage function. The derivative for the mean shortage function is 
\begin{equation}
    \label{eq:49}
    \od{ \phi_k(p)}{p}
 = \od{}{p} \int_p^\infty (1 - F_k(y)) \dif y = F_k(p) - 1 
\end{equation}

Therefore, 
\begin{equation}
    \label{eq:dphi_dp}
    \Bigm \lvert
    \od{\phi_k(p)}{p} \Bigm \lvert \ \leq 1,\ \forall p
\end{equation}

Equation~\ref{eq:dphi_dp} combined with Equation~\ref{eq:final_diff_bound} yields the desired bound
\begin{equation}
    \bigm\lvert \phi(\overline{p}_{cr}^*(\mathbf{n}, k, \tau_{\mathtt{arr}}))  - \phi(\overline{p}_{cr}^*(\mathbf{n}, k, t_{i+1})) \bigm\lvert
    \leq M\Delta t
\end{equation}

\medskip
\noindent \textit{Substituting all three terms}: \\
Now we can substitute the upper bounds on each of the three terms, as well as the upper bound on ${\Pr \big(k_\mathtt{arr} = k \mid \mathtt{arr}([t_i, t_{i+1}]) = 1 \big)}$ given in Equation~\ref{eq:the_other_won} into Equation~\ref{eq:39} to obtain an upper bound on ${\overline{Q}_{\pi^*}\big(n_k, \mathtt{rej}, t_i \mid \mathtt{arr}([t_i, t_{i+1}]) = 1 \big)}$:
\begin{multline}
    \overline{Q}_{\pi^*}\big(\mathbf{n}, \mathtt{rej}, t_i \mid \mathtt{arr}([t_i, t_{i+1}]) = 1 \big) \leq 
    \sum_{k \in K}  \frac{ [ \Lambda_k(t_i) + C_k \Delta t ]}{ \sum_{k' \in K}[ \Lambda_{k'}(t_i) - C_{k'} \Delta t ]} \cdot 
    \Bigg[ \sum_{\mathbf{n}' \in \mathbf{N}} \Pr (\mathbf{n}', t_{i+1} \mid \mathbf{n}, t_i) \cdot \overline{Q}_{\pi^*}(\mathbf{n}', \mathtt{rej}, t_{i+1}) +  \\ 
    \int_{t_i}^{t_{i+1}} \sum_{\mathbf{n}' \in \mathbf{N}}  \Big( \Pr(\mathbf{n}', t_{i+1} \mid \mathbf{n}, t_i) + N_{serv} !^{|K|} \big(1 - \exp(-\mu_k N_{serv} |K| \Delta t) \big) \Big) \cdot 
    \Big(\Pr \big(\mathtt{fin}(k)  \mid t_{i+1} \big) + \big(1 - \exp(-\Delta t \cdot \mu_{k}) \big) \Big)   \cdot \\
    \Big(\phi(\overline{p}_{cr}^*(\mathbf{n}', k, t_{i+1})) + M\Delta t \Big)  \cdot 
    \dif P \big(t_{\mathtt{arr}} \leq \tau_{\mathtt{arr}} \mid \mathtt{arr}_k([t_i, t_{i+1}]) = 1 \big) \Bigg]
\end{multline}

None of the terms in the integral depend on $\tau_{\mathtt{arr}}$, so we can rewrite this without the integral
\begin{multline}
    \overline{Q}_{\pi^*}\big(\mathbf{n}, \mathtt{rej}, t_i \mid \mathtt{arr}([t_i, t_{i+1}]) = 1 \big) \leq 
    \sum_{k \in K}  \frac{ [ \Lambda_k(t_i) + C_k \Delta t ]}{ \sum_{k' \in K}[ \Lambda_{k'}(t_i) - C_{k'} \Delta t ]} \cdot 
    \Bigg[ \sum_{\mathbf{n}' \in \mathbf{N}} \Pr (\mathbf{n}', t_{i+1} \mid \mathbf{n}, t_i) \cdot \overline{Q}_{\pi^*}(\mathbf{n}', \mathtt{rej}, t_{i+1}) +  \\ 
   \sum_{\mathbf{n}' \in \mathbf{N}}  \Big( \Pr(\mathbf{n}', t_{i+1} \mid \mathbf{n}, t_i) + N_{serv} !^{|K|} \big(1 - \exp(-\mu_k N_{serv} |K| \Delta t) \big) \Big) \cdot 
    \Big(\Pr \big(\mathtt{fin}(k)  \mid t_{i+1} \big) + \big(1 - \exp(-\Delta t \cdot \mu_{k}) \big) \Big)   \cdot \\
    \Big(\phi(\overline{p}_{cr}^*(\mathbf{n}', k, t_{i+1})) + M\Delta t \Big)  \Bigg]
\end{multline}

\subsubsection{Case where $N_{\mathtt{arr}} > 1$ \\} 

For the case when more than one task arrives,

\begin{multline}
     \Pr(\mathtt{arr}([t_i, t_{i+1}]) = N_{\mathtt{arr}}) =
     \frac{1}{N_{\mathtt{arr}}!} \Big( \int_{t_i}^{t_{i+1}} \textstyle \sum_{k \in K} \Lambda_k(\tau) \displaystyle \dif \tau \Big)^{N_{\mathtt{arr}}}
     \exp \big(-\int_{t_i}^{t_{i+1}} \textstyle \sum_{k \in K}\Lambda_k(\tau) \dif \tau \big) \\
     \leq \Big(  \Delta t \textstyle \sum_{k \in K} [\Lambda_k(t_i) + C_k \Delta t] \Big)^{N_{\mathtt{arr}}} \exp \big( - \Delta t \sum_{k \in K}[\Lambda_k(t_i) - C_k \Delta t] \big)
\end{multline}

For the value, we use the following simple upper bound

\begin{equation}
    \label{eq:upper_bound_lots}
    \overline{Q}_{\pi^*}\big(\mathbf{n}, \mathtt{rej}, t_i \mid \mathtt{arr}([t_i, t_{i+1}]) = N_{\mathtt{arr}} \big) \leq N_{\mathtt{arr}} \cdot \max_{k \in K} \mathbb{E}[p_k] + \overline{Q}_{\pi^*}\big(\mathbf{n}, \mathtt{rej}, t_i)
\end{equation}

Equation~\ref{eq:upper_bound_lots} is an upper bound as it assumes that all tasks are the most valuable task class, and that all tasks completed instantaneously.

\subsubsection{Final Upper Bound for $\overline{Q}_{\pi^*}\big(\mathbf{n}, \mathtt{rej}, t_i \big)$ \\}
We are finally ready to state the upper bound on the value in the continuous-time problem at $t_i$ as a function of the value at $t_{i+1}$. 

\begin{multline}
    \overline{Q}_{\pi^*}(\mathbf{n}, \mathtt{rej}, t_i) \leq 
    \Pr(\mathtt{arr}([t_i, t_{i+1}]) = 0) \cdot \sum_{\mathbf{n}' \in \mathbf{N}} \Pr (\mathbf{n}', t_{i+1} | \mathbf{n}, t_i) \cdot \overline{Q}_{\pi^*}(\mathbf{n}', \mathtt{rej}, t_{i+1})\\
    + \Pr(\mathtt{arr}([t_i, t_{i+1}]) = 1)\cdot 
    \sum_{k \in K}  \frac{ [ \Lambda_k(t_i) + C_k \Delta t ]}{ \sum_{k' \in K}[ \Lambda_{k'}(t_i) - C_{k'} \Delta t ]} \cdot 
    \Bigg[ \sum_{\mathbf{n}' \in \mathbf{N}} \Pr (\mathbf{n}', t_{i+1} \mid \mathbf{n}, t_i) \cdot \overline{Q}_{\pi^*}(\mathbf{n}', \mathtt{rej}, t_{i+1}) +  \\ 
   \sum_{\mathbf{n}' \in \mathbf{N}}  \Big( \Pr(\mathbf{n}', t_{i+1} \mid \mathbf{n}, t_i) + N_{serv} !^{|K|} \big(1 - \exp(-\mu_k N_{serv} |K| \Delta t) \big) \Big) \cdot 
    \Big(\Pr \big(\mathtt{fin}(k)  \mid t_{i+1} \big) + \big(1 - \exp(-\Delta t \cdot \mu_{k}) \big) \Big)   \cdot \\
    \Big(\phi(\overline{p}_{cr}^*(\mathbf{n}', k, t_{i+1})) + M\Delta t \Big)  \Bigg] 
    + \sum_{N_{\mathtt{arr}} = 2}^\infty \Bigg [\Pr(\mathtt{arr}([t_i, t_{i+1}]) = N_{\mathtt{arr}}) \cdot \Big( N_{\mathtt{arr}} \cdot \max_{k \in K}\mathbb{E}[p_k] + \overline{Q}_{\pi^*}\big(\mathbf{n}, \mathtt{rej}, t_i) \Big) \Bigg] \\
    = \mathtt{Upper Bound}\big(\overline{Q}_{\pi^*}(\mathbf{n}, \mathtt{rej}, t_i) \big)
\end{multline}

\subsection{Lower Bound for $Q_{\pi^*}\big(\mathbf{n}, \mathtt{rej}, t_i \big)$ }
The optimal value in the STA-HMDP is computed as follows

\begin{multline}
    \label{eq:57}
    Q_{\pi^*}(\mathbf{n}, \mathtt{rej}, t_i) = 
    \exp(- \Delta t \textstyle \sum_{k \in K} \displaystyle \Lambda_k(t_{i}) ) \cdot \sum_{\mathbf{n}' \in \mathbf{N}} \Pr (\mathbf{n}', t_{i+1} | \mathbf{n}, t_i) \cdot Q_{\pi^*}(\mathbf{n}', \mathtt{rej}, t_{i+1}) + \\
     \big(1 - \exp(- \Delta t \textstyle \sum_{k \in K} \displaystyle \Lambda_k(t_{i}) ) \big) \cdot 
    \sum_{k \in K}  \frac{ \Lambda_k(t_i)}{ \sum_{k' \in K}\Lambda_{k'}(t_i)  }
    \sum_{\mathbf{n}' \in \mathbf{N}} \Pr (\mathbf{n}', t_{i+1} | \mathbf{n}, t_i) \Big[ Q_{\pi^*}(\mathbf{n}', \mathtt{rej}, t_{i+1}) + \\ \Pr\big(\mathtt{fin}(k)  \mid t_{i+1} \big) \cdot \phi_{k}(p^*_{cr}(\mathbf{n}, k, t_{i+1})) \Big] 
    \geq \exp(- \Delta t \textstyle \sum_{k \in K} \displaystyle \Lambda_k(t_{i}) ) \cdot \sum_{\mathbf{n}' \in \mathbf{N}} \Pr (\mathbf{n}', t_{i+1} | \mathbf{n}, t_i) \cdot Q_{\pi^*}(\mathbf{n}', \mathtt{rej}, t_{i+1}) + \\
    \Delta t \textstyle \sum_{k \in K} \displaystyle \big(\Lambda_k(t_i) \big)\exp(- \Delta t \textstyle \sum_{k \in K} \displaystyle \Lambda_k(t_{i}) )  \cdot 
    \sum_{k \in K}  \frac{ \Lambda_k(t_i)}{ \sum_{k' \in K}\Lambda_{k'}(t_i)  }
    \sum_{\mathbf{n}' \in \mathbf{N}}  \Pr (\mathbf{n}', t_{i+1} | \mathbf{n}, t_i) \Big[ Q_{\pi^*}(\mathbf{n}', \mathtt{rej}, t_{i+1}) + \\ \Pr\big(\mathtt{fin}(k)  \mid t_{i+1} \big) \cdot \phi_{k}(p^*_{cr}(\mathbf{n}, k, t_{i+1})) \Big] \\
\end{multline}

\noindent where the inequality holds because $\big(1 - \exp(- \Delta t \textstyle \sum_{k \in K} \displaystyle \Lambda_k(t_{i}) ) \big)$ is the probability that at least one task arrives, whereas $\Delta t \textstyle \sum_{k \in K} \displaystyle \big(\Lambda_k(t_i) \big)\exp(- \Delta t \textstyle \sum_{k \in K} \displaystyle \Lambda_k(t_{i}) ) $ is the probability that exactly one task arrives.
We have that 
\begin{equation}
    \overline{p}^*_{cr}(\mathbf{n}, k, t_{i+1}) - \epsilon_{max} (t_{i+1}) \leq p^*_{cr}(\mathbf{n}, k, t_{i+1}) \leq \overline{p}^*_{cr}(\mathbf{n}, k, t_{i+1})
\end{equation}
\noindent where the first inequality is by the definition of the error function in Equation~\ref{eq:err2}. The second inequality holds because the critical price is defined to be proportional to the marginal value of having an additional server available rather than processing a task (Proposition~\ref{prop:pwl_vf}). This value is lower in the discrete time STA-HDMP than in continuous time because some tasks are automatically rejected in the STA-HDMP formulation due to Assumption~\ref{ass:arrivals}. 

Remembering that $d\phi_k(p) / dp \leq 0$ for all $p$ (Equation~\ref{eq:49}), this means that \begin{equation}
   \phi_{k}(p^*_{cr}(\mathbf{n}, k, t_{i+1})) \geq \phi_{k}(\overline{p}^*_{cr}(\mathbf{n}, k, t_{i+1}))
\end{equation} 

Substituting this into Equation~\ref{eq:57} and utilising the definition for the error function
 $\delta_{max}(t)$ in Equation~\ref{eq:err1}  yields
\begin{multline}
    Q_{\pi^*}(\mathbf{n}, \mathtt{rej}, t_i) \geq \exp(- \Delta t \textstyle \sum_{k \in K}  \displaystyle \Lambda_k(t_{i}) ) \cdot \sum_{\mathbf{n}' \in \mathbf{N}} \Pr (\mathbf{n}', t_{i+1} | \mathbf{n}, t_i) \cdot \Big(\overline{Q}_{\pi^*}(\mathbf{n}', \mathtt{rej}, t_{i+1}) - \delta_{max}(t_{i+1}) \Big) \\
    + \Delta t \textstyle \sum_{k \in K} \displaystyle \big(\Lambda_k(t_i) \big)\exp(- \Delta t \textstyle \sum_{k \in K} \displaystyle \Lambda_k(t_{i}) ) \big) \cdot 
    \sum_{k \in K}  \frac{ \Lambda_k(t_i)}{ \sum_{k' \in K}\Lambda_{k'}(t_i)  }
    \sum_{\mathbf{n}'}\Pr (\mathbf{n}', t_{i+1} | \mathbf{n}, t_i) \Big[ \overline{Q}_{\pi^*}(\mathbf{n}', \mathtt{rej}, t_{i+1}) - \\\delta_{max}(t_{i+1}) +  
    \Pr\big(\mathtt{fin}(k)  \mid t_{i+1} \big) \cdot  \phi_{k}(\overline{p}^*_{cr}(\mathbf{n}, k, t_{i+1}))  \Big]
\end{multline}

\begin{multline}
     Q_{\pi^*}(\mathbf{n}, \mathtt{rej}, t_i) \geq \exp(- \Delta t \textstyle \sum_{k \in K}  \displaystyle \Lambda_k(t_{i}) ) \cdot \sum_{\mathbf{n}' \in \mathbf{N}} \Pr (\mathbf{n}', t_{i+1} | \mathbf{n}, t_i) \cdot \overline{Q}_{\pi^*}(\mathbf{n}', \mathtt{rej}, t_{i+1})  \\
    + \Delta t \textstyle \sum_{k \in K} \displaystyle \big(\Lambda_k(t_i) \big)\exp(- \Delta t \textstyle \sum_{k \in K} \displaystyle \Lambda_k(t_{i}) ) \big) \cdot 
    \sum_{k \in K}  \frac{ \Lambda_k(t_i)}{ \sum_{k' \in K}\Lambda_{k'}(t_i)  }
    \sum_{\mathbf{n}'}\Pr (\mathbf{n}', t_{i+1} | \mathbf{n}, t_i) \Big[ \overline{Q}_{\pi^*}(\mathbf{n}', \mathtt{rej}, t_{i+1})   \\
    +  \Pr\big(\mathtt{fin}(k)  \mid t_{i+1} \big) \cdot  \phi_{k}(\overline{p}^*_{cr}(\mathbf{n}, k, t_{i+1}))  \Big] - \delta_{max} (t_{i+1})  \\
    = \mathtt{Lower Bound}\big(Q_{\pi^*}(\mathbf{n}, \mathtt{rej}, t_i) \big) - \delta_{max} (t_{i+1}) 
  \end{multline}

\subsection{Induction Argument}
We are now ready to complete the proof by induction. The error function is

\begin{multline}
    \delta_{max}(t_i) = \max_{\mathbf{n} \in \mathbf{N}} \Big[ \overline{Q}_{\pi^*}(\mathbf{n}, \mathtt{rej}, t_i) - Q_{\pi^*}(\mathbf{n} , \mathtt{rej}, t_i) \Big] \leq \\
    \max_{\mathbf{n} \in \mathbf{N}} \Big[ \mathtt{Upper Bound}\big(\overline{Q}_{\pi^*}(\mathbf{n}, \mathtt{rej}, t_i) \big) - \mathtt{Lower Bound}\big(Q_{\pi^*}(\mathbf{n}, \mathtt{rej}, t_i) \big) + \delta_{max} (t_{i+1})   \Big]
\end{multline}

\begin{equation}
    \delta_{max}(t_i) - \delta_{max} (t_{i+1}) \leq \max_{\mathbf{n} \in \mathbf{N}} \Big[ \mathtt{Upper Bound}\big(\overline{Q}_{\pi^*}(\mathbf{n}, \mathtt{rej}, t_i) \big) - \mathtt{Lower Bound}\big(Q_{\pi^*}(\mathbf{n}, \mathtt{rej}, t_i) \big)    \Big]
\end{equation}

At the time horizon, $t_H$, $\overline{Q}_{\pi^*}(\mathbf{n}, \mathtt{rej}, t_H) = Q_{\pi^*}(\mathbf{n}, \mathtt{rej}, t_H) = 0$, for all $\mathbf{n} \in \mathbf{N}$. Therefore, $\delta_{max} (t_H) = 0$. Then the maximum value for $\delta_{max}(t_i)$ at any $t_i \in D$ is given by

\begin{equation}
    \delta_{max}(t_i) \leq \frac{t_H}{\Delta t} \max_{t_i \in D} \max_{\mathbf{n} \in \mathbf{N}} \Big[ \mathtt{Upper Bound}\big(\overline{Q}_{\pi^*}(\mathbf{n}, \mathtt{rej}, t_i) \big) - \mathtt{Lower Bound}\big(Q_{\pi^*}(\mathbf{n}, \mathtt{rej}, t_i) \big)   \Big], \textnormal{\hspace{2pt} for all } t_i \in D 
\end{equation}

We are interested in the limit behaviour as $\Delta t$ approaches zero. We substitute back in the full expressions for $\mathtt{Upper Bound}\big(\overline{Q}_{\pi^*}(\mathbf{n}, \mathtt{rej}, t_i) \big)$ and $\mathtt{Lower Bound}\big(Q_{\pi^*}(\mathbf{n}, \mathtt{rej}, t_i) \big)$.

\begin{multline}
    \lim_{\Delta t \rightarrow 0^+} \delta_{max}(t_i) \leq
    \lim_{\Delta t \rightarrow 0^+} \frac{t_H}{\Delta t} \max_{t_i \in D} \max_{n_k} \Bigg\{ \exp \big( - \Delta t \textstyle \sum_{k \in K} \displaystyle [\Lambda_k(t_i) - C_k \Delta t] \big) 
    \cdot \sum_{\mathbf{n}' \in \mathbf{N}} \Pr (\mathbf{n}', t_{i+1} | \mathbf{n}, t_i) \cdot \overline{Q}_{\pi^*}(\mathbf{n}', \mathtt{rej}, t_{i+1})\\
    +  \Delta t \textstyle \sum_{k \in K} \displaystyle \big[ \Lambda_k(t_i) + C_k \Delta t \big] \exp \big( - \Delta t \textstyle \sum_{k \in K} \displaystyle \big[ \Lambda_k(t_i) - C_k \Delta t \big] \big) \cdot 
    \sum_{k \in K}  \frac{ [ \Lambda_k(t_i) + C_k \Delta t ]}{ \sum_{k' \in K}[ \Lambda_{k'}(t_i) - C_{k'} \Delta t ]} \cdot \\
    \Bigg[ \sum_{\mathbf{n}' \in \mathbf{N}} \Pr (\mathbf{n}', t_{i+1} \mid \mathbf{n}, t_i) \cdot  \overline{Q}_{\pi^*}(\mathbf{n}', \mathtt{rej}, t_{i+1}) +  
   \sum_{\mathbf{n}' \in \mathbf{N}}  \Big( \Pr(\mathbf{n}', t_{i+1} \mid \mathbf{n}, t_i) + N_{serv} !^{|K|} \big(1 - \exp(-\mu_k N_{serv} |K| \Delta t) \big) \Big) \cdot \\
    \Big(\Pr \big(\mathtt{fin}(k)  \mid t_{i+1} \big) + \big(1 - \exp(-\Delta t \cdot \mu_{k}) \big) \Big)   \cdot 
    \Big(\phi(\overline{p}_{cr}^*(\mathbf{n}', k, t_{i+1})) + M\Delta t \Big)  \Bigg] \\
    + \sum_{N_{\mathtt{arr}} = 2}^\infty \Big( \big(  \Delta t \textstyle \sum_{k \in K} [\Lambda_k(t_i) + C_k \Delta t] \big)^{N_{\mathtt{arr}}} \exp \big( - \Delta t \sum_{k \in K}[\Lambda_k(t_i) - C_k \Delta t] \big) \cdot \big( N_{\mathtt{arr}} \cdot \max_{k \in K}\mathbb{E}[p_k] + \overline{Q}_{\pi^*}\big(\mathbf{n}, \mathtt{rej}, t_i) \big) \Big) \\
    - \exp(- \Delta t \textstyle \sum_{k \in K}  \displaystyle \Lambda_k(t_{i}) ) \cdot \sum_{\mathbf{n}' \in \mathbf{N}} \Pr (\mathbf{n}', t_{i+1} | \mathbf{n}, t_i) \cdot \overline{Q}_{\pi^*}(\mathbf{n}', \mathtt{rej}, t_{i+1})  \\
    - \Delta t \textstyle \sum_{k \in K} \displaystyle \big(\Lambda_k(t_i) \big)\exp(- \Delta t \textstyle \sum_{k \in K} \displaystyle \Lambda_k(t_{i}) ) \big) \cdot 
    \sum_{k \in K}  \frac{ \Lambda_k(t_i)}{ \sum_{k' \in K}\Lambda_{k'}(t_i)  }
    \sum_{\mathbf{n}'}\Pr (\mathbf{n}', t_{i+1} | \mathbf{n}, t_i) \Big[ \overline{Q}_{\pi^*}(\mathbf{n}', \mathtt{rej}, t_{i+1})   \\
    +  \Pr\big(\mathtt{fin}(k)  \mid t_{i+1} \big) \cdot  \phi_{k}(\overline{p}^*_{cr}(\mathbf{n}, k, t_{i+1}))  \Big] \Bigg\}
\end{multline}

Applying the sum and product laws of limits and rearranging terms yields
\begin{multline}
    \lim_{\Delta t \rightarrow 0^+} \delta_{max}(t) \leq 
    \lim_{\Delta t \rightarrow 0^+} \frac{t_H}{\Delta t} \max_{t_i \in D} \max_{\mathbf{n} \in \mathbf{N}} \Bigg[ \Big( \exp \big( - \Delta t \textstyle \sum_{k \in K} \displaystyle [\Lambda_k(t_i) - C_k \Delta t] \big) 
    - \exp(- \Delta t \textstyle \sum_{k \in K}  \displaystyle \Lambda_k(t_{i}) ) \Big) \cdot\\
     \sum_{\mathbf{n}' \in \mathbf{N}} \Pr (\mathbf{n}', t_{i+1} | \mathbf{n}, t_i) \cdot \overline{Q}_{\pi^*}(\mathbf{n}', \mathtt{rej}, t_{i+1}) \Bigg] \\
    + \lim_{\Delta t \rightarrow 0^+} t_H \max_{t_i \in D} \max_{\mathbf{n}} \Bigg[ \Big(\textstyle \sum_{k \in K} \displaystyle \big[ \Lambda_k(t_i) + C_k \Delta t \big] \exp \big( - \Delta t \textstyle \sum_{k \in K} \displaystyle \big[ \Lambda_k(t_i) - C_k \Delta t \big] \big) -  \textstyle \sum_{k \in K} \displaystyle \big(\Lambda_k(t_i) \big)\exp(- \Delta t \textstyle \sum_{k \in K} \displaystyle \Lambda_k(t_{i}) ) \big) \Big) \cdot \\
    \sum_{k \in K}  \frac{ \Lambda_k(t_i)}{ \sum_{k' \in K}\Lambda_{k'}(t_i)  }
    \sum_{\mathbf{n}'}\Pr (\mathbf{n}', t_{i+1} | \mathbf{n}, t_i) \Big[ \overline{Q}_{\pi^*}(\mathbf{n}', \mathtt{rej}, t_{i+1})   
    +  \Pr\big(\mathtt{fin}(k)  \mid t_{i+1} \big) \cdot  \phi_{k}(\overline{p}^*_{cr}(\mathbf{n}, k, t_{i+1}))  \Big] \\
    =  \lim_{\Delta t \rightarrow 0^+} t_H \max_{t_i \in D} \max_{\mathbf{n}} \Bigg[ \frac{1}{\Delta t} \cdot \frac{ \exp \big( \textstyle \sum_{k \in K} \displaystyle C_k \Delta t^2) -1 
    }{  \exp \big(\Delta t \textstyle \sum_{k \in K} \Lambda_k(t_{i}) \big) }
    \sum_{\mathbf{n}' \in \mathbf{N}} \Pr (\mathbf{n}', t_{i+1} | \mathbf{n}, t_i) \cdot \overline{Q}_{\pi^*}(\mathbf{n}', \mathtt{rej}, t_{i+1}) \Bigg]
\end{multline}

Applying  L'H$\hat{\textnormal{o}}$pital's rule we have
\begin{multline}
    \lim_{\Delta t \rightarrow 0^+} \delta_{max}(t_i) \leq \\
    \lim_{\Delta t \rightarrow 0^+} t_H \max_{t_i \in D} \max_{\mathbf{n}} \Bigg[ \frac{2 \Delta t \exp \big(\textstyle \sum_{k \in K} \displaystyle C_k  \Delta t^2)  \textstyle \sum_{k \in K} \displaystyle C_k 
    }{ \exp(\textstyle \Delta t \sum_{k \in K} \displaystyle \Lambda_k(t_{i}) ) + \Delta t \exp(\textstyle \Delta t \sum_{k \in K} \displaystyle\Lambda_k(t_{i})) \textstyle \sum_{k \in K} \displaystyle \Lambda_k(t_{i})}
    \cdot \sum_{\mathbf{n}'} \Pr (\mathbf{n}', t_{i+1} | \mathbf{n}, t_i) \cdot \overline{Q}_{\pi^*}(\mathbf{n}', \mathtt{rej}, t_{i+1}) \Bigg] \\
    = 0 \textnormal {, for all }t_i \in D
\end{multline}

We now consider the error in the critical values, $\epsilon_{max}(t_i)$. We have that $\epsilon_{max}(t_H) = 0$ by definition. For any $t_i < t_H$

\begin{multline}
    \overline{p}^*_{cr}(\mathbf{n}, k, t_i) - p^*_{cr}(\mathbf{n}, k, t_i) = \frac{1}{\Pr \big(\mathtt{fin}(k)  \mid t_i \big)} \Big( \overline{Q}_{\pi^*}(\mathbf{n}, \mathtt{rej}, t_i) - \overline{Q}_{\pi^*}(\mathbf{x}^{\mathtt{acc}}_{(\mathbf{n}, k)}[\mathbf{n}], \mathtt{rej}, t_i) \\
    - Q_{\pi^*}(\mathbf{n}, \mathtt{rej}, t_i)  + Q_{\pi^*}(\mathbf{x}^{\mathtt{acc}}_{(\mathbf{n}, k)}[\mathbf{n}], \mathtt{rej}, t_i)\Big)
\end{multline}

\begin{multline}
    \lim_{\Delta t \rightarrow 0^+} \Big( \overline{p}^*_{cr}(\mathbf{n}, k, t_i) - p^*_{cr}(\mathbf{n}, k, t_i) \Big) = \\
    \frac{1}{\Pr \big(\mathtt{fin}(k)  \mid t_i \big)} \Big( \lim_{\Delta t \rightarrow 0^+}  \big[ \overline{Q}_{\pi^*}(\mathbf{n}, \mathtt{rej}, t_i) - Q_{\pi^*}(\mathbf{n}, \mathtt{rej}, t_i) \big] + \\
    \lim_{\Delta t \rightarrow 0^+}  \big[ Q_{\pi^*}(\mathbf{x}^{\mathtt{acc}}_{(\mathbf{n}, k)}[\mathbf{n}], \mathtt{rej}, t_i) - \overline{Q}_{\pi^*}(\mathbf{x}^{\mathtt{acc}}_{(\mathbf{n}, k)}[\mathbf{n}], \mathtt{rej}, t_i) \big] \Big)\leq \\
    \frac{2}{\Pr \big(\mathtt{fin}(k)  \mid t_i \big)} \cdot \lim_{\Delta t \rightarrow 0^+} \delta_{max}(t_i)  = 0
\end{multline}

Therefore $\lim_{\Delta t \rightarrow 0^+} \epsilon_{max}(t_i) = 0$ for all $t_i \in D$.
Finally, by the definitions of $\overline{V}_{\pi^*}(\mathbf{x}, t)$ and $V_{\pi^*}(\mathbf{x}, t_i)$,  $\lim_{\Delta t \rightarrow 0^+} \delta_{max}(t_i) = 0$ and $\lim_{\Delta t \rightarrow 0^+} \epsilon_{max}(t_i) = 0$ for all $t_i \in D $ implies that $\lim_{\Delta t \rightarrow 0^+} V(\mathbf{x}, t_i) = \overline{V}(\mathbf{x}, t_i)$ for all $t_i \in D $, which completes the proof. \qed

\section{Proof of Proposition~\ref{prop:5}}
\abstractionerr*
\vspace{-4mm}
\begin{equation*}
\forall_{a \in A} \mid Q_{\pi^*} (\mathbf{x}_1, a, t) - Q_{\pi^*} (\mathbf{x}_2, a, t) \mid\ \leq \epsilon
\end{equation*}
    
\bigskip
\textit{Proof}: We must show that this holds for both the $\mathtt{acc}$ and $\mathtt{rej}$ action. It is immediately clear that this holds for the $\mathtt{rej}$ action due to the assumption that ${\mid Q_{\pi^*}(\mathbf{x}_1[\mathbf{n}], \mathtt{rej}, t) - Q_{\pi^*}(\mathbf{x}_2[\mathbf{n}], \mathtt{rej}, t) \mid \leq \epsilon / 2}$. For the $\mathtt{acc}$ action, by substituting Equation~\ref{eq:q_val_relation} we have that
\begin{multline}
\mid Q_{\pi^*} (\mathbf{x}_1, \mathtt{acc}, t) - Q_{\pi^*} (\mathbf{x}_2, \mathtt{acc}, t) \mid = \\
\Bigm\lvert \Big\{ \Pr\big(\mathtt{fin}(\mathbf{x}_1[{k^+}]) \mid t \big) \cdot \big(\mathbf{x}_1[p] - p^*_{cr}(\mathbf{x}_1[\mathbf{n}], \mathbf{x}_1[{k^+}], t) \big) 
     + Q_{\pi^*}(\mathbf{x}_1[\mathbf{n}], \mathtt{rej}, t) \Big\} - \\
     \Big\{ \Pr\big(\mathtt{fin}(\mathbf{x}_2[{k^+}]) \mid t \big) \cdot \big(\mathbf{x}_2[p] - p^*_{cr}(\mathbf{x}_2[\mathbf{n}], \mathbf{x}_2[{k^+}], t) \big) 
     + Q_{\pi^*}(\mathbf{x}_2[\mathbf{n}], \mathtt{rej}, t) \Big\}  \Bigm\lvert \\
\end{multline}

Because $\mathbf{x}_1[k^+] = \mathbf{x}_2[k^+] = k^+$ and $\mathbf{x}_1[p] = \mathbf{x}_2[p]$, 
\begin{multline}
    \mid Q_{\pi^*} (\mathbf{x}_1, \mathtt{acc}, t) - Q_{\pi^*} (\mathbf{x}_2, \mathtt{acc}, t) \mid  = \\
    \bigm\lvert \Pr\big(\mathtt{fin}(k^+) \mid t \big) \cdot \big(p^*_{cr}(\mathbf{x}_2[\mathbf{n}], k^+, t) -  p^*_{cr}(\mathbf{x}_1[\mathbf{n}], k^+, t) \big)  \big) + \\ Q_{\pi^*}(\mathbf{x}_1[\mathbf{n}], \mathtt{rej}, t) - Q_{\pi^*}(\mathbf{x}_2[\mathbf{n}], \mathtt{rej}, t)  \bigm\lvert 
    \ \leq \Pr\big(\mathtt{fin}(k^+) \mid t \big) \cdot \frac{\epsilon}{2} + \frac{\epsilon}{2} \leq \epsilon \hspace{10pt} \square
\end{multline}

\section{Proof of Proposition~\ref{prop:6}}

\abstractvalue*
\noindent \textit{where the critical price function is}
\begin{equation*}
    \widehat{p}^*_{cr}(n_A, {k^+}, t) =  \frac{1}{\Pr(\mathtt{fin}({k^+}) \mid t) } \times 
    \Big(\widehat{Q}_{\pi^*}(n_A, \mathtt{rej}, t)  -   \sum_{\mathbf{n} \in \psi_{\mathbf{n}}^{-1}(n_A)}  \hspace{-3mm} w(\mathbf{n}) \cdot \widehat{Q}_{\pi^*}(\psi(\mathbf{x}^{\mathtt{acc}}_{(\mathbf{n}, k^+)}), \mathtt{rej}, t)
    \Big). 
\end{equation*}
\noindent \textit{if $\mathbf{x}[{k^+}] \neq \bot$ and $t < t_H$, and 0 otherwise. } 

\bigskip

\textit{Proof}: To prove Proposition~\ref{prop:6}, we must show that $ \widehat{Q}_{\pi^*}(\widehat{\mathbf{x}}, \mathtt{acc}, t) \geq  \widehat{Q}_{\pi^*}(\widehat{\mathbf{x}}, \mathtt{rej}, t)$ if and only if $\widehat{\mathbf{x}}[p] \geq \widehat{p}^*_{cr}(\widehat{\mathbf{x}}[n_A], \widehat{\mathbf{x}}[{k^+}], t)$. The proof proceeds in the same manner as the proofs of Propositions~\ref{prop:pwl_vf} and~\ref{prop:thr_pol}.

In the case that a task is accepted, we adapt Equation~\ref{eq:qv_accept_act} to compute the $Q$-value as follows
\begin{multline}
    \widehat{Q}_{\pi^*}(\widehat{\mathbf{x}}, \mathtt{acc}, t_i) = \widehat{\mathbf{x}}[p] \cdot \Pr(\mathtt{fin}(\widehat{\mathbf{x}}[{k^+}] \mid t_i) + 
    \sum_{\mathbf{n} \in \psi_{\mathbf{n}}^{-1}(n_A)}   w(\mathbf{n}) \cdot \widehat{V}_{\pi^*}(\psi(\mathbf{x}^{\mathtt{acc}}_{(\mathbf{n}, \widehat{\mathbf{x}}[{k^+}])}), t_{i}) \\
    = \widehat{\mathbf{x}}[p] \cdot \Pr(\mathtt{fin}(\mathbf{x}[{k^+}]) \mid t) +
    \sum_{\mathbf{n} \in \psi_{\mathbf{n}}^{-1}(n_A)}  \hspace{-3mm} w(\mathbf{n}) \cdot \widehat{Q}_{\pi^*}(\psi(\mathbf{x}^{\mathtt{acc}}_{(\mathbf{n}, \widehat{\mathbf{x}}[{k^+}])}), \mathtt{rej}, t)
\end{multline}

\noindent where $\mathbf{x}^{\mathtt{acc}}_{(\mathbf{n}, k)}$ was defined previously in Equation~\ref{eq:x_accept}, and the second equality is because $\mathbf{x}^{\mathtt{acc}}_{(\mathbf{n}, \mathbf{x}[{k^+}])}[{k^+}] = \bot$ so only the reject action is enabled. We add and subtract $\widehat{p}^*_{cr}(\widehat{\mathbf{x}}[n_A], \widehat{\mathbf{x}}[{k^+}], t) $ in the previous expression to arrive at
\begin{multline}
    \widehat{Q}_{\pi^*}(\widehat{\mathbf{x}}, \mathtt{acc}, t) =  \Pr(\mathtt{fin}(\widehat{\mathbf{x}}[{k^+}]) \mid t) \cdot \big(\widehat{\mathbf{x}}[p] + \widehat{p}^*_{cr}(\widehat{\mathbf{x}}[n_A], \widehat{\mathbf{x}}[k^+], t) - \widehat{p}^*_{cr}(\widehat{\mathbf{x}}[n_A], \widehat{\mathbf{x}}[k^+], t) \big)+ \\
    \sum_{\mathbf{n} \in \psi_{\mathbf{n}}^{-1}(n_A)}  \hspace{-3mm} w(\mathbf{n}) \cdot \widehat{Q}_{\pi^*}(\psi(\mathbf{x}^{\mathtt{acc}}_{(\mathbf{n}, \widehat{\mathbf{x}}[{k^+}])}), \mathtt{rej}, t).
\end{multline}

Substituting in the definition of $\widehat{p}^*_{cr}(\widehat{\mathbf{x}}[n_A], \widehat{\mathbf{x}}[{k^+}], t)$ we have
\begin{equation}
    \label{eq:result_abstract_pol}
    \widehat{Q}_{\pi^*}(\widehat{\mathbf{x}}, \mathtt{acc}, t) =  \Pr(\mathtt{fin}(\mathbf{x}[{k^+}]) \mid t) \cdot \big(\widehat{\mathbf{x}}[p]- \widehat{p}^*_{cr}(\widehat{\mathbf{x}}[n_A], \widehat{\mathbf{x}}[{k^+}], t) \big)+
    \widehat{Q}_{\pi^*}(\widehat{\mathbf{x}}[n_A], \mathtt{rej}, t)
\end{equation}
From Equation~\ref{eq:result_abstract_pol}, we have the desired result that $ \widehat{Q}_{\pi^*}(\widehat{\mathbf{x}}, \mathtt{acc}, t) \geq  \widehat{Q}_{\pi^*}(\widehat{\mathbf{x}}, \mathtt{rej}, t)$ if and only if $\widehat{\mathbf{x}}[p] \geq \widehat{p}^*_{cr}(\widehat{\mathbf{x}}[n_A], \widehat{\mathbf{x}}[{k^+}], t)$, thereby proving Proposition~\ref{prop:6}. $\square$

\section{State Abstraction Details}
In our state abstraction approach, we propose two approaches for defining the state aggregation function. In the first approach, we propose to solve an approximation of Problem~\ref{prob:continuous_time}, which we refer to as the~\emph{stationary problem}. The stationary problem considers constant arrival rates and an infinite horizon. We use the solution to the stationary problem to aggregate similar combinations of task classes, as a proxy to the solution for the STA-HDMP as suggested by Proposition~\ref{prop:5}. In the second approach, we aggregate states based on summary statistics for the distribution over task completion times. We provide details for both of these approaches here.

\subsection{State Aggregation via Stationary Solution}
We first compute the average arrival rate for each task class over the horizon, $\mathbb{E}_t[\Lambda_k(t)]$. For the stationary problem, we assume that the arrival rate for each task is always $\mathbb{E}_t[\Lambda_k(t)]$. Additionally, we assume that we are optimising the average reward over an infinite horizon. Due to the constant dynamics and infinite horizon, the optimal solution is stationary, i.e. does not depend on time, and therefore is much easier to solve than the time-varying and finite horizon problem given by Problem~\ref{prob:continuous_time}. To arrive at a solution algorithm for the stationary problem, we formulate the stationary problem as a~\emph{semi-MDP}~\cite{puterman2014markov}, a common model for queuing systems.

\subsubsection{Semi-MDPs}
Here we give a brief introduction to semi-MDPs. For a more rigorous introduction, see~\citet{puterman2014markov} (Chapter 11). In a semi-MDP, the decision maker is allowed to choose which action to apply at each decision epoch. After applying the action, the state evolves to a successor state at the next decision epoch. However, unlike a discrete-time MDP, in a semi-MDP the time between each decision epoch is randomly distributed. In queuing problems with admission control, such as the one we address, each decision epoch corresponds to each time a task arrives.

A semi-MDP with stationary dynamics and infinite horizon can be defined by the tuple, $\mathcal{M}^{s} = (\mathbf{X}, A, \mathcal{P}, \mathcal{R}, \mathcal{G})$. $\mathbf{X}$ is the state space, and $A$ is the action space. $\mathcal{R}(\mathbf{x}, a)$ is the reward for applying action $a$ in state $\mathbf{x}$.
$\mathcal{P}(\mathbf{x}' \mid \mathbf{x}, a)$ is the probability that the state at the next decision epoch will be $\mathbf{x}'$ given that the current state is $\mathbf{x}$ and action $a$ is applied.
$\mathcal{G}(t \mid \mathbf{x}, a)$ is the probability that the next decision epoch will occur within duration $t$ of the current decision epoch, given that the current state is $\mathbf{x}$ and action $a$ is applied.

It is possible to convert a semi-MDP to an equivalent discrete-time infinite horizon MDP which has the same optimal value function using the method introduced by~\citet{schweitzer1971iterative}, and described in~\citet{puterman2014markov} (Chapter 11.4). We denote the quantities in the transformed model with ``$\sim$". 
Let $\widetilde{\mathbf{X}} = \mathbf{X}$, $\widetilde{A} = A$, 
\begin{equation}
    \label{eq:conversion_start}
    \widetilde{\mathcal{R}}(\mathbf{x}, a) = \mathcal{R}(\mathbf{x}, a)/y(\mathbf{x}, a),
\end{equation}

\begin{equation}
    \widetilde{\mathcal{P}}(\mathbf{x}' \mid \mathbf{x}, a) = 
    \begin{cases}
        \eta \mathcal{P}(\mathbf{x}' \mid \mathbf{x}, a)/y(\mathbf{x}, a), \textnormal{ if } \mathbf{x}' \neq \mathbf{x} \\
        1 + \eta \big[\mathcal{P}(\mathbf{x} \mid \mathbf{x}, a) - 1 \big]/y(\mathbf{x}, a), \textnormal{ if } \mathbf{x}' = \mathbf{x}
    \end{cases}
\end{equation}
\noindent where $y(\mathbf{x}, a)$ is the expected length of time until the next decision epoch, given that action $a$ is applied at state $\mathbf{x}$ at the current decision epoch. $\eta$ is a constant which satisfies
\begin{equation}
    \label{eq:conversion_final}
    0 < \eta < y(\mathbf{x}, a) \textnormal{ for all } a \in A \textnormal{ and } \mathbf{x} \in \mathbf{X},
\end{equation}

\noindent thereby ensuring that $\widetilde{\mathcal{P}}$ outputs valid probabilities.

\subsubsection{Stationary Problem as a Semi-MDP}
We now explain how the functions required for the Semi-MDP can be computed for the stationary version of Problem~\ref{prob:continuous_time} which we aim to solve. We will use the same notation for states as was used for the STA-HMDP. To simplify the presentation, we will assume that a decision epoch occurs whenever a new task arrives, or a task finishes being processed. At decision epochs where a new task has arrived, the decision-maker may play the $\mathtt{acc}$ or $\mathtt{rej}$ action to decide whether to accept the task. At decision epochs where a task has finished being processed, the decision maker is not able to accept a new task. Taking this perspective allows us to write the required equations in a simpler format.

First, the expected time until the next decision epoch is $y(\mathbf{x}, a) = 1/{\Lambda_{total}}(\mathbf{x})$, where $\Lambda_{total}(\mathbf{x})$ is the total rate at which tasks are arriving and being processed
\begin{equation}
    \Lambda_{total}(\mathbf{x}) = \sum_{k} \mathbb{E}_t[\Lambda_k(t)] + \sum_{k} \mathbf{x}[n_k] \cdot \mu_k.
\end{equation}

We set $\eta < \min_{\mathbf{x}} \big[ 1/ \Lambda_{total}(\mathbf{x}) \big]$. The transition dynamics for the state at the next decision epoch are

\begin{equation}
    \label{eq:cases_semi_mdp}
    \mathcal{P}(\mathbf{x}' \mid \mathbf{x}, a) = 
    \begin{cases}
        \frac{\mu_k \cdot \mathbf{x}[n_k]}{\Lambda_{total}(\mathbf{x})} f_\bot (\mathbf{x}'[p]),  \textnormal{  \ if } \mathbf{x}'[k^+] = \bot, \mathbf{x}'[n_k] = \mathbf{x}[n_k] - 1, \textnormal{ and } \mathbf{x}'[n_j] = \mathbf{x}[n_j]\ \forall_{j \neq k} \vspace{2mm} \\
        \frac{\mathbb{E}_t[\Lambda_{\mathbf{x}'[k^+]}(t)]}{\Lambda_{total}(\mathbf{x})} f_{\mathbf{x}'[k^+]}(\mathbf{x}'[p]), \textnormal{ \ if } \mathbf{x}'[k^+] \neq \bot \textnormal{ and } \mathbf{x}'[n_k] = \mathbf{x}[n_k]\ \forall_{k} \vspace{2mm} \\
        0, \textnormal{ otherwise}
    \end{cases}
\end{equation}

The first case in Equation~\ref{eq:cases_semi_mdp} considers when the next decision epoch occurs due to a task finishing being processed. The second case considers when the next decision epoch occurs due to a task arriving. The reward function for the stationary problem which assumes an infinite horizon is 

\begin{equation}
    \mathcal{R}(\mathbf{x}, \mathtt{acc}) = \mathbf{x}[p].
\end{equation}

In accordance with the previous subsection (Equations~\ref{eq:conversion_start} - \ref{eq:conversion_final}), we can now compute the reward and transition function for the equivalent discrete-time MDP with infinite horizon, $\widetilde{\mathcal{R}}(\mathbf{x}, a)$ and $ \widetilde{\mathcal{P}}(\mathbf{x}' \mid \mathbf{x}, a)$ respectively.
For the resulting discrete-time MDP, we adapt the
 Modifed Policy Iteration Algorithm (MPI) for the infinite horizon average reward criterion, which is described in~\citet{puterman2014markov}, Chapter 8.7.1. 
This enables us to compute the optimal critical price function for the stationary problem, $\widetilde{p}^*_{cr}(\mathbf{n}, k)$. 

In the infinite horizon average reward setting, the value function is infinite. 
Therefore we use the MPI to compute the optimal~\emph{relative} value function, and the associated relative $Q$-values for the $\mathtt{rej}$ action, $\widetilde{Q}_{\pi^*}(\mathbf{n}, \mathtt{rej})$. 
The relative value function is also referred to as the~\emph{bias} function (see~\cite{puterman2014markov}, Chapter 8.2.1.), and represents the difference in value between states.

\subsubsection{State Aggregation Based on Stationary Solution}
For each $\mathbf{n}\in \mathbf{N}$, we then compose a vector $v_\mathbf{n}$ of the associated relative $Q$-value and critical price values from the solution to the stationary problem:  $$v_{\mathbf{n}}^{ss} = [ \widetilde{Q}_{\pi^*}(\mathbf{n}, \mathtt{rej}), \widetilde{p}^*_{cr}(\mathbf{n}, k_1), \ldots, \widetilde{p}^*_{cr}(\mathbf{n}, k_2) ]$$

We then performing clustering on the $v_\mathbf{n}^{ss}$ to aggregate the task combinations into the desired number of abstrations, $|\mathcal{N}_A|$. These clusters define aggregation function, $\psi_\mathbf{n}$. For the clustering, we use the k-means algorithm using 20 random initialisations of the centroids.

\subsection{State Aggregation via Order Statistics}
The scalability of performing the state aggregation based on the stationary solution is still limited as it requires solving for the stationary solution which may not be feasible for a large number of task classes. Here we present an alternative state aggregation approach, based on summary statistics for each combination of tasks. For each $\mathbf{n} \in \mathbf{N}$, we compose a vector of the form 
$$v_{\mathbf{n}}^{os} = \\ \big[ \mathbb{E}[t_{N_{f} \geq 0} \mid \mathbf{n}], \mathbb{E}[t_{N_{f} \geq 1} \mid \mathbf{n}], \ldots, \mathbb{E}[t_{N_{f} = N_{serv}} \mid \mathbf{n}] \big]$$

\noindent where $N_{f}$ denotes the number of servers free, i.e. $N_f = N_{serv} - \sum_{k\in K} n_k$. Additionally, $\mathbb{E}[t_{N_{f} \geq q} \mid \mathbf{n}]$ denotes the expected time until at least $q$ servers are free given that the combination of tasks currently being processed is $\mathbf{n}$, and assuming that no further tasks are accepted. The intuition for $v_{\mathbf{n}}^{os}$ is that it approximately summarises the distribution over task completion times for the tasks currently being processed. 

Computing $v_{\mathbf{n}}^{os}$ requires computing the mean of order statistics of independent and non-identical exponential distributions. This is computationally challenging to compute exactly. Therefore, we use a Monte Carlo approximation to estimate each value of $\mathbb{E}[t_{N_{f} \geq q} \mid \mathbf{n}]$.

For each $\mathbf{n} \in \mathbf{N}$ we use the following approach to compute the Monte Carlo approximation of $v_{\mathbf{n}}^{os}$. For each server in combination $\mathbf{n}$, we randomly sample the amount of time, $t_{\mathtt{fin}}$, for each server to complete the task they are currently processing. We sample these values from the exponential distributions for the appropriate task class. If a server is not processing a task, $t_{\mathtt{fin}} = 0$.
We then sort the sampled processing times for each server in ascending order.
The resulting sorted vector, $v_{\mathbf{n}}^{sample}$, is a sample of the form

$$v_{\mathbf{n}}^{sample} = \\ \big[ (t_{N_{f} \geq 0} \mid \mathbf{n}), (t_{N_{f} \geq 1} \mid \mathbf{n}), \ldots, (t_{N_{f} = N_{serv}} \mid \mathbf{n}) \big]$$

To arrive at an estimate of the expected value of each element, as required by $v_{\mathbf{n}}^{os}$, we sample many random vectors $v_{\mathbf{n}}^{sample}$, and compute the element-wise mean. 

In our implementation, we use the mean of 5000 samples of $v_{\mathbf{n}}^{sample}$ to estimate $v_{\mathbf{n}}^{os}$. Once we have computed the estimate of $v_{\mathbf{n}}^{os}$ for each $\mathbf{n} \in \mathbf{N}$, we perform clustering on these vectors to group the task combinations into the desired number of abstractions. 
For the clustering, we use the k-means algorithm using 20 random initialisations of the centroids.

\section{Value Iteration Algorithm for the Abstract STA-HMDP}
We refer to the abstract version of the STA-HDMP as the Abstract STA-HDMP. As mentioned, the states in the Abstract STA-MDP are comprised of the state variables $\widehat{\mathbf{x}} = \{n_A,k^+, p\}$. We denote by  $\widehat{V}_{\pi^*}(\widehat{\mathbf{x}}, t_i)$, and $\widehat{p}^*_{cr}(\widehat{\mathbf{x}}[n_A], \widehat{\mathbf{x}}[k^+], t)$ the optimal value and critical price respectively in the Abstract STA-HDMP. As in the original STA-HDMP, the $Q$-value for rejecting the task does not depend on the task class or price that was rejected. Therefore, we write $\widehat{Q}_{\pi^*}(n_A, \mathtt{rej}, t_i)$ for the value of rejecting the task in the Abstract STA-HDMP. The $Q$-value for the $\mathtt{rej}$ action can be computed as follows
\begin{multline}
    \label{eq:abstract_q_val}
    \widehat{Q}_{\pi^*}(n_A, \mathtt{rej}, t_i) =
    \sum_{\widehat{\mathbf{x}}'[n_A]}  \sum_{\widehat{\mathbf{x}}'[k^+]} \int_{\widehat{\mathbf{x}}'[p]} P(\widehat{\mathbf{x}}' \mid \widehat{\mathbf{x}}, \mathtt{rej}, t_i) \cdot \widehat{V}_{\pi^*}(\mathbf{x}', t_{i+1})\cdot \dif (\mathbf{x}'[p]) \\
    = \sum_{\widehat{\mathbf{x}}'[n_A]} \Pr(\widehat{\mathbf{x}}'[n_A] \mid  \widehat{\mathbf{x}}[n_A]) \cdot  \sum_{\widehat{\mathbf{x}}'[k^+]} \Pr(\widehat{\mathbf{x}}'[k^+]= {k^+}' \mid t_i) \cdot \\
    \Big[ \widehat{Q}_{\pi^*}(\widehat{\mathbf{x}}'[n_A], \mathtt{rej}, t_{i+1} ) +  \Pr\big(\mathtt{fin}(\widehat{\mathbf{x}}'[{k^+}]) \mid t_{i+1} \big) \cdot  \phi_{\widehat{\mathbf{x}}'[{k^+}]} \big( \widehat{p}^*_{cr}(\widehat{\mathbf{x}}'[\mathbf{n}], \widehat{\mathbf{x}}'[{k^+}], t_{i+1}) \big) \Big]
\end{multline}

\noindent where from Equation~\ref{eq:abstract_trans_probs} the transition probabilities between the abstractions of the task combinations is 
\begin{equation}
    \label{eq:abstr_probs}
    \Pr(\widehat{\mathbf{x}}'[n_A] \mid  \widehat{\mathbf{x}}[n_A]) = \\ \sum_{\mathbf{n} \in \psi_{\mathbf{n}}^{-1}(n_A)} \sum_{\mathbf{n}' \in \psi_{\mathbf{n}}^{-1}(n_A')}  w(\mathbf{n}) \cdot \prod_{k \in K} \Pr(\mathbf{n}'[n_k] \mid  \mathbf{n}[n_k]),
\end{equation}
\noindent and in Equation~\ref{eq:abstract_q_val} we use the simplification of the integral from Proposition~\ref{prop:simple_calcs}. The probabilities in the product in Equation~\ref{eq:abstr_probs} can be computed using Equation~\ref{eq:server_fin}.

In the case that a task is accepted, we adapt Equation~\ref{eq:qv_accept_act} to compute the $Q$-value as follows
\begin{equation}
    \label{eq:qv_accept_abstract}
    \widehat{Q}_{\pi^*}(\widehat{\mathbf{x}}, \mathtt{acc}, t_i) = \widehat{\mathbf{x}}[p] \cdot \Pr(\mathtt{fin}(\widehat{\mathbf{x}}[{k^+}] \mid t_i) + \\
    \sum_{\mathbf{n} \in \psi_{\mathbf{n}}^{-1}(n_A)}   w(\mathbf{n}) \cdot \widehat{V}_{\pi^*}(\psi(\mathbf{x}^{\mathtt{acc}}_{(\mathbf{n}, \widehat{\mathbf{x}}[{k^+}])}), t_{i}).
\end{equation}

\noindent where $\mathbf{x}^{\mathtt{acc}}_{(\mathbf{n}, k)}$ was defined previously in Equation~\ref{eq:x_accept}.

Like the original STA-HMDP, the optimal value function for the Abstract STA-HMDP has the piecewise-linear form given in Proposition~\ref{prop:pwl_vf}. The critical price function and optimal policy for the Abstract STA-HDMP can be computed according to Proposition~\ref{prop:6}.

\abstractvalue*
\noindent \textit{where the critical price function is}
\begin{equation}
    \label{eq:pcrit_abstract}
    \widehat{p}^*_{cr}(n_A, {k^+}, t) =  \frac{1}{\Pr(\mathtt{fin}({k^+}) \mid t) } \times 
    \Big(\widehat{Q}_{\pi^*}(n_A, \mathtt{rej}, t)  -   \sum_{\mathbf{n} \in \psi_{\mathbf{n}}^{-1}(n_A)}  \hspace{-3mm} w(\mathbf{n}) \cdot \widehat{Q}_{\pi^*}(\psi(\mathbf{x}^{\mathtt{acc}}_{(\mathbf{n}, {k^+})}), \mathtt{rej}, t)
    \Big). 
\end{equation}
\noindent \textit{if $\mathbf{x}[{k^+}] \neq \bot$ and $t < t_H$, and 0 otherwise. }

The value iteration algorithm for solving the Abstract STA-HDMP proceeds in a similar manner to Algorithm~\ref{alg:1}. Pseudocode is provided in Algorithm~\ref{alg:2}. The algorithm steps backwards over time. At each time step, it computes the $Q$-values for the $\mathtt{rej}$ action for each abstract combination, $n_A \in \mathcal{N}_A$, using Equation~\ref{eq:abstract_q_val}. Then, the critical price values for each abstract combination and task class are computed using Proposition~\ref{prop:6}. Once the policy has been computed, the policy to apply to the original STA-HMDP is derived using Equation~\ref{eq:abstract_to_ground}.

\begin{algorithm}
\caption{Value iteration for Abstract STA-HDMP \label{alg:2}}
\begin{algorithmic}
\State {initialise} $\widehat{Q}_{\pi^*}(n_A, \mathtt{rej}, t_H) = 0
\textnormal{ for all } n_A \in \mathcal{N}_A$
\For{ $t_i = t_{H-1}, t_{H-2}, \ldots t_0$}
    \For{$n_A \in \mathcal{N}_A$}
        \State{compute $\widehat{Q}_{\pi^*}(n_A, \mathtt{rej}, t_i)$ using Eq.~\ref{eq:abstract_q_val} }
    \EndFor
    \For{$n_A \in \mathcal{N}_A$}
        \For{$k \in K$}
            \State{compute $\widehat{p}^*_{cr}(n_A, k, t_i)$ using Eq.~\ref{eq:pcrit_abstract}}
        \EndFor
    \EndFor
\EndFor
     
\end{algorithmic}
\end{algorithm}

\FloatBarrier 
\newpage
\section{Additional Results}
\subsection{Mean Reward Performance for Synthetic Domain with Step Arrival Rate Functions}

\begin{figure}[h!tb]
    \centering
        \includegraphics[width=0.85\textwidth]{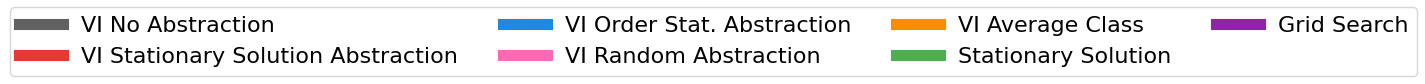}
    \vspace{-3mm}
\end{figure}
\begin{figure}[h!tb]
    \centering
    \begin{subfigure}[t]{0.35\textwidth}
        \centering
        \includegraphics[width=\textwidth]{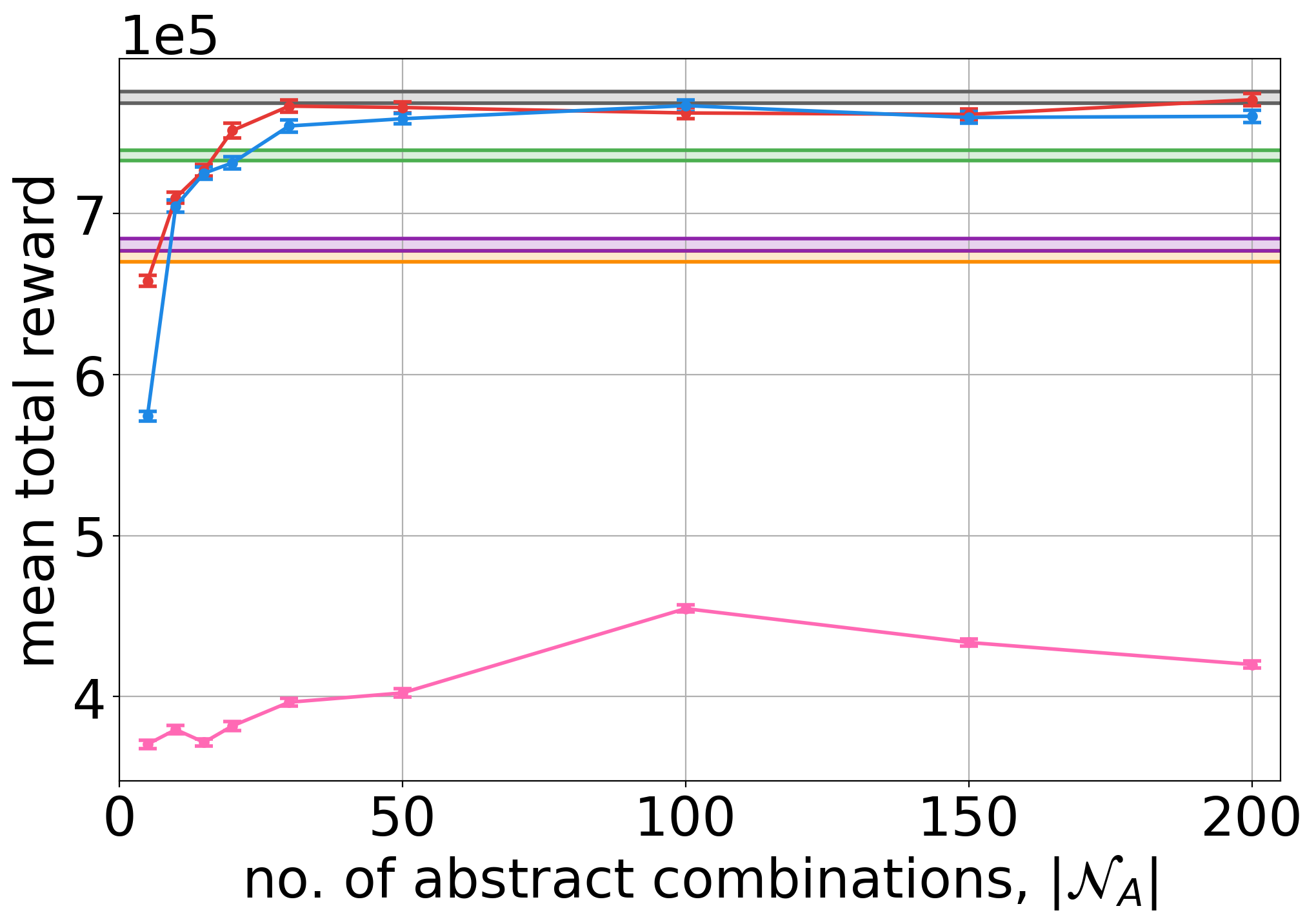}
        \caption{Synthetic Small (step arrival rates).}
    \end{subfigure}%
    ~
    \begin{subfigure}[t]{0.37\textwidth}
        \centering
        \includegraphics[width=\textwidth]{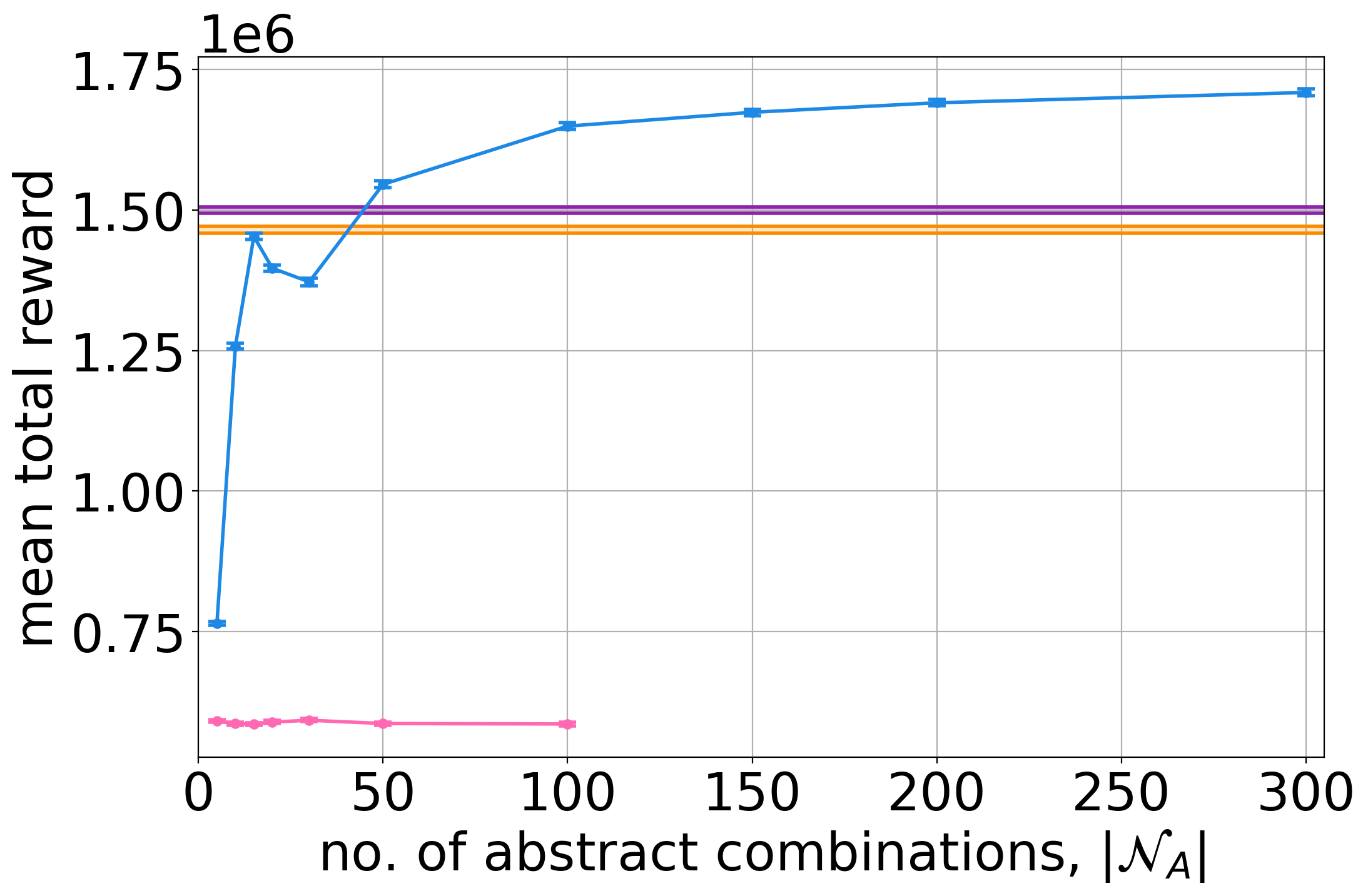}
        \caption{Synthetic Large (step arrival rates).}
    \end{subfigure}
    \caption{Mean total reward performance on 300 evaluation runs for each domain. Error bars and shaded regions indicate standard errors. We only include methods and numbers of abstract combinations which can be computed within 30,000 seconds. }
\end{figure}

\FloatBarrier

\subsection{Computation Times}
\begin{figure}[h!tb]
    \centering
        \includegraphics[width=0.85\textwidth]{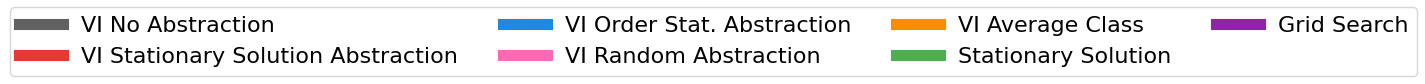}
    \vspace{-3mm}
\end{figure}

\begin{figure}[h!tb]
    \centering
    \begin{subfigure}[t]{0.33\textwidth}
        \centering
        \includegraphics[width=\textwidth]{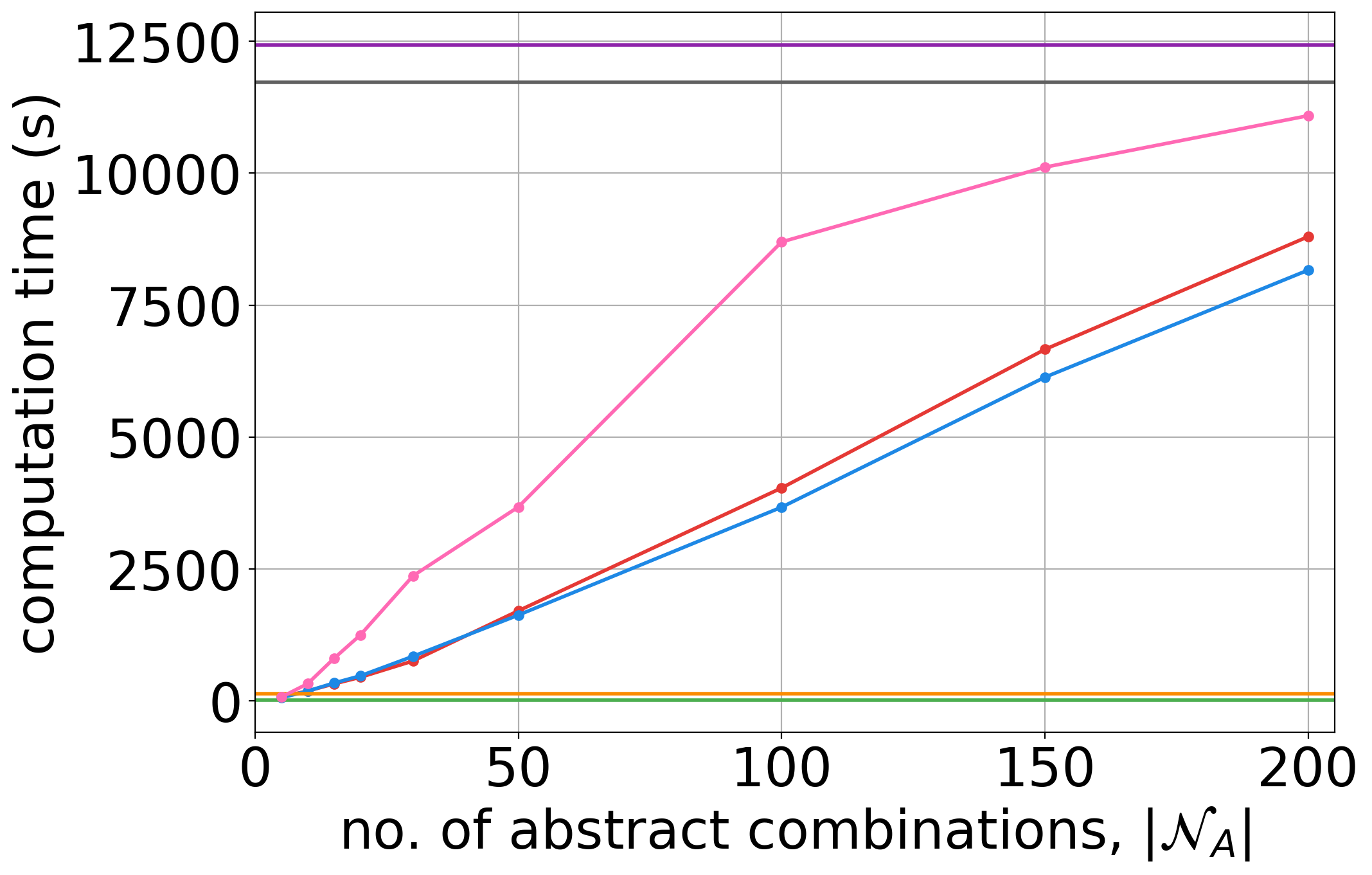}
        \caption{Synthetic Small (step arrival rates).}
    \end{subfigure}%
    ~
    \begin{subfigure}[t]{0.33\textwidth}
        \centering
        \includegraphics[width=\textwidth]{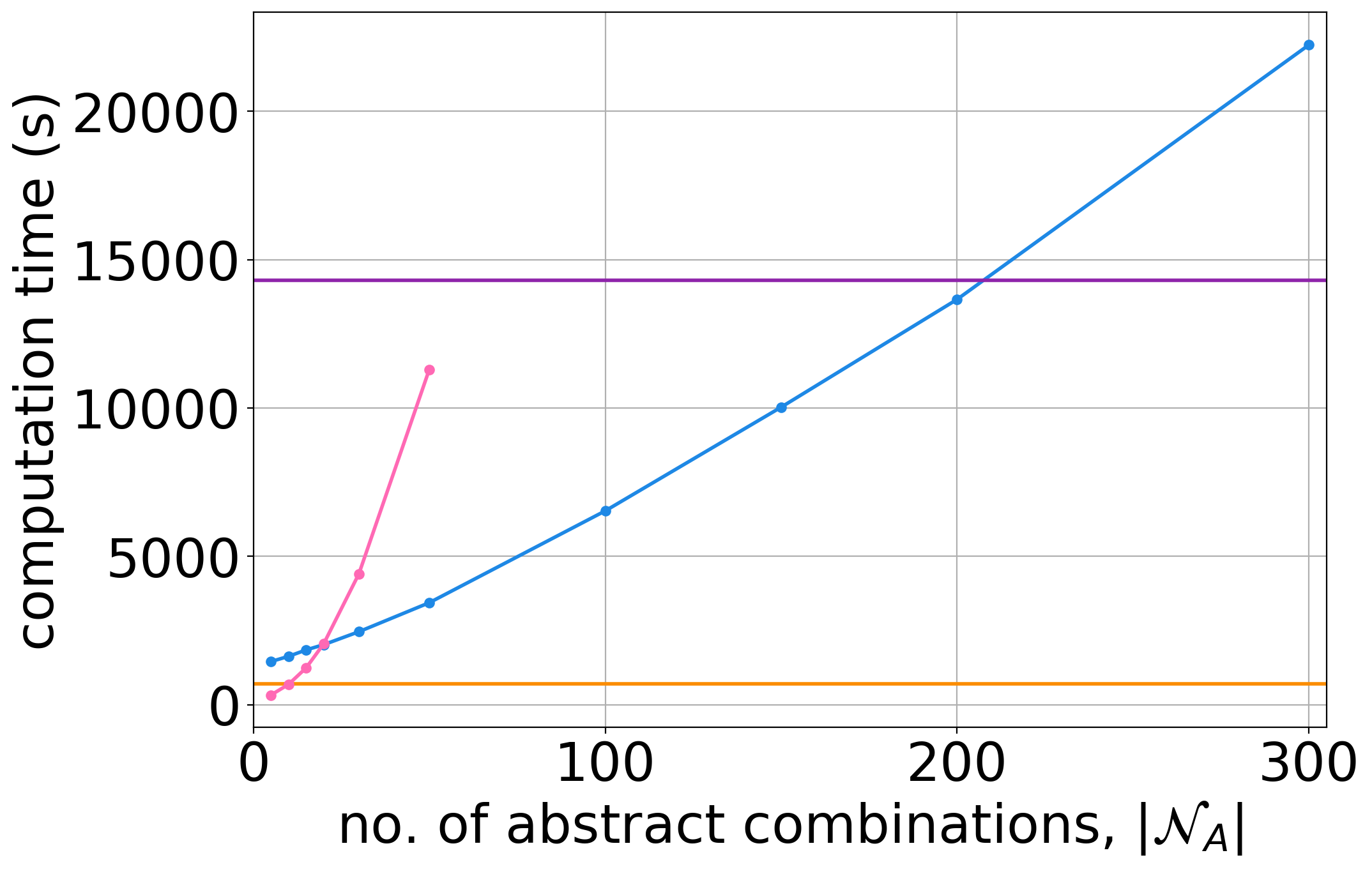}
        \caption{Synthetic Large (sinusoid arrival rates).}
    \end{subfigure} \\
    \smallskip
    \begin{subfigure}[t]{0.33\textwidth}
        \centering
        \includegraphics[width=\textwidth]{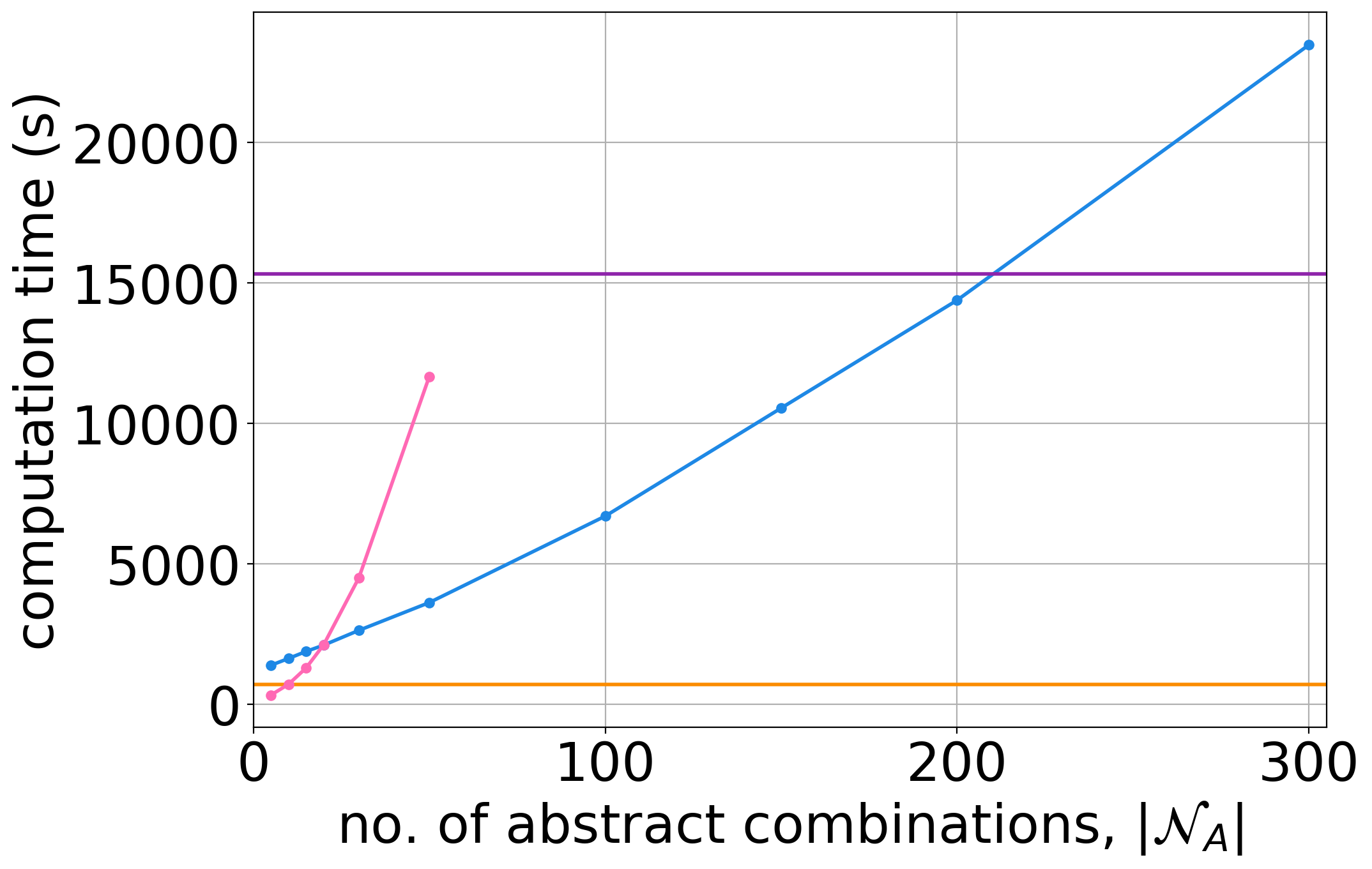}
        \caption{Synthetic Large (step arrival rates).}
    \end{subfigure}%
    ~
    \begin{subfigure}[t]{0.33\textwidth}
        \centering
        \includegraphics[width=\textwidth]{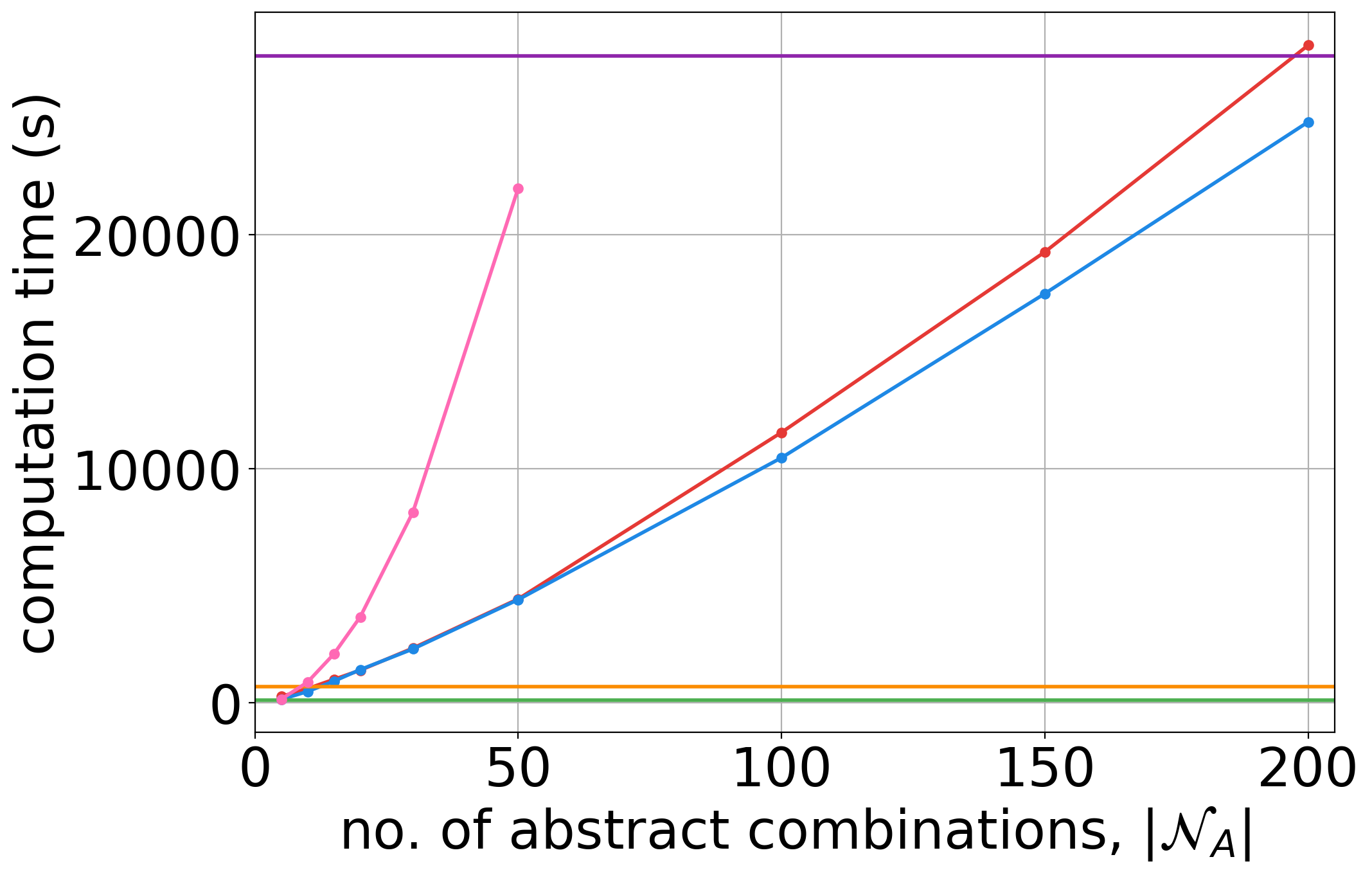}
        \caption{Public Fraud Dataset.}
    \end{subfigure} \\
    
    \caption{Computation times for each method on each domain. }
\end{figure}

\FloatBarrier
\newpage 
\subsection{Arrival Rates}

\begin{figure}[h!tb]
    \centering
    \begin{subfigure}[t]{0.47\textwidth}
        \centering
        \includegraphics[width=\textwidth]{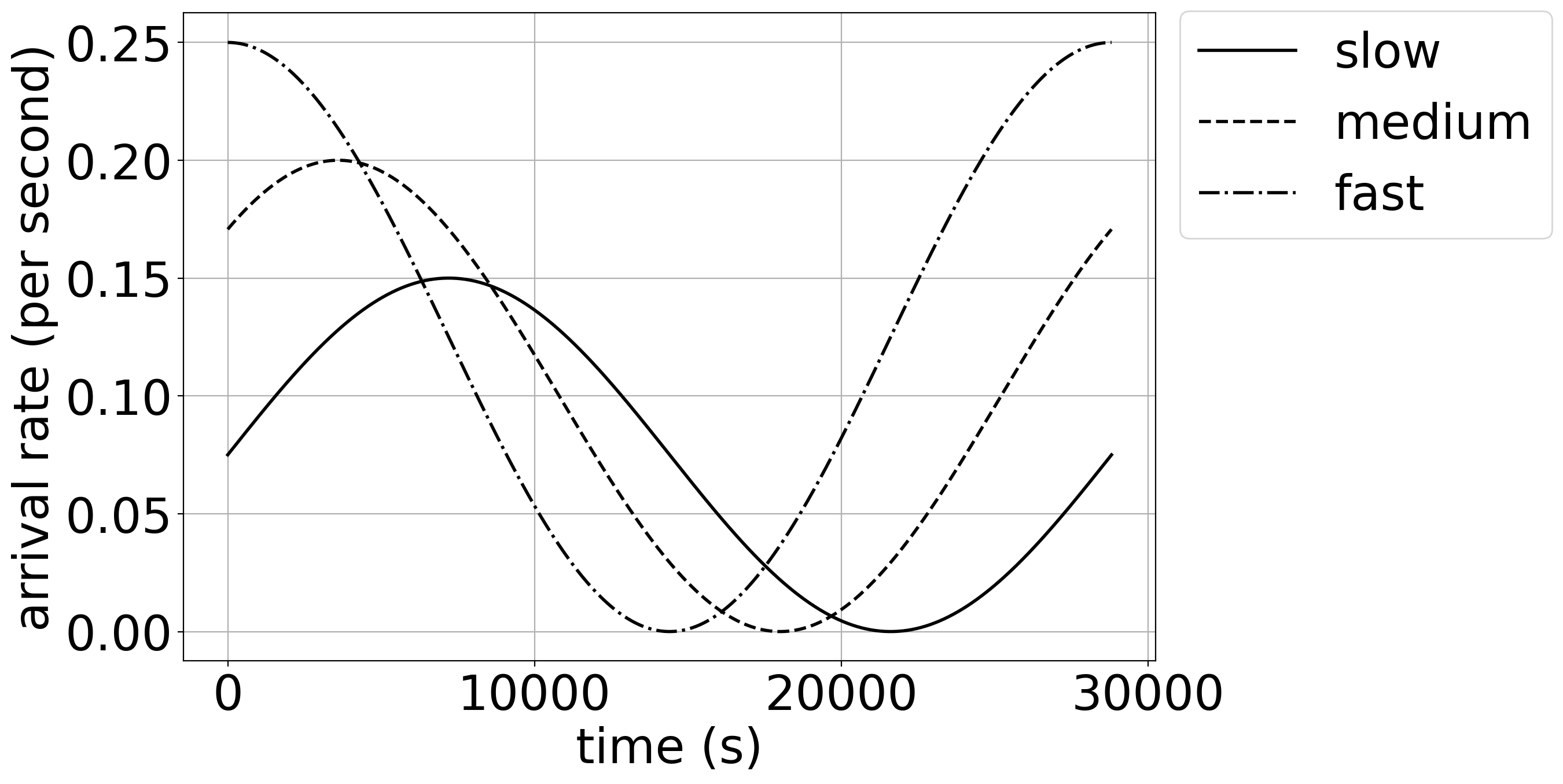}
        \hspace{-10mm} 
        \caption{Synthetic Small (sinusoid arrival rates).}
    \end{subfigure}%
    \hspace{5pt}
    \begin{subfigure}[t]{0.47\textwidth}
        \centering
        \includegraphics[width=\textwidth]{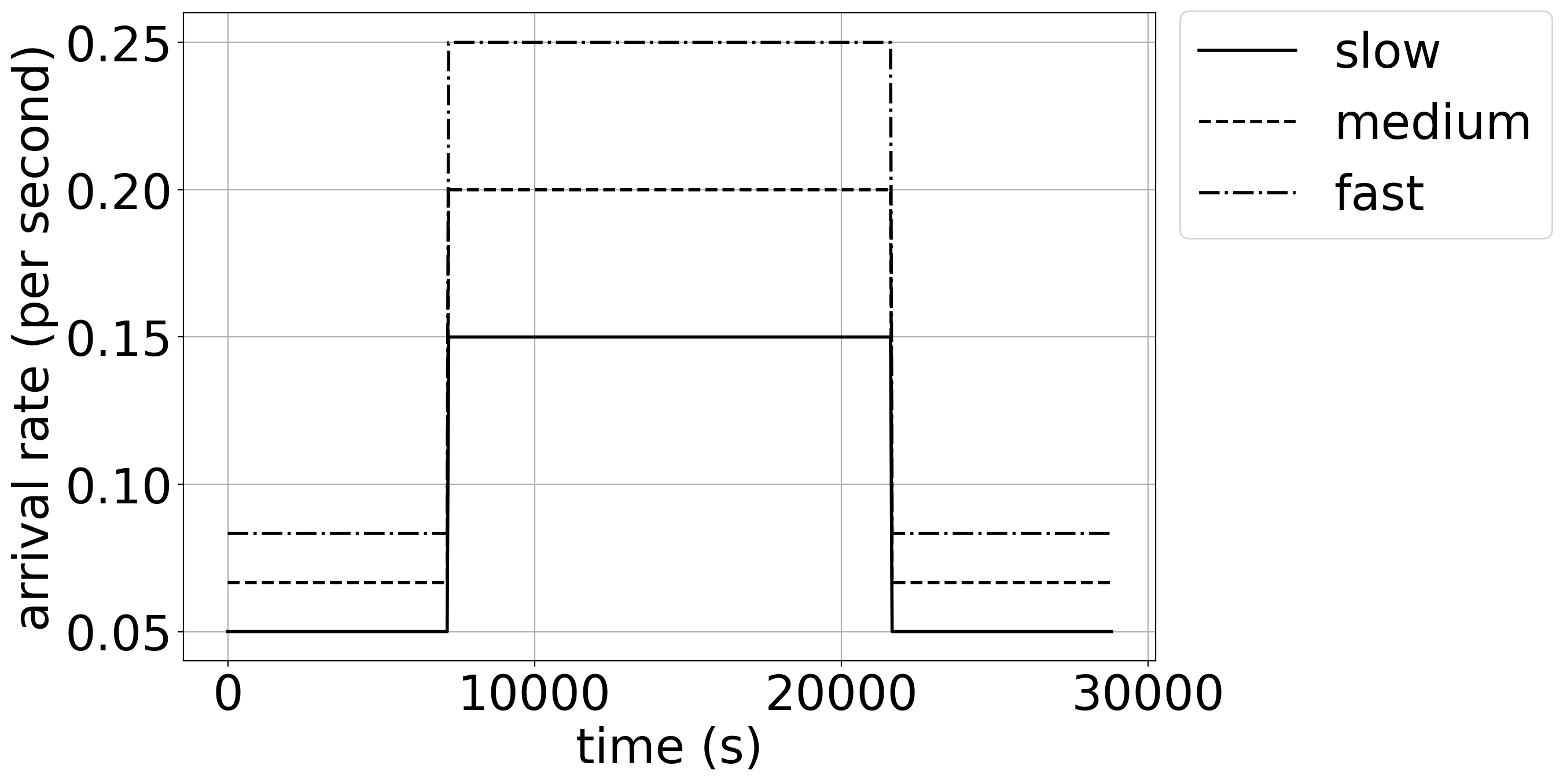}
        \hspace{-10mm} 
        \caption{Synthetic Small (step arrival rates).}
    \end{subfigure} \\
    \smallskip
    \begin{subfigure}[t]{0.47\textwidth}
        \centering
        \includegraphics[width=\textwidth]{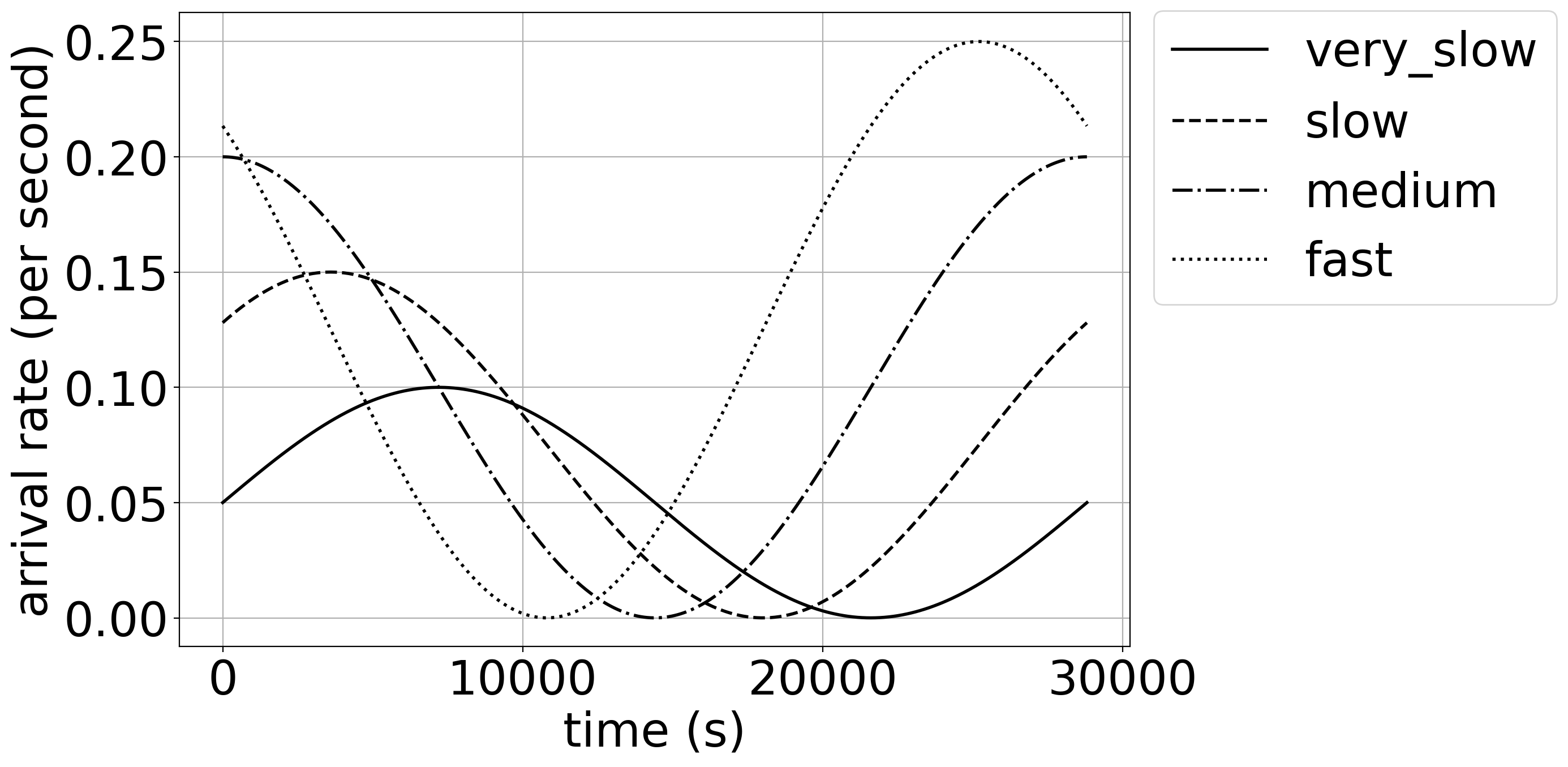}
        \hspace{-10mm} 
        \caption{Synthetic Large (sinusoid arrival rates).}
    \end{subfigure}%
    \hspace{5pt}
    \begin{subfigure}[t]{0.47\textwidth}
        \centering
        \includegraphics[width=\textwidth]{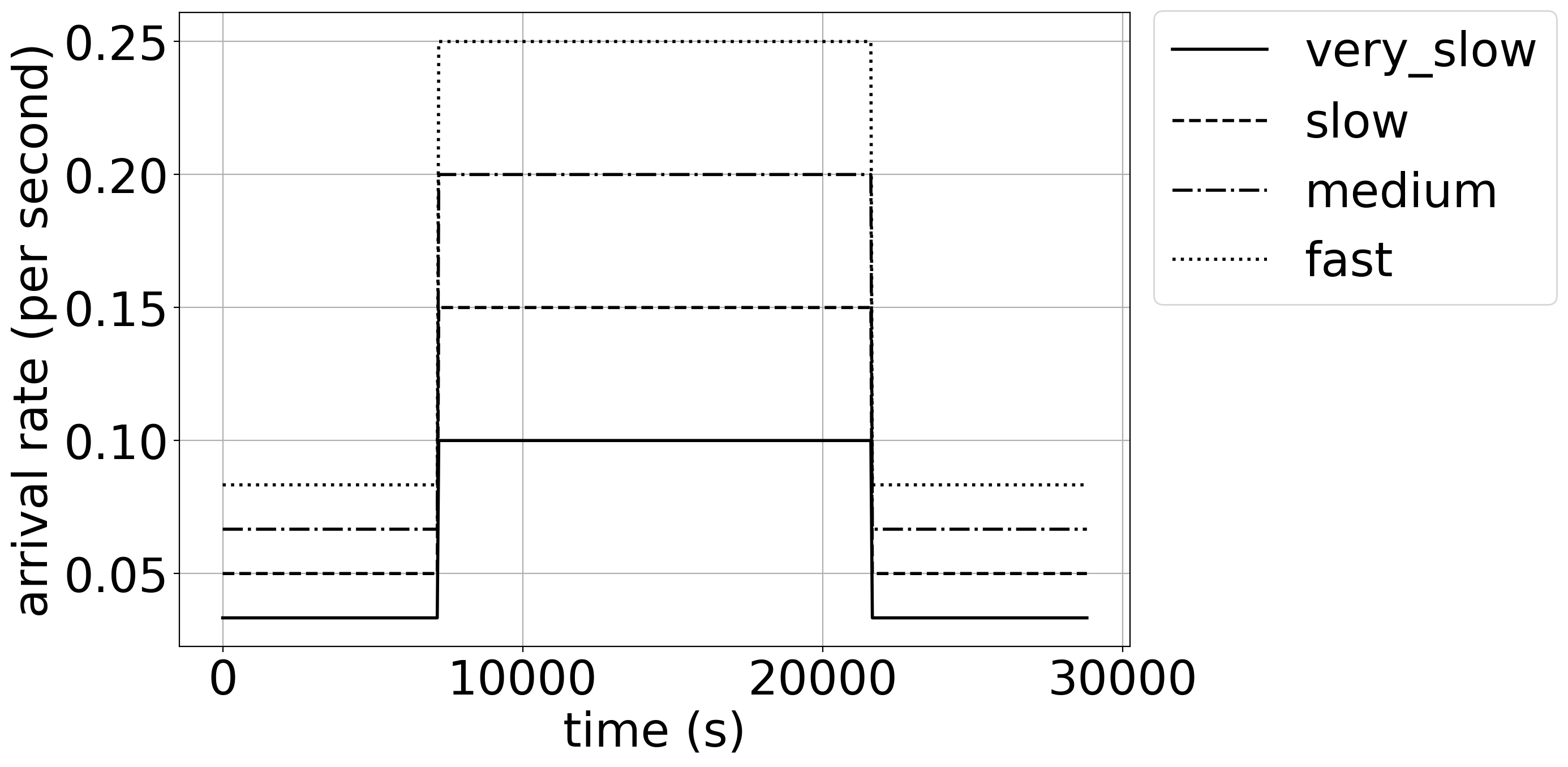}
        \hspace{-10mm}
        \caption{Synthetic Large (step arrival rates).}
    \end{subfigure} \\
    
    \caption{Arrival rates for each domain.}
\end{figure}

\end{document}